\documentclass[acmlarge,screen]{acmart}

\AtBeginDocument{%
  }

\setcopyright{acmcopyright}
\copyrightyear{2023}
\acmYear{2023}
\acmDOI{XXXXXXX.XXXXXXX}

\acmJournal{CSUR}
\acmVolume{0}
\acmNumber{0}
\acmArticle{0}
\acmMonth{6}

\usepackage{multirow}
\usepackage{subfigure}
\usepackage{wrapfig}
\begin{document}

\title{A Review and Roadmap of Deep Causal Model \\
from Different Causal Structures and Representations}

\author{Hang Chen}
\authornote{Both authors contributed equally to this research.}
\email{albert2123@stu.xjtu.edu.cn}
\orcid{0000-0002-9141-174X}
\author{Keqing Du}
\authornotemark[1]
\email{dukeqing@stu.xjtu.edu.cn}
\affiliation{%
  \institution{Xi’an Jiaotong University}
  \streetaddress{No.28, Xianning West Road}
  \city{Xi'an}
  \state{Shaanxi}
  \country{China}
  \postcode{710049}
}

\author{Chenguang Li}
\affiliation{%
  \institution{Xi’an Jiaotong University}
  \streetaddress{No.28, Xianning West Road}
  \city{Xi'an}
  \state{Shaanxi}
  \country{China}}
\email{adgjl357159@stu.xjtu.edu.cn}

\author{Xinyu Yang}
\affiliation{%
  \institution{Xi’an Jiaotong University}
  \city{Xi'an}
  \country{China}
}
\email{yxyphd@mail.xjtu.edu.cn}

\renewcommand{\shortauthors}{Hang Chen et al.}

\begin{abstract}
  The fusion of causal models with deep learning 
  introducing increasingly intricate data sets, 
  such as the causal associations within images or 
  between textual components, has surfaced as a focal research area. 
  Nonetheless, the broadening of original causal concepts 
  and theories to such complex, non-statistical data 
  has been met with serious challenges. In response, 
  our study proposes redefinitions of causal data 
  into three distinct categories from the standpoint 
  of causal structure and representation: 
  definite data, semi-definite data, and indefinite data. 
  Definite data chiefly pertains to statistical data 
  used in conventional causal scenarios, 
  while semi-definite data refers to a spectrum 
  of data formats germane to deep learning, 
  including time-series, images, text, among others. 
  Indefinite data is an emergent research sphere, 
  inferred from the progression of data forms by us. 
  To comprehensively present these three data paradigms, 
  we elaborate on their formal definitions, differences 
  manifested in datasets, resolution pathways, and 
  development of researches. We summarize key tasks 
  and achievements pertaining to definite and semi-definite 
  data from myriad research undertakings, present a roadmap 
  for indefinite data, beginning with its current 
  research conundrums. Lastly, we classify and scrutinize 
  the key datasets presently utilized within these three paradigms.
\end{abstract}


\received{20 July 2023}
\received[revised]{12 March 2009}
\received[accepted]{5 June 2009}

\maketitle

\section{Introduction}

\textbf{Causal model}, lies in between mechanistic models 
and statistical models~\cite{scholkopf2021toward}. 
Like statistical models, 
they analyze the relationships of system components 
in a data-driven approach~\cite{
goudet2017causal,kpotufe2014consistency,lopez2015towards}
\footnote{The most effective way 
for recovering causal relationships is randomized controlled trials (RCTs). 
However, conducting RCTs in the real world often proves 
time-consuming or excessively costly. 
Consequently, the causal discovery  
from observational data has become a popular choice
~\cite{mooij2016distinguishing,peters2017elements,peters2012identifiability,peters2014causal,
shimizu2006linear,sun2006causal,zhang2012identifiability}.}
. However, they possess the ability to maintain 
robustness in distribution shifts~\cite{vapnik1999nature}, 
meaning that causal models can retain accuracy out of 
$i.i.d.$ environments~\cite{pearl2000models,peters2017elements,scholkopf2012causal,spirtes2000causation}
. For instance, consider the joint 
distribution of the same system $P(X, Y)$
under two different experimental conditions. In statistical models, 
these two joint distributions may not be equivalent. 
However, by decomposing them causally as the factorization: 
$P(X)P(Y|X)$, we may obtain a robust distribution, $P(Y|X)$, 
which potentially represents $X$ as the cause of $Y$ in this system. 
When we have learned all the component relationships, 
we effectively acquire the $dx/dt$ equivalent found 
in mechanistic models.

Another domain driven by $i.i.d.$ data is machine learning, 
which has a close relationship with causal models. 
Machine learning has achieved remarkable success with extensive 
$i.i.d$. datasets~\cite{deng2009imagenet,lecun2015deep,mnih2015human,
schrittwieser2020mastering}, 
such as nearest neighbor classifiers~\cite{smola1998learning}, 
support vector machines~\cite{hearst1998support}, 
and neural networks~\cite{vapnik1999nature}. 
However, the objects accurately identified in machine learning, 
often fail to achieve the same level of correctness and unbiasedness 
in causal models~\cite{dittadi2020transfer,
goyal2020object}. 
Machine learning appears fragile 
when confronted with tasks that violate the $i.i.d.$ assumption
~\cite{kulkarni2019unsupervised,locatello2019fairness,sanchez2020learning}. 
This issue becomes more evident as machine learning, 
particularly deep learning, is applied in broader scenarios. 
Consequently, there has been a cross-pollination between the 
two fields: deep learning methods and causal discovery. 
with the efficient 
utilization and development on vast $i.i.d.$ data, deep learning 
have facilitated the emergence of causal discovery tasks 
in numerous scenarios, while causal models, 
through intervention and disentanglement, are gradually 
compensating for the generalization capacity and 
interpretability of deep learning. 
As a result, causal models are gradually being applied 
to data types involved in deep learning, 
such as computer vision~\cite{oh2021causal,guo2021causal,
hafiani2023rare,anciukevicius2022unsupervised}, 
natural language processing~\cite{yang2022survey,
sridhar2022causal,hu2021causal}, and speech recognition
~\cite{nicolson2020masked,zhang2020multi,gabler2023reconsidering}.

\begin{table}
  \caption{Highlight of existing surveys on causal model discovery}
  \label{tab:freq}
  \resizebox{\linewidth}{!}{
  \begin{tabular}{cl}
    \toprule
    Surveys&Highlights\\
    \midrule
    The development of causal reasoning~\cite{kuhn2012development}
    &initial reviews and development of causal inference\\
    Matching methods for causal inference: A review and a look forward~\cite{stuart2010matching}
    &matching methods in causal effect estimation\\
    Machine learning methods for estimating heterogeneous causal effects~\cite{athey2015machine}
    &tree-based and ensemble-based method for causal model\\
    Dynamic treatment regimes~\cite{chakraborty2014dynamic}
    &dynamic treatment regimes for estimating causal effect\\
    A survey on causal inference~\cite{yao2021survey}
    &causal effect under the potential outcome framework\\ 
    Toward causal representation learning~\cite{scholkopf2021toward}
    &Structural Causal Model and Machine Learning\\
    A survey of learning causality with data: Problems and methods~\cite{guo2020survey}
    &learning causal model from observational data\\
    Review of causal discovery methods based on graphical models~\cite{glymour2019review}
    &computational methods for causal discovery with practical issues in past three decades\\
    D’ya like dags? a survey on structure learning and causal discovery~\cite{vowels2022d}
    &continuous optimization approaches compared to other surveys\\
    Causal Discovery from Temporal Data: An Overview and New Perspectives~\cite{gong2023causal}
    &causal temporal data in multivariate time series and event sequences\\
    Causal inference for time series analysis: Problems, methods and evaluation~\cite{moraffah2021causal}
    &current progress to analyze time series causal data\\
    Granger causality: A review and recent advances~\cite{shojaie2022granger}
    &time series data between Granger causality and causation\\
    Survey and evaluation of causal discovery methods for time series~\cite{assaad2022survey}
    &causal discovery in time series with comparative evaluations\\
    Causality for machine learning~\cite{scholkopf2022causality}
    &difference between causal model and machine learning\\
    Causal inference~\cite{pearl2010causal}
    &machine learning interpretability for counterfactual inference\\
    Causal machine learning for healthcare and precision medicine~\cite{sanchez2022causal}
    &examples for illustrating how causation can be advantageous in clinical scenarios\\
    Causal machine learning: A survey and open problems~\cite{kaddour2022causal}
    &Structural Causal Model and corresponding problems in machine learning\\
    Causal inference in recommender systems: A survey and future directions~\cite{gao2022causal}
    &causal optimisation in recommendation systems\\
    Deep Causal Learning: Representation, Discovery and Inference~\cite{deng2022deep}
    &causal discovery under deep learning\\
  \bottomrule
\end{tabular}}
\label{tabcontras}
\end{table}

There exist several surveys that discuss how to discovery causal model 
from diverse senarios or methods of deep learning. 
In Table~\ref{tabcontras}, 
we list some the representative surveys and their highlight reviews.
Some reviews focus on causal inference methods, 
such as matching-based methods~\cite{stuart2010matching}, tree-based methods 
and ensemble-based methods~\cite{athey2015machine}, 
and dynamic treatment regimes methods~\cite{chakraborty2014dynamic}. 
Other reviews focus on the construction frameworks of causal models, 
such as the Granger causal model
~\cite{shojaie2022granger,gong2023causal,moraffah2021causal,assaad2022survey}, 
potential outcome frameworks~\cite{yao2021survey,kuhn2012development,gao2022causal}, 
and structural causal model~\cite{scholkopf2021toward,guo2020survey,deng2022deep,kaddour2022causal}. 
Some reviews examine the application scope of causal analysis 
in various domains, such as time series data~\cite{assaad2022survey,moraffah2021causal}, 
medical data~\cite{sanchez2022causal}, 
and machine learning multi-modal data~\cite{scholkopf2021toward,kaddour2022causal,deng2022deep}. 

Additionally, we classify these studies from two novel perspectives: 
based on whether the causal model's structure 
is fixed, we categorize them into single-structure
~\cite{kuhn2012development,stuart2010matching,athey2015machine,yao2021survey,scholkopf2021toward,glymour2019review,vowels2022d,scholkopf2022causality,pearl2010causal,sanchez2022causal,kaddour2022causal,gao2022causal,deng2022deep}
and multi-structure studies
~\cite{chakraborty2014dynamic,guo2020survey,gong2023causal,moraffah2021causal,shojaie2022granger,assaad2022survey}; 
based on the whether the causal variables need to be transformed 
to deep representation, we classify them as single-value 
~\cite{kuhn2012development,stuart2010matching,chakraborty2014dynamic,yao2021survey,guo2020survey,glymour2019review,vowels2022d,gong2023causal,moraffah2021causal,shojaie2022granger,assaad2022survey,pearl2010causal,sanchez2022causal} 
and multi-value~\cite{athey2015machine,scholkopf2021toward,scholkopf2022causality,kaddour2022causal,gao2022causal,deng2022deep} 
variable studies. The structure and variables 
are two crucial characteristics serving deep learning. 
If the causal discovery task involves multi-structure data types, 
the corresponding deep neural network should consider 
the discriminability of samples for different structures
~\cite{chen2023affective,kew2006membranous,yao2023causal,ward2016unpacking}, 
even constructing parameter-sharing modules which 
can facilitate the learning of dynamics and invariance between 
different structures~\cite{wang2022causal,yao2022learning,zimboras2022does}. 
Conversely, when dealing with data types 
that encompass multi-value variables, the causal varaibles are 
converted into deep representation, where the several statistical 
strengths need to be reexamined, including imprecise mapping 
of causal representation~\cite{veitch2020adapting,sun2022multi,xie2019boosting}, 
lack of independence and samplability~\cite{zhang2022causal,dougherty1985precursors,dai1997study}, 
and estimation of causal strength~\cite{sontakke2021causal,wang2022causalr,trabasso1985causal,ke2020amortized}. 
However, there has not been a comprehensive review summarizing 
research from these two perspectives, leading to confusion among 
researchers on which causal inference frameworks and processing  
to employ when applying deep learning to causal discovery, 
given the wide range of data types. 

Therefore, we propose three data paradigms, 
each resulting from the combination of structure quantity 
and variable complexity. The data paradigm characterized 
by a single-structure causal model and single-value variables 
is referred to the \textbf{definite data paradigm}. 
The data paradigm characterized by a multi-structure causal model 
and multi-value variables is termed the 
\textbf{indefinite data paradigm}. 
The \textbf{semi-definite data paradigm}, 
lies between the definite and indefinite paradigms, 
capturing the combination of a single-structure causal model and 
multi-value variables, or a multi-structure causal model and 
single-value variables. 
Surprisingly, there has been extensive research in the 
definite and semi-definite domains, while significant progress 
has been lacking in the indefinite data paradigm. 

To provide a detailed discussion on the existing work in the 
definite and semi-definite data paradigms, 
as well as the research gap in the indefinite data paradigm, 
our survey makes the following contributions: 

\begin{itemize}
  \item In Section 2, we introduce expanded concepts and terminology 
related to causal data. Furthermore, in Section 3, 
we propose definitions for the three data paradigms 
and analyze their differences in the computational process 
of causal discovery.

  \item In Section 4 and 5, we summarize the existing work 
  in the definite and semi-definite data paradigms, respectively. 
  \item In Section 6, we present the challenges 
  faced by indefinite data and propose corresponding theoretical 
  roadmaps. We discuss addressing theoretical issues 
  such as causal discriminability, confounding disentanglement, 
  and causal consistency. 

  \item In Section 7, we compile commonly datasets 
  for the three data paradigms. We provide information on 
  dataset sizes and typical tasks associated with them. 
\end{itemize}

\section{Preliminaries}\label{secp}
In this section, we will provide a detailed explanation of 
some relevant terminology. Particularly, due to our expansion 
of data paradigms into broader domains, 
certain previously defined terms need to be modified or redefined. 
We begin with the ``causal variables''. 
To align with data processing and mainstream task settings 
in deep learning, we have deconstructed the traditional definition 
of ``causal variables'' into three new terms: 
observed variables, causal variables, and causal representations. 
The connection from causal variables to observed variables is defined 
as follows:

\begin{definition}[Causal model]
  The basic causal model can be represented graphically as 
  $\mathcal{G} = (\mathcal{E}, \mathcal{V})$, 
  where $\mathcal{E}$ refers to causal strength, indicating 
  the direction and value of the relationship, and $\mathcal{V}$ 
  represents the set of causal representations, 
  signifying the value of each causal variable $X$ when 
  constructing this causal model. 
  \end{definition}
  
In different tasks, 
the definition of a causal model can undergo changes. 
For instance, when studying confounding factors, $V$ 
includes not only the causal representations of the 
observed variables but also those representation of latent variables. 
In the field of causal abstraction, 
a causal model may encompass intervention set $\mathcal{I}$ 
and distribution set $\mathcal{P}$. 

\begin{definition}[Structural Causal Model]
  An SCM is a 3-tuple $\langle X,\mathcal{F},\mathcal{P}\rangle $, where 
  $X$ is the entire set of causal variables  
  $X=\{X_{i}\}^{n}_{i=1}$. Structural equations 
  $\mathcal{F}=\{f_{i}\}^{n}_{i=1}$ are functions that determine 
  $X$ with $X_{i}=f_{i}(Pa_{i}, U_{i})$, 
  where $Pa_{i}\subseteq X$ represents the parent set of $X_{i}$, 
  $U_{i}$ represents the $i.i.d.$ noise term.  
  $\mathcal{P}(X)$ is a distribution over $X$. 
  \label{def3}
\end{definition}

The distinctive feature of Structural Causal Models (SCMs) 
is the construction of causal models 
using a deterministic function $f$ and a noise term $U$. 
The inclusion of the noise term allows causal relationships 
to be represented as general conditional distributions, 
e.g., $P(X_{i}|PA_{i})$. Such probabilistic representations 
not only enable interventions (e.g., $P(X_{i}|do(X_{j}))$), 
but also support counterfactual predictions 
that causal graph models cannot accommodate 
(e.g., $P(X_{i}|X_{k}, do(X_{j}))$). 

\begin{definition}[Causal variables]
The observed variables always represent the raw data in dataset 
while the causal variables represents the direct variables 
emerging in causal model. We define a projection to connect 
observed variables to causal variables which reads:
\begin{equation}
  \begin{aligned}
X&=H(S_{1}, S_{2}, \dots, S_{n})\\ 
(where:~ X&=(X_{1}, X_{2}, \dots, X_{d}))
\end{aligned}
\end{equation}
where $H$ is a transformation function, $X$ represents the causal variables 
and $S$ represents the observed variables. 
\end{definition}

\begin{definition}[Causal representation]
The causal representation $V$ represents the computed values 
of causal variables when constructing a causal model, 
i.e., the quantified values from causal variables. 
\end{definition}

\begin{table}
  \caption{Examples about observerd variables, causal varaibles, and causal representation}
  \label{tab:freq}
  \resizebox{\linewidth}{!}{
  \begin{tabular}{cllll}
    \toprule
    \textbf{Modal}&\textbf{Observed variables}&\textbf{Causal variables}&\textbf{Deep model}&\textbf{Causal representation}\\
    \midrule
    \multirow{2}{*}{Statistic Data}&Age (e.g., one observed variable: 25 years old)&Age (e.g., one causal variable: 25 years old)&-& 25 (1-dimension data)\\
    &Electric potential (e.g., 2 observed variables: V$_{A}$: 5V, V$_{B}$: 3V)&Voltage (e.g., one causal variable: $\Delta V_{AB}$=5V-3V=2V)&-&2 (1-dimension data)\\
    \midrule
    Text&Alphabet (e.g., 9 observed varaibles: \textit{``I, a, m, h, u, m, a, n, .''})&Token (e.g., 4 causal variables: \textit{``I, am, human, .''})&RoBERTa&1024-dimension tensor\\
    Image&Pixel (e.g., 32*32 observed variables: an image with 32*32 pixels)&A local part (e.g., two causal variables: figure pattern and background pattern)&LeNet-5&5*5-dimension tensor\\
    Video&Pixel  (e.g., 320*240*500 observed varaibles: a video with 500 frames)&Segment (e.g., 5 causl variables: action segments: take, pour, open, put, fold)&I3D&64*500-dimension tensor\\
    \bottomrule
\end{tabular}}
\label{tabexam}
\end{table}

The progressive transition of ``observed variables $\rightarrow$ 
causal variables $\rightarrow$ causal representations'' 
embodies the algebraic transformation of diverse data types 
from the real world phenomenon into numerical forms 
that can be effectively employed in causal modeling. For example 
in Table~\ref{tabexam}, 
we assume that observed variables of a text 
are alphabet, the causal variables could be tokens. 
The causal representations is the word embedding 
$\in \mathbb{R}^{n\times m}$, where $m$ is 768 or 1024 
given the prevalent language model RoBERTa~\cite{liu2019roberta}. 
It it noted that if the causal model seizes the relationship 
in sentences, the causal variable should turn to 
sentence rather than tokens. 
Additionally, statistic data often exhibit the 
scenario where observed variables, causal variables, 
and causal representations are identical or satisfied the linear 
projection. 

Furthermore, we observe a clear distinction between 
statistical data and modal data (non-statistical data). 
Statistical data inherently exist in numerical form and 
do not require deep representations for computations. 
On the other hand, modal data must be transformed into 
representations prior to computation. Consequently, 
we define these two types of variables as single-value variable 
and multi-value variable, respectively: 

\begin{definition}[Single-value and Multi-value variable]
\label{defsmv}

\textbf{Single-value variable}: This type of variable inherently 
exists in numerical form, and thus, there is no necessity 
for the use of deep representation.
\textbf{Multi-Value variable}: This type of variable doesn't 
inherently exist in numerical form and must be transformed 
into deep representations to enable computations.
\end{definition}

Definition~\ref{defsmv} delineates the distinction 
between single-value and multi-value variables. 
It should be noted that single-value variables are not 
necessarily defined as one-dimensional representations, and, 
correspondingly, multi-value variables are not inherently 
multi-dimensional representations, though they often 
manifest in such forms.  

Multi-value variables often facilitate 
the quantification by deep representation, such as 
text $\rightarrow$ embeddings, image $\rightarrow$ matrices, 
audio $\rightarrow$ spectrum map, and video $\rightarrow$ optical flow. 
Compared to single-value variable, it involves more complex environments. 
The statistical advantages of single-value variable are more significant, 
such as computing independence between two single-value 
variables. On the contrary, determining such ``independence'' 
among multi-value variables is challenging, 
often approximated through algorithms like cosine similarity. 
In Structural Causal Models (SCMs), one can assume that the noise 
of single-value data follows a specific distribution, 
but in multi-value data, the noise items are multi-value 
and interdependent among dimensions, causing many traditional 
causal discovery methods to make no efforts with multi-value data.

\begin{definition}[Single-structure and Multi-structure data]
For any given dataset or task, if there exists only a single 
causal graph $\mathcal{G}$, indicating a fixed causal structure 
within the dataset, we refer to such data as 
\textbf{single-structure} data. Conversely, 
if multiple causal graphs $\{\mathcal{G}\}^{M}_{m=1}$ exist (where 
$M$ represent the amount of structures), implying an 
not unique causal structure for each sample in the dataset, 
we term this as \textbf{multi-structure} data.
\end{definition}

In single-structure data, 
it is often unnecessary to consider multiple causal relationships 
between two variables. The relationship between two variables 
can be inferred directly from the statistical features 
manifested across all samples of these two variables. 
However, in multi-structure data, 
it is essential to consider whether the models are 
capable of clustering samples from the same structure, 
the robustness against low utilization of samples, 
and the generalization ability to learn the invariances 
among different structures. Simultaneously, multi-structure data 
often have trouble in low sample utilization 
since samples from other structures contribute nothing when 
identifying a specific causal structure.

\section{Definition and Difference Between Three Data Paradigms}\label{secdd}
\subsection{The Definition of Causal Data}
Recently, many studies on deep learning-based causal models 
have discussed the issue of data types. 
In addition to the number of structures and variable complexity, 
this also includes high-level tasks and low-level tasks
~\cite{5980160,chalupka2017causal,ukita2020causal,geiger2023finding}, 
high-dimensional data and low-dimensional data
~\cite{belloni2017program,li2022bounds,loh2014high,wang2020high}, 
as well as structured data and unstructured data
~\cite{mei2016signal,lally2017watsonpaths,claggett2018unpacking,badura2018can}. 
Regarding high-level and low-level tasks, 
the focus is primarily on the robustness of the models 
under different $i.i.d.$ experimental conditions, 
without distinguishing data types. 
High-dimensional and low-dimensional data are commonly 
used to indicate the richness of information, 
yet lacking clear criteria for determination. 
The discussion of structured and unstructured data 
often revolves around a specific data type, 
without disconnect the concepts of structures and variables, 
thus making it difficult to achieve comprehensive coverage.
In comparison, we consider the number of structures and 
variable complexity to be more fundamental and comprehensive 
characteristics that provide a framework for constructing 
causal models in the majority of classification 
and generation tasks. Building on the definition presented 
in Paper~\cite{chen2023learning}, we provide detailed 
definitions and examples for the three data paradigms. 

\begin{definition}[Causal Data]
  The causal relationships exist in a dataset 
  $\mathbf{D} = \{X_{s}\}^{S}_{s=1}$ which has 
  $S$ samples and $M$ ($M\geqslant 1$) causal structures 
  ($\mathcal{G}=\{\mathcal{E}_{m}, \mathcal{V}_{m}\}^{M}_{m=1}$). 
  Each structure $\mathcal{G}_{m}$ corresponds to several samples separately. 
  Hence, each sample $X_{s,m} \in \mathbb{R}^{N_{m} \times D}$ 
  belongs to a causal structure  
  $\mathcal{G}_{m}=\{\mathcal{E}_{m}, \mathcal{V}_{m}\}$ and 
  consists of $N_{m}$ variables: 
  $X_{s}=\{x_{s,m,n}\}^{N_{m}}_{n_{m}=1}$. 
  $\hat{x}_{s,m,n} \in \mathbb{R}^{1 \times D} (D \geqslant 1)$ 
  represents the causal representation of a varaible $x_{s,m,n}$ 
  where $D$ denotes the dimension of the causal representation. 
  Based on the above datasets, we define three data paradigms: 
\begin{itemize}
\item \textbf{Definite Data}: The causal structure is single-structure ($M=1$) 
and the causal variable is single-value ($D=1$).
\item \textbf{Semi-Definite Data}: The causal structure is 
single-structure ($M=1$) and the causal variable is 
multi-value ($D>1$), or the causal structure is 
multi-structure ($M>1$) and the causal variable is 
single-value ($D=1$).
\item \textbf{Indefinite Data}: The causal structure is 
multi-structure ($M>1$) and the causal variable is 
multi-value ($D>1$).
\end{itemize}
  \label{defcausalmodel}
\end{definition}

\begin{example}[Definite Data]
  Arrhythmia Dataset~\citep{647926} is a case record dataset from 
  patients with arrhythmias including 452 samples. 
  All samples contribute one causal structure with 279 
  single-value variables (e.g., age, weight, heart rate, etc.), 
  where some causal relationship are involved, 
  such as how age affects heart rate. 
  \label{exp1}
\end{example}

\begin{example}[Semi-definite Data (Multi-structure and Single-value)]
  The Netsim dataset~\citep{smith2011network} is a simulated fMRI dataset. 
  Because different activities in brain regions over time imply 
  different categories, a set of records of one patient corresponds 
  to one causal sturcture. This dataset includes 50 sturctures and 
  each sturcture consists of 15 single-value variables that measure the signal 
  strength of 15 brain regions. 
  \label{exp2}
\end{example} 

\begin{example}[Semi-definite Data (Single-structure and Multi-value)]
  CMNIST-75sp~\citep{fan2022debiasing} is a graph classification dataset 
  with controllable bias degrees. In this dataset, all researchers 
  concentrate on one causal structure including 4 causal variables: 
  causal pattern $C$, background pattern $B$, observed results $G$ 
  and label $Y$. The four causal variables all exist in the 
  form of deep representations, i.e., multi-value variables. 
  \label{exp3}
\end{example}

\begin{example}[Indefinite Data]
  IEM Dataset~\citep{busso2008iemocap} is a conversation record dataset 
  with each sample including a dialogue between two speakers. All 100 
  samples are assigned into 26 structures based on the 
  speaker identifies and turns. Each sample consists of 5-24 
  causal variables where each variable is an 
  utterance represented by word embeddings. 
  \label{exp4}
\end{example}

Definite data possesses precise quantification, 
and in most cases, the observed variables are equal to 
the causal variables or even the causal representations. 
Additionally, due to the ease of data collection, 
definite data exhibits an sufficient sample size, 
enabling significant statistical advantages. 
Furthermore, in single-structure data, the entire dataset satisfies 
the $i.i.d.$ condition, removing the need to consider 
multiple causal relationships between two variables. 
In summary, the data type characterized by 
signle-value and a single-structure is the simplest and 
earliest type involved in causal model research.

Semi-definite data encounters the problem of variables and 
structures, respectively.  Multi-value variables 
cannot be accurately quantified. 
For instance, text and audio data pose challenges in embedding. 
Although raw images can be quantified through precise mappings 
at the pixel level, such as RGB values, there still 
exist discrepancies between the resulting causal representations 
and causal variables when mapping them to high-dimensional 
feature spaces. Furthermore, a key issue with multi-structure 
data is the difficulty in ensuring structural characteristics 
as like definite data, leading to sparse and insufficient 
sample sizes. multi-structure data imposes the requirement 
for causal discovery methods to possess discriminative power 
across different structures, and even further, 
to learn the dynamics and invariances shared among distinct 
structures. These challenges necessitate the integration of 
causal discovery with deep models in order to learn from 
unstructured data in the context of semi-defined data.

Both satisfying the characteristics of multi-structure 
and multi-value data significantly increases the difficulty 
of causal discovery beyond the two types of semi-defined data. 
For instance, for multi-structure data, 
clustering can be executed accurately using the numerical 
precision of single-value variables; however, multi-value variables 
demonstrate poor discriminative performance during clustering. 
In the case of multi-value data, representation relationships 
can be learned through a fixed causal structure, 
whereas the advantage is undermined in the presence of 
multi-structure data.

\subsection{The Distinctions Among Three Data Paradigms}

Specifically, we employ the theory illustrated in
~\citet{scholkopf2021toward} to explicate why the structure (\textbf{M}) and 
variable dimension (\textbf{D}) are pivotal in capturing differences 
in causal discovery algorithms. Accroding to the assumption 
in~\citet{scholkopf2021toward}, the domain of causal variables 
$\mathcal{X}$ is projected onto the domain of causal representations $\hat{\mathcal{X}}$ 
via the encoder $p_{\varphi}$ and decoder $q_{\theta}$, 
showcasing the causal mechanism in structural equations: 

\begin{equation}
  \hat{x}_{i}=f_{i}(Pa_{i},U_{i})
\end{equation}

where $Pa_{i}$ represent the parent node set of $x_{i}$. 
For instance, $p_{\varphi}:U=(1-A)X $ and 
$q_{\theta}: \hat{X}=(1-A)^{-1}U$. Without prior knowledge, 
there exist two pathways to recover the causal model: 
1) Given a fixed causal structure and known causal representation, 
the causal strength can be estimated by the statistical strength 
observable in the samples. 
2) If encoder and decoder are feasible, optimal solutions 
of the causal model can be achieved by minimizing the 
reconstruction loss $p_{\varphi}\circ f\circ q_{\theta}$. 
Here we would like to delimit the solvability of this process 
for different combinations via M=1, M$>$1, D=1, and D$>$1.

\textbf{For a single-structure data (M=1)}: 
When the causal structure is fixed, causal strengths $f$ 
can be calculated. If the causal representation is single-value (D=1), 
the causal structure can be determined without the 
encoder $p_{\varphi}$ or decoder $q_{\theta}$. 
The reconstruction loss in this case is $f$. However, 
for multi-value data (D$>1$), in the reconstruction loss function 
$p_{\varphi}\circ f\circ q_{\theta}$, $f$ represents the 
being determined part. 

\textbf{For a multi-structure data (M$>1$)}: 
The multi-structure data induce uncertainty in causal structures, 
unclear of which samples correspond to the same causal structure 
and therefore making causal strengths $f$ unsolvable directly. 
However, under single-value (D=1) condition without generated 
representation, the precision of clustering is guaranteed. 
We can approach by first clustering the samples, 
and then separate the problem to several tasks of 
definite data problem-solving (M=1, D=1). 
In this regard, reconstruction loss amounts to $\{f_{m}\}^{M}_{m=1}$, 
representing the set containing each sub-task's $f_{m}$. 
Reconstruction loss can be regarded as a 
multi-task optimization problem, 
$\alpha_1 f_1 + \alpha_2 f_2 +\dots+\alpha_M f_M$, where $\alpha_m$ 
is the weights of the sample quantity per structure. 
The worst-case scenario arises with multi-value data (D$>$1), 
only able to attain an approximate encoder 
$\tilde{p}_{\varphi}=p_{\varphi}\circ f_m$, 
which results in a final reconstruction loss of 
$\tilde{p}_{\varphi} \circ q_{\theta}$. Causal strength $f_m$ 
comprises an unassigned part. 

In summary, for definite data, 
it suffices to identify the causal strength 
between any two causal variables under a certain causal structure. 
Semi-definite data addresses the problem of discriminating 
multi-structure structures and encoding multi-value variables 
separately. As for indefinite data, in the absence of additional 
assumptions, causal discovery in such datasets presents an 
ill-posed problem, given it requires both  
variable encoding and resolving structure discernibility. 

Beyond the distinct resolution pathways, 
the differences amongst the three data paradigms can 
also be identified in the development of the research. 
Definite data, devoid of the confusing from multi-value 
and multi-structure aspects, is dedicated to discovering 
more precise causal relationships within the 
purest experimental environments. Semi-definite data 
emerges from the convergence of causal discovery and deep learning, 
aiming to leverage the formidable pattern recognition 
and fitting capabilities of deep learning to address the 
challenges of multi-value variables or multi-structure data. 
Indefinite data presents a promising future issue. 
As a data paradigm that is closest to the real world, 
it necessitates resolving specific problems in certain scenarios, 
of course, within a framework that simultaneously handles 
multi-value and multi-structure issues. 

\section{The Tasks and Exsiting Work about Definite Data Paradigm}

In this section, we demonstrate the research progress 
on single-structure and single-value data types by 
introducing different tasks related to the data paradigm 
and their corresponding existing works. 

\begin{itemize}
\item The first task is causal discovery based on observed variables, 
aiming to recover the causal model or partial causal model 
of a complete and non-confounder set of observed variables 
through various methods. We summarize an overview 
of traditional causal discovery methods (e.g., 
constraint-based methods, score-based methods, 
and SCM-based methods) as well as recent works that 
incorporate deep learning. 

\item The second task involves causal discovery with confounders, 
aiming to estimate and recover the causal model under 
assumptions that various confounding factors exist 
(e.g., assuming confounders have a pervasive influence on 
all observed variables or assuming the presence 
of only one confounder as a parent node of observed variables). 
These studies include methods based on graphical causal models 
and SCMs. 

\item The third task is causal effect estimation, 
aiming to estimate the process by which the value of an 
observed target achieves an ideal value when the value of a 
treatment target is changed. This task requires the 
prerequisite of recovering a causal model or 
combining insights from causal models with effect estimation. 
These studies mostly rely on the potential outcome frameworks , 
specifically the Rubin causal models (RCMs), 
and can be categorized according to the classification 
provided in Review~\cite{yao2021survey}, 
including re-weighting methods, stratification methods, 
matching methods, tree-based methods, representation-based methods, 
multi-task methods, and meta-learning methods. 

\end{itemize}

\subsection{Causal Discovery based Observed Variables}

Traditional methods are primarily based on the assumptions 
of causal Markov condition~\cite{kang2009markov} 
and causal faithfulness~\cite{balashankar2021learning}. 
These methods include constraint-based approaches 
(such as IC~\cite{pearl2000models}, 
PC~\cite{spirtes2000causation,kalisch2007estimating}, 
FCI, CD-NOD~\cite{colombo2012learning,zhang2008completeness}), 
score-based methods (such as 
GES~\cite{chickering2002optimal,hauser2012characterization}, 
fGES~\cite{ramsey2017million}), as well as methods based on 
functional causal models (FCMs)(such as 
LiNGAM~\cite{shimizu2006linear}, ANM~\cite{hoyer2008nonlinear}, 
PNL~\cite{zhang2015estimation}, IGIC~\cite{janzing2012information}, 
and FOM~\cite{cai2020fom}). The first two categories often 
result in PDAGs, which present a challenge in distinguishing 
causal relationships within Markov equivalence classes. 
For example, the independence relations for 
$X \rightarrow Y \leftarrow Z$ and $X \rightarrow Y \rightarrow Z$ are the same, 
where $X \perp \!\!\! \perp Z| Y$, $X \not \! \perp \!\!\! \perp Y$, 
and $Z \not \! \perp \!\!\! \perp Y$. The FCM-based methods address 
this issue by introducing noise terms. Moreover, 
these methods have provided new insights into utilizing 
higher-order statistical quantities. For instance, 
$A \perp \!\!\! \perp B :=_{t-sep=d-sep}rank(cov(A,B))\leqslant 
C_{A}+C_{B}$~\cite{spirtes2013calculation}; the asymmetrical measure 
of the fourth-order $E[E[(y-f(x))^{4}|x]]$ can determine 
the causal direction of heteroscedastic data~\cite{cai2020fom}. 

Recent research in causal discovery has introduced 
deep generative models that can generate counterfactuals 
and model causal graphs, thereby enabling dimensionality reduction 
and structured processing of complex data. With the advent of 
NOTEARS~\cite{zheng2018dags}, causal discovery has been 
transformed into a continuous optimization problem, 
where its assumptions on Structural Equation Models (SEMs) 
are as follows: 
\begin{equation}
\mathcal{G}(A)\in DAGs\Leftrightarrow h(A)=0\Leftrightarrow tr(e^{A\circ A})-d=0 
\label{eqatr}
\end{equation} 
Equation~\ref{eqatr} above algebraically represents the 
acyclicity of a graph as the trace of the Hadamard product 
of its adjacency matrix, laying the foundation for nonlinear 
causal discovery. Subsequently, numerous studies have extended 
the continuous optimization and probabilistic inference methods 
of causal discovery through various baseline models. 
In Table~\ref{tabobcd}, we present the baseline models and key modifications 
of these methods. DAG-GNN and CasualVAE leverage the 
VAE framework to obtain causal structures and decouple 
causal representations. RL-BIC/BIC2 and CORL combine 
reinforcement learning methods with causal discovery, 
greatly enhancing the algorithm's search capability. DAG-GAN, 
CAN, and cGAN are all based on GANs to simulate causal 
generative mechanisms. GAE, AEQ, and CASTLE utilize auto-encoders (AEs) 
to reconstruct causal features.  

\begin{table}
  \caption{Overviews of deep causal models for causal discovery}
  \label{tab:freq}
  \resizebox{\linewidth}{!}{
  \begin{tabular}{clll}
    \toprule
    Base models&Methods&Contributions&Cores\\
    \midrule
    \multirow{2}{*}{FNN}&NOTEARS~\cite{zheng2018dags}&transformation from combinatorial optimization to a continuous optimization&$h(A)=0\Leftrightarrow tr(e^{A\circ A})-d=0$\\
    &GarN-DAG~\cite{lachapelle2019gradient}&it can handle parameter families of various conditional probability distributions&$C\triangleq |W^{(L+1)}|\dots|W^{(2)}||W^{(1)}|$\\
    \midrule
    \multirow{2}{*}{VAE}&DAG-GNN~\cite{yu2019dag}& generalization of linear SCMs via generative models and variational Bayesian&$\frac{1}{2} \sum_{i = 1}^{m}\sum_{j = 1}^{d}  Z^{2}_{ij}=\frac{1}{2}||(I-A^{T})||^{2}_{F}$\\
    &CausalVAE~\cite{lin2022cascade}&decoupling causal representation enhances model interpretability&$KL(q_{\varphi}(\hat{Z}|\hat{X} ||p_{B}(\hat{Z})))\thickapprox E_{q_{\varphi}(\hat{Z}|\hat{X})}[log\frac{D(\hat{Z})}{1-D(\hat{Z})}]$\\
    \midrule
    \multirow{2}{*}{RL}&RL-BIC~\cite{zhu2019causal}&automatic determination of positions to search&$reward:=-(\mathcal{S}(\mathcal{G})+\lambda_{1}I(\mathcal{G}\notin DAGs)+\lambda_{2}h(A))$\\
    &CORL~\cite{wang2021ordering}&finding suitable variable sequences in the variable space&$reward:=\sum_{k = 1}^{m}log p(x^{k}_{j}|U(x^{k}_{j});\theta_{j})-\frac{\theta_{j}}{2} log m $\\
    \midrule
    \multirow{3}{*}{GAN}&DAG-GAN&acyclic nature of the adjacency matrix parameterized by the generator&$W_{i,j}=||ith - column(A^{(1)}_{j})||$\\
    &CAN~\cite{moraffah2020causal}&generated samples based on conditional and intervention distributions&$tr((I+\beta A\odot A)^{n})-n=0$\\
    &cGAN~\cite{ng2019graph}&condition information is added into the generator and the discriminator&$\min_{G} \max_{D} V(D,G)=log D(x|a)+log (1-G(z|a))$\\
    \midrule
    \multirow{3}{*}{AE}&GAE~\cite{xu2019modeling}&graph structure information is fully utilized to discover causal structures&$f(X^{j},A)=g(A^{T}H^{j})$\\
    &AEQ~\cite{galanti2020critical}& reconstruction error is introduced to distinguish causal directions&$L(G,F,R)_{re}=\frac{1}{m}\sum_{i = 1}^{m}||G(F(a_{i}),R(b_{i}))-b_{i}||^{2}_{2}$\\
    &CASTLE~\cite{kyono2020castle}&regularization is introduced to reconstruct features with causal adjacency&$\hat{W} \in \min_{W} \frac{1}{N}||Y-\tilde{X}W_{:,1}||^{2}+\lambda\mathcal{R}_{DAG}(\tilde{X},W)$\\
    \bottomrule
\end{tabular}}
\label{tabobcd}
\end{table}

\subsection{Causal Discovery based Latent Confounders}
\subsubsection{Graph-based Methods}

Methods based on graph primarily learn the causal structure 
with confounding factors either by making assumptions 
from a graph representation perspective on latent variables 
or by modeling conditional independence on the MAG. 
The trek-separation~\cite{sullivant2010trek} 
separates variables by blocking all common parent nodes 
and directed paths between variables, under the constraint 
\begin{equation}
Rank(\sum_{A,B})\leq min {|C_{A}|+|C_{B}|}
\end{equation}
This approach can identify latent variables within 
linear Gaussian models via clustering observed variables, 
such as vanishing tetrad~\cite{silva2006learning}: 
$t_{i,j,u,v}=det(cov([ij],[uv]))=\rho_{iu}\rho_{jv}-\rho_{iv}\rho_{ju} =0$, 
FOFC~\cite{kummerfeld2016causal}, LFCMs~\cite{squires2022causal}. 
Based on bipartite, Unique Cluster $\&$ Triple-Child, 
and Double Parent assumptions, observed variables are 
divided into different clusters. Through finding the 
topological order of the maximum clusters, 
the causal strength matrix of clusters and confounding factors 
can be calculated, which allows for the recovery of complex 
latent causal structures even with low sample sizes. 
However, these methods fail when the number of observed variables 
for a latent variable is less than 3. 

The Triad constraint~\cite{cai2019triad} 
assumes that each latent variable has at least two pure children. 
It designs a ``pseudo-residual" for any three variables: 
\begin{equation}
  E_{(i,j|k)}:=X_{j}-\frac{Cov(X_{i},X_{k})}{Cov(X_{i},X_{k})} \cdot X_{j}
\end{equation}
which can identify the linear causal direction 
between latent variables with non-Gaussian noise. 
Similar methods all depend on covariance matrix rank constraints 
and employ certain high-order statistics. 
For instance, \cite{anandkumar2013learning} identifies 
latent factors through extracting second-order statistics, 
while~\cite{huang2020causal} and~\cite{zhang2017causal} 
consider a special type of confounding factor 
caused by distribution changes. 

Many methods interpret latent confounding based on 
equivalence between m-separation on a maximal ancestral graph (MAG) 
and d-separation on a directed acyclic graph (DAG). 
\cite{kocaoglu2019characterization} proposes the \textit{do-see} principle: 
$P_{I}(y|w)=P_{J}(y|w)$, which captures the I-Markov equivalence 
representation between two causal graphs with latent variables 
given an intervention set I, and consequently recovers the 
causal structure by learning a Consistently Extended 
$MAG_{\mathcal{M}}=MAG(Aug_{I}(D_{2}))$. 
The L-MARVEL algorithm~\cite{akbari2021recursive} 
obtains a Markov equivalence class of the MAG based on 
recursive equivalence constraints 
$(X\bot Y|Z)_{\mathcal{M}}\Leftrightarrow (X \perp \!\!\! \perp Y|Z)_{P_{O|S}}$ 
in the presence of latent variables and selection bias. 
\cite{bernstein2020ordering} introduces a mapping from a 
biased set (posets) to DMAGs, the sparse topology formula, 
such that the Ancestral Graph $AG(\pi ,\mathbb{P})$ meets 
the condition $derected edge set: \{i \rightarrow j: i \preccurlyeq_{\pi} j, X_{i} \not \! \perp \!\!\! \perp X_{j}| X_{pre_{\pi}(i,j)/(i,j)}\}$, 
and $bidirected edge set: \{i \leftrightarrow j: i \npreceq \nsucceq {\pi} j, X_{i} \not \! \perp \!\!\! \perp X_{j}| X_{pre_{\pi}(i,j)/(i,j)}\}$ 
then transforms AG to the Minimal IMAP. 
It proves that, under weakened faithfulness assumptions, 
any sparse independent mapping of the distribution belongs 
to the Markov equivalence class of the true causal model. 
Regarding unmeasurable latent confounding, 
\cite{bhattacharya2021differentiable} introduces 
differentiable constraints 
$trace(e^{D})-d+sum(e^{D}\circ B)=0$ and $trace(e^{D})-d + GREENERY (D,B)=0$
on various types of ADMG and 
uses continuous optimization to discover causal structures. 
However, in practical tasks, it is usually sufficient to 
estimate the causal effects of specific variables. 
The ACIC algorithm~\cite{wang2021actively} 
proposes a new graph characterization, 
the Minimal Conditional Set (MCS), as the equivalent condition 
of $P(Y|do(X=x))=\int P(\bar{D})P(Y|\bar{D},X=x) ~d\bar{D}$ 
in MAG, which effectively estimates all possible 
causal effects in PMAG. 

The latest methods introduce nonparametric assumptions, 
eliminating the need to presume a known number of latent variables, 
and each latent variable requires at most one unknown intervention. 
GSPo~\cite{jiang2023learning} defines two new representation: 
Imaginary subsets and Isolated edges. 
It then learns the bipartite graph, $G_{B}$, 
in accordance with the constraints of the maximal measurement model 
that $Pa(X_{i}) \neq \emptyset $ and $\{\mathcal{T}_{X}(G^{(I)})\}_{I\in \mathcal{I}}=\{\mathcal{T}_{X}(G^{(I')})\}$ 
using this to recover the latent DAG, $G_{H}$.

\subsubsection{SCM-based Methods}

Methods based on structured causal models (SCMs) primarily 
involve making assumptions and constraints on latent variables 
or noise distributions from the perspective of 
Structural Equation Models (SEM). They then recover causal 
structures via independence tests conducted on SCMs. 

If the latent confounding factor follows a Gaussian distribution, 
\cite{chen2013causality} can identify the causal order 
and causal effects of observed variables when the external noise 
of observed variables is super-Gaussian or sub-Gaussian. 
In methods for learning non-Gaussian linear causal models, 
\cite{hoyer2008estimation} introduces latent variables into SEM, 
normalizing non-Gaussian linear SEMs and proposes the 
lvLiNGAM method. \cite{entner2011discovering} 
recovers all unconfounded sets to learn the causal relationship 
of each pair of variables in the set, identifying the 
causal structure. However, when the latent confounding factor 
is the parent node of most observed variables, 
it will fail and return empty unconfounded sets. 

\cite{salehkaleybar2020learning} rewrites the total causal effect 
of $V_{i}$ on $V_{j}$, denoted as $B=(I-A)^{-1}$, 
in an expression containing only zero-order matrices, 
$B=\sum ^{d-1}_{k=0}A^{k}$. Based on the assumption of 
faithfulness and the reducible definition of the matrix, 
it determines the number of system variables, 
and through the over-complete ICA method, 
it identifies the unique causal order amongst variables. 
CGNN~\cite{goudet2018learning} formally provides the definition 
of $FCM(\mathcal{G},f,\mathcal{E})$ in the presence of latent confounding:
$X_{i}=f_{i}(X_{Pa(i;\mathcal{G})},E_{i})$, where $E_{i}$ includes 
all unobserved variables, thus introducing max average pool layer: 
$\widehat{MMD}_{k}(D,\hat{D})=\frac{1}{n^{2}}\sum ^{n}_{i,j=1}k(x_{i},x_{j})+\frac{1}{n^{2}}\sum ^{n}_{i,j=1}k(\hat{x}_{i},\hat{x}_{j})-\frac{2}{n^{2}}\sum ^{n}_{i,j=1}k(x_{i},\hat{x}_{j}) $ 
to minimize the difference between generated variables and observed 
varaibles. SAM~\cite{kalainathan2019generative} introduces the 
noise matrix and differentiable parameters to the new definition: 
$X_{j}=\hat{f}_{j}(X,U_{j})=\sum^{n_{h}}_{k=1}m_{j}\varphi_{j,k}(X,U_{j})z_{j,k}+m_{j,0}$, 
settling down the computing limitation of CGNN. 
GIN~\cite{xie2020generalized} leverages the non-Gaussianity of 
data and independent noise to estimate latent variable graphs. 
It presumes that no observed variable $X$ is an ancestor of 
the latent confounder $L$. If the noise 
$E_{Y||X}:=\omega^{T}Y$ between $(Z,Y)$ is 
independent of $Z$, then it is considered that $(Z,Y)$ satisfies 
the GIN condition. Based on the GIN condition, the authors 
propose a recursive algorithm that learns the causal order 
of latent variables through causal clustering and thus identifies 
the causal structure.~\cite{xie2022identification} relaxes the 
existing methods' default assumption that all child variables 
of a latent variable are observable and allows child variables 
to also be latent, that is 
\begin{equation}
  X_{i}=\sum _{L_{j}\in Pa(X_{i})}b_{ij}L_{j}+\varepsilon_{X_{i}}
\end{equation}
\begin{equation}
  L_{i}=\sum _{L_{k}\in Pa(X_{j})}c_{jk}L_{k}+\varepsilon_{L_{j}}
\end{equation}
It introduces the concept of a minimal latent layer  
and proposes the LaHME algorithm, which is based on the 
GIN condition, to learn linear non-Gaussian causal latent 
hierarchical models.

Some methods introduce counterfactuals and interventions 
to learn latent DAGs.~\cite{ahuja2022weakly} employs 
weak supervision to learn latent DAGs from pairs of 
counterfactual data, relaxing the assumptions about the 
distribution of latent variables. It assumes that each latent 
variable $Z$ is sampled from an arbitrary unknown 
$P_{Z}$, and the entire data generation process follows 
$z \thicksim  P_{Z}, x \leftarrow g(z), \tilde{z}_k \leftarrow z + \delta_{k}, \tilde{x}_k \leftarrow g(\tilde{z}_k), \forall k \in I$ 
(the intervention set), thereby enabling the identification 
across arbitrary continuous latent distributions.
SuaVE~\cite{liu2022identifying} focus on the representation-context 
linear Gaussian model with $d$ latent variables, 
non-linear mixed function, which has hypotheses: 
$n_{i}\sim \mathcal{N}(\beta_{i,1}(u),\beta_{i,2}(u)), z_{i}:=\lambda^{T}_{i}(u)(z)+n_{i}, x:=f(z)+\varepsilon$ 
where $u$ could be regarded as naturally soft interventions. 
~\cite{seigal2022linear} defined linear latent model: 
$Z=A_{k}Z+\Omega^{\frac{1}{2}}_{k}\varepsilon, cov(\varepsilon)=I_{d}$, 
to utilize prefect intervention $B_{k}=B_{0}+e_{ik}c^{T}_{k}$ and 
RQ factorization of matrix enhance causal deconfounding. It proves that 
the single intervention on each latent varaible enables identifying 
latent causal model. Moreover, DeCAMFoundER~\cite{agrawal2021decamfounder} 
employs spectral decomposition to recover causal graphs 
under non-linear and pervasive conditions. 
It assumes the existence of a statistic $S=E[x|h]$to replace 
$x=Bx+\Theta h+\varepsilon $, and estimates the causal graph 
$\mathcal{G}^{*}$ based on spectral estimators and score functions.

\subsection{Causal Effect Estimation}

Estimation of causal effects typically rely on  
three crucial assumptions: Stable Unit Treatment Value Assumption 
(SUTVA), ignorability, and positivity~\cite{yao2021survey}. 
SUTVA assumes that each unit is independent and that there exists 
only one version of each treatment. Ignorability establishes 
an independent relationship between the background variables $X$, 
the treatment $W$, and the latent outcomes: 
$W \perp \!\!\! \perp Y (W=0), Y(W=1)|X$. 
Positivity hypothesizes that for any given values of $w \in W$ and 
$x \in X$, the treatment assignment probability 
$P(W=w|X=x)>0$. Grounded within these assumptions, 
the core of effect estimation lies in addressing selection bias 
induced by confounding factors, ultimately rectifying estimation 
bias and spurious effects. 

\subsubsection{Re-weighting methods}
The first category of solution tackles selection bias by creating 
an approximation of a pseudo group. Re-weighting methods 
fall into this category, which work by assigning appropriate 
weights $e(x)$ to each unit in the observed data, 
thereby forging a pseudo-population to make the treatment and 
control groups similarly distributed. Certain techniques 
rely on propensity scores (such as inverse probability weighting (IPW)) 
to re-weight samples. In order to resolve the high dependency 
of the IPW estimator on defended propensity, 
\cite{imai2014covariate} introduced the Covariate Balancing 
Propensity Score (CBPS) approach. This method models treatment 
allocation while optimizing covariate balance. 
CBPS leverages the propensity score as dual characteristics 
of the covariate balancing score and the conditional probability 
of treatment assignment, estimating propensity scores 
through the evaluation of the following matric conditions:

\begin{equation}
  \frac{1}{N}\sum_{n = 1}^{N} \{\frac{S_{n}\tilde{X}_{n}}{\pi_{\beta}(X_{n})} -\frac{(1-S_{n})\tilde{X}_{n}}{1-\pi_{\beta}(X_{n})} \}=0   
\end{equation}

Covariate Balancing Generalized Propensity Score (CBGPS)
~\cite{fong2018covariate} extends CBPS to continuous treatments, 
demonstrating that the balance condition for covariates is 
equivalent to $\mathbb{E}(w_{n}T^{*}_{n}X^{*}_{n})=\mathbb{E}(T^{*}_{n})\mathbb{E}(X^{*}_{n})=0$, 
and subsequently identifies the optimal weight 
by constraining $\sum_{n = 1}^{N} \log f(T^{*}_{n})+\log f(X^(*)_{n})-\log w_{i} $. 
Additionally, IPM estimator possesses stability issues 
with infrequent treatment assignments at the tail. 
\cite{ma2020robust} enhances the estimator's stability 
for small probability weights by adapting different asymptotic 
distributions, thereby constructing the bias term: 
  $\hat{\mathsf{B}}_{n,b_{n}}=-\frac{1}{N} \sum_{n = 1}^{N} (\sum_{j = 0}^{p}\hat{\beta}_{j} \hat{e}(X_{i})^{j} ) (where:~\hat{e}(X_{i})\leqslant b_{n})$
to correct the estimation of $\hat{\theta}^{bc}_{n,b_{n}}$ in 
bias-corrected estimator. \cite{li2018balancing} proposes the utilization 
of weights $h(x)=e(x)(1-e(x))$ to minimize the asymptotic 
variance of the average treatment effect among a class of 
balanced weights, which serve to balance the weighted distribution 
of covariates across treatment groups. Other methods re-weight 
samples and covariates to balance confounding factors. 
The Data-Driven Variable Decomposition (D$^{2}$VD) algorithm
~\cite{kuang2017treatment} automatically distinguishes 
between confounders and adjustment variables, 
breaking the convention set by existing methods that treat all 
variables as confounders in effect estimation.

\subsubsection{Stratification methods} Stratification methods 
adjust the biases between 
the treatment and control groups by evenly distributing 
the samples from both groups into blocks. The critical aspects 
of such methods are the creation of these blocks and the 
way they're combined. Equi-frequency methods
~\cite{rosenbaum1983central} split blocks according to 
occurrence probabilities (such as propensity scores), 
so that each sub-group (block) has the same occurrence 
probability (i.e., propensity score). To reduce variance, 
\cite{hullsiek2002propensity} reweights the treatment 
effects specific to each block based on equi-frequency methods. 
To address the impact of treatment on intermediate outcomes, 
\cite{frangakis2002principal} constructs subgroups 
based on the pretreatment potential outcomes, 
to capture the true causal effects of the treatment results.

Apart from addressing bias introduced by confounding factors, 
another challenge in effect estimation is the lack of 
\textbf{counterfactuals}. Match-based methods can estimate these 
counterfactuals and simultaneously reduce estimation biases 
induced by confounders. The core idea is that a single sample, 
within the same spacetime, will not exhibit simultaneous outcomes. 
However, in effect estimation, the counterfactual for a node 
can be inferred using the average of similar neighboring nodes. 
~\cite{stuart2010matching} proposes a distance measure 
based on linear propensity scores $D_{i,j}=|\log (e_{i})-\log (e_{j})|$, 
which ensures that two units are highly similar when evaluated 
on propensity score measures and covariate comparisons. 
The Hilbert-Schmidt Independence Criterion (HSIC)
~\cite{chang2017informative} employs a measure 
($D(x_{i},x_{j})=||W^{(t)T}(x_{i}-x_{j})||_{2}$), 
which maximizes the nonlinear dependence between candidate 
subspaces and measured outcome variables 
$max_{w^{(t)}} HSIC (Z^{(t)}, Y^{F(t)})-\mathcal{R}(W^{(t)})$. 
This allows for a more accurate estimation of counterfactuals 
and avoids the influence of pre-variables on estimation outcomes. 
In terms of matching algorithms, common choices include 
Nearest Neighbor Matching (NNM), caliper, stratification, 
and kernel. Stratified matching~\cite{rosenbaum1985constructing} 
partitions the common support of propensity scores into a 
series of intervals and calculates the average of outcome 
differences between treatment and control within each interval 
to measure the effect. Coarsened Exact Matching (CEM)
~\cite{iacus2012causal} tackles the issue, overlooked 
by other methods, of extrapolative regions, 
where other treatment groups have few or no reasonable matches.

\subsubsection{Decision-tree Methods} 
Methods based on decision trees predict the values 
of target variables according to decision rules learned from data. 
To estimate the heterogeneity of causal effects,
~\cite{athey2016recursive} proposed a data-driven method based 
on Classification and Regression Trees (CART), 
alleviating the assumption of covariate sparsity. 
Similarly, Bayesian Additive Regression Trees (BART)
~\cite{hill2011bayesian} not only readily identifies 
heterogeneous treatment effects but also facilitates more accurate 
estimation of average treatment effects in non-linear simulation 
scenarios. In order to mitigate issues that prior distributions 
of treatments are often difficult to ascertain,~\cite{hahn2020bayesian} 
uses interpolation techniques and regression trees to model 
the conditions of treatment and control entirely independently, 
thereby simplifying modeling response variables as functions 
of binary treatment and control variables. 
Further, random forest models can also be employed for 
heterogeneous treatment effects. The Causal Forest suggested by 
~\cite{wager2018estimation}, as an adaptive nearest-neighbor method, 
effectively leverages non-parametric estimation for bias correction: 

\begin{equation}
  RF(x;Z_{1},\dots,Z_{n})=(\frac {n}{s})^{-1}\sum_{1\leqslant i_{1} < i_{2} < \dots < i_{s} \leqslant n} \mathbb{E}_{\xi \thicksim \Xi}[T(x;\xi,Z_{1},\dots,Z_{i_{s}})]
\end{equation}

\subsubsection{Deep learning Methods}
Methods based on deep learning learn representations 
of input data by transforming original covariates 
or extracting features from the covariate space. 
Since the counterfactual distribution generally differs from 
the factual one, some methods aiming to predict counterfactual 
outcomes from factual data have emerged, leveraging effective 
feature representations to address domain adaptivity. 
~\cite{shalit2017estimating} lays out an intuitive generalization 
error bound that indicates the expected estimation error 
of a representation is constrained by the sum of the conventional 
generalization error of this representation and the distance 
between treatment and control distributions based on 
this representation. Minimizing the following function can help 
alleviate the problem of selection bias in Individual Treatment 
Effect (ITE) estimation.

\begin{equation}
  min_{h,\Phi} \frac{1}{n} \sum_{n = 1}^{N} r_{n} \cdot L(h(\Phi (x_{n}),W_{n}),y_{n})+\lambda \cdot R(h)+\alpha \cdot IPM_{G}(\{\Phi (x_{n})\}_{n:W_{n}=0}, \{\Phi (x_{n})\}_{n:W_{n}=1})
\end{equation}

\cite{yao2019ace,yao2019estimation} proposed a method 
for estimating Similarity-preserved Individual Treatment Effects 
(SITE) based on deep representation learning. 
This method leverages local similarity information 
to constrain the estimation process, thereby reducing bias. 
Other representation learning methods based on matching have 
also been suggested.~\cite{chang2017informative} estimates 
the counterfactual outcomes of treated samples by learning a 
projection matrix, which maximizes the nonlinear dependence 
between the subspace of control samples and outcome variables. 
\cite{chu2020matching} put forth a method combining deep 
representation learning and match-based Feature Selection 
for Representation Matching (FSRM), which maps the original 
covariate space to a selective, nonlinear, and balanced 
representation space. The IPM term is defined as 

\begin{equation}
  W_{ass}(P,Q)=\gamma \inf_{k \in \mathcal{K}} \int_{\Phi(x)} || k(\Phi(x))-\Phi(x)||P(\Phi(x))d(\Phi(x)) 
\end{equation}
to balance the representation, and then match in the representation space from learning. 

In addition, causal inference can also be conceptualized 
as a multi-task learning problem where the treatment and 
control groups share a set of layers, as well as having their 
respective specific layers. In such multi-task learning scenarios, 
the impact of selection bias can be mitigated through 
the propensity-dropout regularization approach~\cite{alaa2017deep}, 
or by employing DRNet~\cite{schwab2020learning} to learn the 
individual effects of various treatments characterized by 
dosage parameters.

\subsubsection{Meta-learning based Methods} 
In the previously mentioned methods, all strive to provide 
precise estimates of the Conditional Average Treatment Effect 
(CATE), while controlling for confounders. On the other hand, 
meta-learning based methods divide this process 
into two steps, first estimating the conditional average outcome, 
then deriving the CATE estimates. Entity T-learner
~\cite{kunzel2019metalearners}, for example, employs two trees 
to estimate the conditional treatment/control outcomes, 
with its estimator being the S-learner learned
${\hat{\tau}}_S(x)=\hat{\mu}(x,1)-\hat{\mu}(x,0)$. 
When the number of units is extremely unbalanced between two groups, 
the performance of the base models trained on the smaller group 
is poor. To overcome this issue, X-learner calculates the 
difference between observed outcomes and estimated outcomes 
as the estimated treatment effect, yielding the outcomes 
for the control group and the treatment group as 
${\hat{D}}_i^C={\hat{\mu}}_1(x)-Y^F,\ \ {\hat{D}}_i^T=Y^F-{\hat{\mu}}_0(x)$, respectively. 
R-learner, based on Robinson's transformation, proposes a new 
CATE estimator's loss function to enhance estimation performance: 

\begin{equation}
  \tau(\cdot)=\arg \min_{\tau} \{\frac{1}{n} \sum_{n = 1}^{N} ((Y^{F}_{n}-\hat{m}(x_{n}))-(w_{n}-\hat{e}(x_{n}))\tau(x_{n}))^{2}+\Lambda (\tau (\cdot))\}
\end{equation}

\section{The Tasks and Exsiting Work about Semi-Definite Data Paradigm}

In this section, we present an overview of the tasks and 
existing work involved in the semi-definite data paradigm 
to demonstrate the research progress made on two types of data: 
single-structure $\&$ multi-value and multi-structure $\&$ single-value data.

\begin{itemize}

\item For the multi-structure $\&$ single-value data type, 
the focus is primarily on time series data. 
The objective of such tasks is to identify causal relationships 
between multiple temporal components 
(where the causal structure may vary across different samples). 
Following the classification approach of Review~\cite{gong2023causal}, 
these tasks are further divided into multivariate time series 
and event sequence, depending on the presence of calibrated data. 

\item Regarding the single-structure $\&$ multi-value data type, it 
encompasses various tasks related to different 
multi-value data modalities, such as image-related tasks, 
text-related tasks, speech-related tasks, 
and representation-related tasks. These tasks have 
distinct high-level fields, involving recognition, 
classification, generation, extraction, and discrimination. 
However, the common field is to recover a fixed 
(potentially containing only essential parts) 
causal model among the multi-value data at a lower level.

\end{itemize}

\subsection{Multi-structure $\&$ Single-value Data}

The causal relationship within $d$ variables in a time series $\{x^{t}\}_{t\in \mathbb{Z}^{+}}=\{(x^{t}_{1},x^{t}_{2},\dots,x^{t}_{d})^{T}\}_{t\in \mathbb{Z}^{+}}$ 
is determined by structural equation $x^{t}_{i}:=f_{i}(Pa(x^{t}_{i}),u^{t}_{i})$, where$i=1,\dots,d$. 
Its output are constantly summary causal graph or window causal graph. 
Common methods in definite data often include constraint-based, 
score-based, and SCM-based approaches, each of which also 
has many applicable extensions within the field of time series data. 
Most methods generally fit new causal models whenever they 
encounter a new structure. However, approaches based on SCM 
consider harnessing the conformity between autoregression 
and autoencoders to learn the dynamics of multiple structures.

\subsubsection{Constraint-based methods} 
TE$_{M}$~\cite{zhou2022causality} and PCGCE~\cite{sun2015causal} 
are two such extensions of the constraint-based method, 
expanded through the utilization of transfer entropy. 
Meanwhile, PCMCI~\cite{runge2019detecting} is also an expansion 
of the constraint-based approach which tests 
$X^{t-k}_{i} \not \! \perp \!\!\! \perp X^{t}_{j}|Pa(x^{t}_{j}\{X^{t-k}_{i}\},X^{t-k}_{i})$ 
by the conditional interaction of the parent nodes of $x^{t}_{j}$ 
and $x^{t-k}_{i}$. PCMCI+~\cite{runge2020discovering} 
further extends its application 
to the identification of momentary causal relationships. 

\subsubsection{FCI-based methods} 
Furthermore, there are methodologies that stem from FCI-based 
expansions. For instance, ANLTSM~\cite{chu2008search} incorporates 
an additive noise model, assuming the effects of 
latent confounding factors to be both linear and synchronous, 
as indicated below: 
$x^{t}_{j}=\sum_{1\leqslant i \leqslant d, i\neq j} a_{j,i}x^{t}_{i}+\sum_{1\leqslant i \leqslant d, 1\leqslant l \leqslant \tau }f_{j,i,l}(x^{t-l}_{i})+\sum_{r=1}^{h}b_{j,r}u^{t}_{r}+e^{t}$
which estimates the conditional expectation $\mathbb{E}(x_i^t|x_j^t\bigcup S)$ 
to subsequently investigate the significance of causal relationships. 
TsFCI~\cite{entner2010causal} transforms the original time series 
using a fixed sliding window into a derivative vector, 
wherein each component is an independent random variable. 

\subsubsection{Score-based} 
Similar to the determination methodology, the approaches can be 
categorized into combinatorial optimization and continuous 
optimization. In the combinatorial optimization, 
~\cite{friedman2013learning} is the first to propose the structured 
expectation-maximization algorithm to learn DBN from 
longitudinal data. Subsequently,~\cite{pena2005learning} 
utilized k-fold cross-validation to compute the score 
$\frac{1}{T}\sum_{k=1}^{K}\log{p(D^K|G,{\hat{\theta}}^k)}$ 
in DBN for $D^K$. Using a decomposable score function,~\cite{de2011efficient} 
transformed the structure learning problem in DBN into 
the correspondingly augmented BN, ensuring global optimization 
using a branch and bound method. The causal graph is then 
searched through optimization: 
$(\mathcal{G}^{0_{*}}, \mathcal{G}^{1_{*}})=\arg \max_{\mathcal{G}^{0}, \mathcal{G}^{1}}(s_{D^{0}}(\mathcal{G}^{0})+S_{D^{1:T}}(\mathcal{G}^{1}))$.
For continuous optimization methods, inspired by NOTEARS~\cite{zheng2018dags}, 
DYNOTEARS~\cite{pamfil2020dynotears} models time series data as 
$x^{t}=x^{t}W+x^{t-1}A^{1}+\dots+x^{t-p}A^{p}+u^{t}$ and 
determines linear causal relationships through the constraint 
$\min_{W,A} F(W,A)=f(W,A)+\frac{\rho }{2}h(W)^{2}+\alpha h(W)$. 
The recent NTS-NOTEARS~\cite{sun2021nts} simultaneously 
captures both linear and non-linear relationships 
between variables, accomplished through the implementation 
of CNN for $\mathbb{E}[x_j^t| P a(x_j^t)]={\rm CNN}_j(\left\{x^{t-k}:1\le k\le K\right\},\ x_{_j}^t)$.

\subsubsection{SCM-based methods} 
The primary focus of the research is independent additive noise. 
VAR-LiNGAM~\cite{hyvarinen2010estimation} utilizes 
non-Gaussianity to estimate the structural autoregressive model, 
defining $x^t=\sum_{k=0}^{\tau}{B^kx^{t-k}+u^t}=\sum_{k=0}^{\tau}{M^kx^{t-k}+n^t}$, 
and estimating momentary causal effects as ${\hat{B}}^k=(I-{\hat{B}}^0){\hat{M}}^k$. 
Its extended method~\cite{lanne2017identification} 
leverages the variational distribution of latent variables 
for automatic modelling of multi-structure causal models.
Escaping the constraints of linearity and additivity, 
NCDH~\cite{wu2022nonlinear} is employed to extract general 
non-linear relationships. It hypothesizes that observed values 
are generated by mutually independent latent variables: 
$=f(S)\ \ where\ f={(f_1,\ f_2,\ldots,f_d)}^T\ and\ \ S={(S_1,\ S_2,\ldots,S_d)}^T$. 
TiMINo~\cite{peters2013causal} defines time series as 
$x^{t}_{j}=f_{j}(Pa(x^{\tau}_{j})^{t-\tau}),\dots,Pa(x^{1}_{j})^{t-1},Pa(x^{0}_{j})^{t},u^{t}_{j}$, 
with its output being a summarized time graph or an 
indeterminate state. 

\subsubsection{Granger Causality} 
Furthermore, research concerning Granger causality~\cite{granger1969investigating} 
showcases a shift in the study of multi-structure data, 
transitioning from multiple static models serving single structures 
to one dynamic models sustaining multiple structures simultaneously. 

Early work focused on how to integrate Granger causality 
with time series, thus simplifying multi-structure data into 
various single structure causal models. 
Radial basis functions~\cite{ancona2004radial} extended Granger causality 
into binary nonlinear scenarios, while~\cite{marinazzo2008kernel}, 
drawing from the theory of reproducing kernel Hilbert spaces (RKHS), 
proposed a Granger causal analysis model that embedded data 
into space for nonlinear search. \cite{sindhwani2012scalable} 
introduced a matrix-valued extension of kernel methods and 
applied it to a vector-valued RKHS dictionary. 

Recent studies have been integrating Granger causality 
and deep learning, viewing the causal relationships 
of multiple structures as a determined distribution. 
The Minimum Prediction Information Regularization (MPIR) method
~\cite{wu2020discovering} employs the learnable corruption 
of predictive variables, minimizes the risk of mutual information 
regularization, and compensates for inefficiency 
and instability. It provides information of $x^{t}_{j}$ as 
little as possible for each $x^{t-K:t-1}_{i}$: 

\begin{equation}
  R_{X,x_{j}}[f_{\theta,n}]=E_{X^{t-1},x^{t}_{j},u}[(x^{t}_{j}-f_{\theta}(\tilde{X}^{t-K:t-1}_{(n)} ))^{2}]+\lambda \cdot \sum_{p = 1}^{d} I(\tilde{X}^{t-K:t-1}_{i(n)};\tilde{X}^{t-K:t-1}_{i})  
\end{equation}

There's also~\cite{tank2017interpretable} based on the linear 
VAR model, which defines the generation process for each variable 
and Granger causality respectively as: 

\begin{equation}
  x^{t}_{j}:=g_{j}(x^{1:(t-1)}_{1},\dots,x^{1:(t-1)}_{i},\dots,x^{1:(t-1)}_{d})+u^{t}_{j}, for 1\leqslant j\leqslant d 
\end{equation}

\begin{equation}
  g_{j}(x^{1:(t-1)}_{1},\dots,x^{1:(t-1)}_{i},\dots,x^{1:(t-1)}_{d})=g_{j}(x^{1:(t-1)}_{1},\dots,x^{1:(t-1)}_{i^{'}},\dots,x^{1:(t-1)}_{d})
\end{equation}

ACD~\cite{lowe2022amortized} uses predictions for $x^{t+1}$ 
to optimize the ELBO: 

\begin{equation}
  L(X_{train},\phi,\theta)=\sum_{s= 1}^{S} \sum_{t = 1}^{T} loss(x^{t+1}_{s},f_{\theta}(x^{\leqslant t}_{s},f_{\phi}(x_{s})))+r(f_{\phi}(x_{s}))   
\end{equation}

\subsection{Single-structure $\&$ Multi-value Data}
single-structure $\&$ multi-value Data focuses on 
transforming modal information (such as images, text, audio, etc.) 
into computable deep representations (matrices, embeddings, 
spectrograms). Therefore, the structure of such data is fixed, 
allowing for the determination of correlation between two 
causal representations using metrics like cosine similarity, which 
aims to orientate the causal directions. 

\subsubsection{Image} 
In tasks related to images,~\cite{lopez2017discovering} initially 
proposed the presence of causal tendencies 
in image signals and defined features in the image as 
object features and context features. They determine 
the causal features ($X\rightarrow Y$) and the counter-causal 
features ($X\leftarrow Y$). The discerning method of 
the Neural Causation Coefficient (NCC) resemble to a 
fusion of ANM and neural networks. 
For the features $x_{ij}$ and $y_{ij}$, they define the 
feedforward neural network as: 

\begin{equation}
  NCC(\{(x_{ij},y_{ij})\}^{m_{i}}_{j=1})=\psi (\frac{1}{m_{i}}\sum_{j = 1}^{m_{i}}\phi(x_{ij},y_{ij}))
\end{equation}
which orientates the causal direction via bi-classifer. 
The goal of CAPNET~\cite{oh2021causal} is to leverage past facial 
images to fulfill emotional predictions. Its causal relation 
extractor consists of an LSTM layer and two FC layers, 
executing causal inference on emotional states by training 
$p(y_t)=p(y_t|x_{t-(d+\frac{n}{f})},\ n=f\times(w-d),\ldots,2s,s,0)$. 
Similarly, ~\cite{shadaydeh2021analyzing} employs hypothesis 
testing to assist past facial expressions in emotional causality 
inference, constructing different directional autoregressive 
models for the extracted image features and labels respectively: 

\begin{equation}
  X_{t}=\sum_{j = 1}^{M}a_{j}X_{t-j}+  \sum_{j = 1}^{M}b_{j}Y_{t-j}+\varepsilon_{t}
\end{equation}

\begin{equation}
  Y_{t}=\sum_{j = 1}^{M}c_{j}X_{t-j}+  \sum_{j = 1}^{M}d_{j}Y_{t-j}+\vartheta_{t} 
\end{equation}

The causal direction can be determined via conducting F tests for 
parameters in the model.~\cite{mao2022causal} defines the SCM: 
$<V=\{X,Y\},U=\{U_{X},U_{XY}\},\mathcal{F}=\{f_{X},f_{Y}\},P(U)>$ 
and orientates causal direction according to two equations: 
$X \leftarrow f_{X}(U_{X},U_{XY})$ and $Y \leftarrow f_{Y}(X,U_{XY})$. 

\subsubsection{Text}  
Causality extraction methods based on text can be subdivided 
into knowledge-based, statistical machine learning-based, 
and deep learning-based strategies. 

In knowledge-based methods, for explicit causal relationships 
within sentences, \cite{khoo2000extracting} offers an approach 
for exploring causality in medical text corpora by 
treating specific words as causal clues based on domain knowledge. 
WordNet~\cite{beamer2008automatic} captures noun features 
to extract causality according to the similarity pairing 
principle of noun pairs. ~\cite{girju2009classification} eschews 
dependency on manual construction, employing clustering algorithms 
to automatically establish a matching pattern. For implicit 
causal relationships, ~\cite{beltagy2019scibert} utilizes 
weak supervision to iterate over the identification of 
verb patterns, inseparable causal patterns, 
and nonverbal patterns. 

In contrast, statistical machine learning-based methods 
require fewer manually predefined patterns. For explicit causal 
relationships, ~\cite{girju2003automatic} proposes a model 
to detect causal relationships in QA systems, verifying whether 
the verbs representing causal morality entail causality via 
a feature set of decision trees. ~\cite{zhao2016event} 
utilizes Restricted Hidden 
Naive Bayes (RHNB) to handle interaction between features, 
causal conjunctions and lexical syntactic patterns. 
For implicit causal relationships. 

Deep learning-based methods effectively mitigate 
the problem of feature sparsity. In explicit causality, 
\cite{xu2015classifying} employs LSTM to learn higher-level 
semantics and \cite{li2021causality} amalgamates 
BiLSTM with Multi-Head Self-Attention (MHSA) 
to direct attention towards remote dependencies between words, 
~\cite{chen2016implicit} combines BiLSTM with GRU to 
capture complex semantic interactions between text segments. 
For implicit causality, ~\cite{chicco2020advantages} 
employs Reinforcement Learning (RL) to relabel noisy instances, 
then uses PCNN to iteratively retrain the relationship extractor 
with adjusted labels. ~\cite{zhao2021modeling} 
leverages the Cross-Modal Attention Network (CMAN), 
built by stacking two attention units: the BiLSTM-enhanced 
Self-Attention (BSA) unit and BiLSTM-enhanced Label-Attention 
(BLA) unit, to achieve dense correlation in the tagging and 
label space. 

Recently,~\cite{chen2023affective,chen2023learninga} 
have unified models of explicit and implicit causes, 
defining implicit causes as $i.i.d.$ noise terms, 
while defining explicit causes as known contexts. 
The causality identifiability of their models has been 
theoretically substantiated. Additionally, they have converted 
the Structural Causal Models (SCM) into autoencoders, 
demonstrating that it is possible to achieve effective 
causal representations even when the implicit causes are unknown, 
provided that the textual prior is combined as structural knowledge: 

\begin{equation} 
  E=f_{2}((I-A)f_{1}(H))
  \label{eqt4}
 \end{equation}
 \begin{equation} 
  H=f_{4}((I-A)^{-1}f_{3}(E))
  \label{eqt5}
 \end{equation}

 \subsubsection{Audio} 
\cite{hazan2007causal} employs a causality unsupervised structure 
to study the structure of audio streams and make predictions, 
computing ZCR, SC, MFCC, and Pitch respectively for event encoding. 
Given a previous context sequence $e_{(i-n)+1}^{i-1}$, 
they utilize the PPM algorithm~\cite{cleary1984comparison} 
to predict the next symbol $e_{i}$: 

\begin{equation}
  \begin{split}
    p(e_{i}|e^{i-1}_{(i-n)+1})=\left\{
      \begin{array}{lr}
        \alpha(e_{i}|e^{i-1}_{(i-n)+1}), & if c(e_{i}|e^{i-1}_{(i-n)+1})>0\\
        p(e_{i}|e^{i-1}_{(i-n)+2}) \gamma(e^{i-1}_{(i-n)+1}), &if c(e_{i}|e^{i-1}_{(i-n)+1})=0
      \end{array}
    \right.
  \end{split}
  \label{eqnsims}
\end{equation}

WaveNet[20] is likewise employed for audio sequence predictions, 
leveraging the network stacked with dilated causal convolution 
layers and GRU to train the conditional distribution of 
the audio to identify causal relationships within the sequence: 

\begin{equation}
  p(x|h)=\prod^{T}_{t=1}p(x_{t}|x_{1},\dots,x_{t-1},h)
\end{equation} 

\subsubsection{Deep Representation} 

\begin{wrapfigure}{r}{0.5\textwidth}
  \centering
  \subfigure[DAG $\mathcal{G}$]{
    \includegraphics[width=0.35\textwidth]{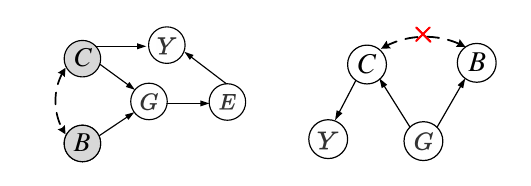}}
  \subfigure[DAG $\mathcal{H}$]{
    \includegraphics[width=0.12\textwidth]{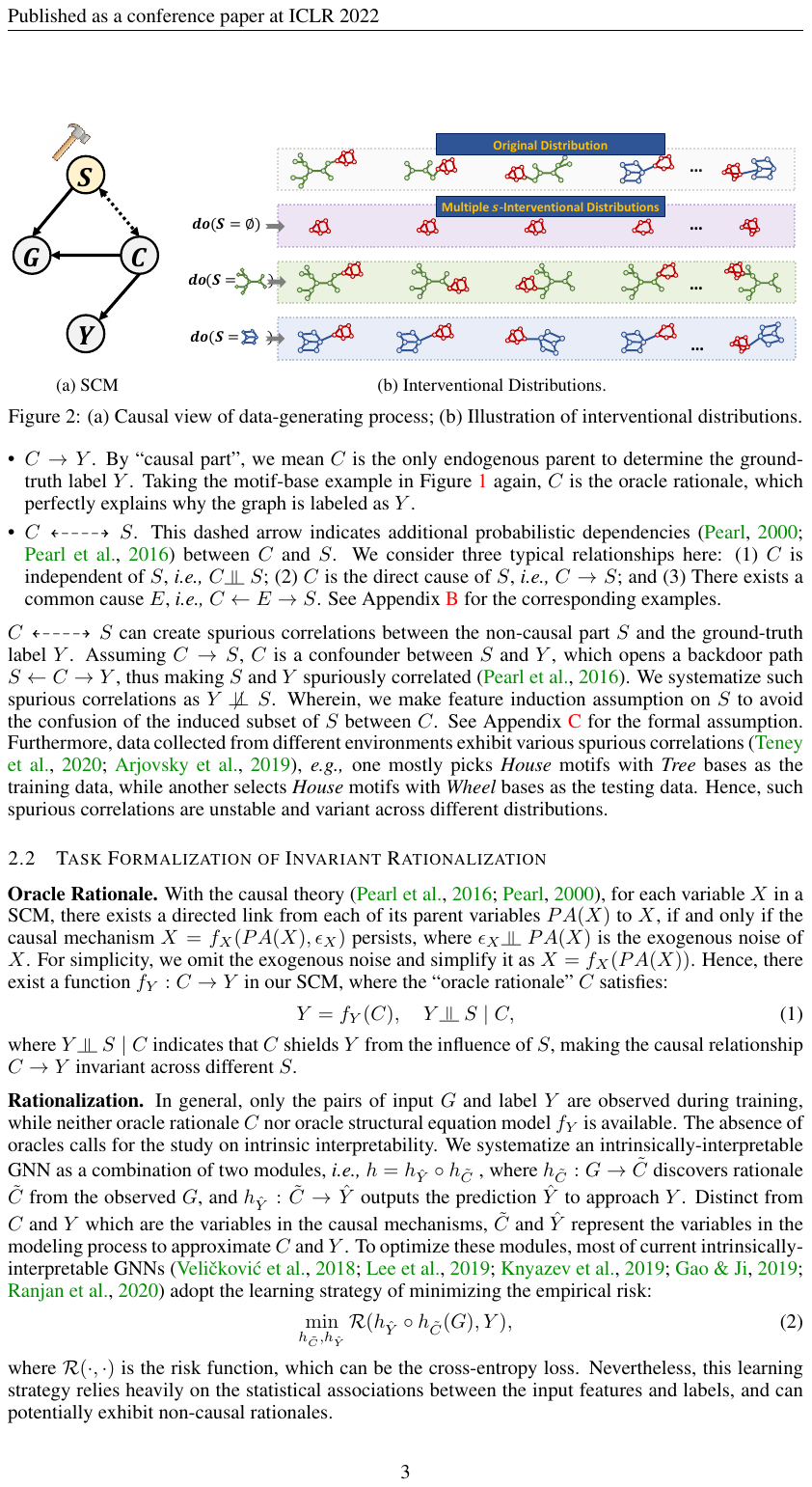}}
  \caption{two causal models driven from~\cite{fan2022debiasing,wu2021discovering}}
  \label{figpatterns}
\end{wrapfigure}

Recently, some studies~\cite{feng2021degree,fan2022debiasing,wu2021discovering} 
have conducted direct analysis 
on the causal relationships between deep representations. 
Drawing inspiration from domain generalization, 
these works divide deep representations into two distinct patterns: 
causal patterns and shortcut patterns. 
The causality graph constructed by this approach is as shown in Figure~\ref{figpatterns}. 

Within their approach, intervention is defined similarly 
to the creation of negative samples in self-supervised methods. 
The intervention is achieved by swapping the causal patterns 
and shortcut patterns of different samples, with the aim of 
predicting the correct causal pattern:
 
\begin{equation}
  \min \mathcal{R}_{DIR}=\mathbb{E}_{s}[\mathcal{R}(h(G),Y|do(S=s))]+\lambda Var_{s}(\{\mathcal{R}(h(G),Y|do(S=s))\})
\end{equation} 

\section{The Challenges and Roadmaps of Indefinite Data Paradigms}
Despite the challenges posed by the combination of multiple 
structures and multi-value variables, we aim to explore 
the two perspectives of both multiple structures  
and multi-valued variables, separately. 
In other words, when discussing the issues arising 
from multiple structures, it is assumed that multi-value data 
leads to the existence of $p: \mathcal{S} \rightarrow \mathcal{X}$. 
$q: \mathcal{X} \rightarrow \mathcal{\hat{X}}$ 
and similarly, multi-value data assumes that variable 
$f$ cannot be solved through statistical strength.

\subsection{Multi-structure Perspectives}

As mentioned in Section~\ref{secdd}, 
the issue of multiple structures in the indefinite data paradigm 
has surpassed the fundamental challenges of clustering ability 
of samples and the dynamic capability to learn from 
different structures. Due to the disturbance caused by 
quantification inaccuracies, the learning ability of 
multi-structure models cannot be independent of the 
robustness and accuracy of deep learning. 
However, the success of deep learning techniques is 
primarily attributed to large-scale $i.i.d.$ datasets, 
which endow them with exceptional generalization capabilities 
on the same distribution (i.e., from one datapoint to another). 
However, their generalization abilities lag significantly 
on different distributions (i.e., from one dataset to another).

The relevant studies in Paper~\cite{chen2023affective} 
elucidated this situation by evaluating the performance 
of various methods for emotion inference in dialogues. 
These methods include task-specific supervised approaches 
and unsupervised approaches primarily based on 
large language models (LLMs) and pre-trained models. 
Table~\ref{tabr1} presents the specific results, 
involving two specifically designed causal relation 
testing experiments: 

\begin{wraptable}{r}{0.4\textwidth}
  \caption{two challenges of causation examination~\cite{chen2023affective}}
  \label{tab:freq}
  \resizebox{0.4\textwidth}{!}{
  \begin{tabular}{clllll}
    \toprule
    \multirow{2}{*}{Methods} & \multicolumn{2}{c}{Challenge 1} &  \multicolumn{3}{c}{Challenge 2} \\
    \cline{2-6}
    &($U_{a}$,$U_{b}$)&($U_{b}$,$U_{a}$)&($U_{a}$,$U_{b}$)&($U_{b}$,$U_{c}$)&($U_{a}$,$U_{c}$)\\
    \midrule
     GPT-3.5~\cite{radford2019language}&112 & 102& 108&114 & 109\\
     GPT-4~\cite{openai2023gpt4}&127 &114 &111 & 105& 103\\
     \midrule
    RoBERTa~\cite{liu2019roberta}&95 & 97&94 &91 & 83\\
    RoBERTa$^{+}$~\cite{liu2019roberta}&97 &91& 105& 101&106\\
    \midrule
    RANK-CP~\cite{wei-etal-2020-effective}&142 & 125& 147& 129& 131\\
     ECPE-2D~\cite{ding-etal-2020-ecpe}&151 &153 &142 &138 & 146\\
     EGAT~\cite{chen2022learning}& 166& 154&157 & 139&148 \\
  \bottomrule
\end{tabular}}
\label{tabr1}
\end{wraptable}

\begin{itemize}
\item Negative samples that involve interchanging outcome  
utterances and cause utterances. For such negative samples, 
the ability of deep models to recognize the Markovian causal 
relationships of $X_{i} \rightarrow X_{j}$ and 
$X_{j} \rightarrow X_{i}$ can be examined.

\item Negative samples with intermediate variables. 
For such negative samples, the ability of the models to 
recognize the causal relationships of $X_{i} \rightarrow X_{j}$ and 
$X_{i} \rightarrow X_{k} \rightarrow X_{j}$ can be examined.

\end{itemize}

However, all the models did not exhibit significant causality 
discrimination in these two tests. Specifically, 
irrespective of positive or negative samples, 
the models learned similar associations. 
As mentioned in Section~\ref{secp}, 
existing deep model methods can only learn the existence 
of associations between two variables but are unable to discern 
the specific direction, such as 
$X_{i} \rightarrow X_{j}$ and $X_{j} \rightarrow X_{i}$. 
This lack of causality discrimination is a noteworthy limitation. 

\begin{wraptable}{r}{0.5\textwidth}
  \caption{Overall performance of CORR2CAUSE~\cite{jin2023large}}
  \label{tab:freq}
  \resizebox{0.5\textwidth}{!}{
  \begin{tabular}{cllll}
    \toprule
    &F1&Precision&Recall&Accuracy\\
    \midrule
    \textbf{Random Baselines}& & &\\
    Always Majority &0.0&0.0&0.0&84.77\\
    Random (Proportional)&13.5&12.53&14.62&71.46\\
    Random (Uniform)&20.38&15.11&31.29&62.78\\
    \midrule 
    \textbf{BERT-Based Models}& & &\\
    BERT MNLI~\cite{devlin2019bert}&2.82&7.23&1.75&81.61\\
    RoBERTa MNLI~\cite{liu2019roberta}&22.79&34.73&16.96&82.50\\
    DeBERTa MNLI~\cite{he2020deberta}&14.52&14.71&14.33&74.31\\
    DistilBERT MNLI~\cite{sanh2019distilbert}&20.70&24.12&18.13&78.85\\
    DistilBAET MNLI~\cite{shleifer2020pre}&26.74&15.92&83.63&30.23\\
    BART MNLI~\cite{lewis2019bart}&33.38&31.59&35.38&78.50\\
    \midrule 
    \textbf{LLaMa-Based Models}& & & \\
    LLaMa-6.7B~\cite{touvron2023llama}&26.81&15.50&99.42&17.36\\
    Alpaca-6.7B\cite{taori2023stanford}&27.37&15.93&97.37&21.33\\
    \midrule 
    \textbf{GPT-Based Models}& & &\\
    GPT-3 Ada~\cite{brown2020language}&0.00&0.00&0.00&84.77\\
    GPT-3 Babbage~\cite{brown2020language}&27.45&15.96&97.95&21.15\\
    GPT-3 Curie~\cite{brown2020language}&26.43&15.23&100.00&15.23\\
    GPT-3 Davinci~\cite{brown2020language}&27.82&16.57&86.55&31.61\\
    GPT-3 Instruct (text-davinci-001)~\cite{ouyang2022training}&17.99&11.84&37.43&48.04\\
    GPT-3 Instruct (text-davinci-002)~\cite{ouyang2022training}&21.87&13.46&58.19&36.69\\
    GPT-3 Instruct (text-davinci-003)~\cite{ouyang2022training}&15.72&13.4&19.01&68.97\\
    GPT-3.5~\cite{radford2019language}&21.69&17.79&27.78&69.46\\
    GPT-4~\cite{openai2023gpt4}&29.08&20.92&47.66&64.60\\
  \bottomrule
\end{tabular}}
\label{tabr2}
\end{wraptable}

Coincidentally, Paper~\cite{jin2023large} 
introduces a dataset called CORR2CAUSE, 
which is designed to assess the causal discrimination ability 
of LLMs. The dataset focuses on when causal inferences 
can be made based on correlation and when they cannot. 
Table~\ref{tabr2} presents a more comprehensive evaluation of LLMs' 
causality discrimination capability, 
further affirming that pure causal reasoning remains a 
highly challenging task for all current LLMs. 

Fortunately, SCMs provide evidence for analyzing more precise 
causal relationships in the case of fitting two similar variables. 
Under the assumption of causal sufficiency, the common cause 
theorem implies that the noise terms must be independent. 
This independence introduces a discrepancy between 
the dependencies of $P(X_{i}|pa_{i})$ and $P(X_{j}|Pa_{j})$ 
with respect to the correlation between $X_{i}$ and $X_{j}$ 
(note that the independence of $U_{i}$ cannot be directly 
translated into the independence of $X_{i}$). 
In other words, when fitting $X_{i}$ and $X_{j}$, 
in addition to capturing their correlation, 
it becomes possible to uncover the causal relationship 
between $X_{i}$ and $X_{j}$. Following the setup in Paper~\cite{chen2023affective}, 
let's assume $\Sigma$ represents the residual after 
fitting variables $X_{i}$ and $X_{j}$. In this case, 
the conditional independence between the residual term 
and the variables is given by: 

\begin{definition}[Independent Conditions for causations from correlations]
  The relationship of two utterances 
  $X_{i}$ and $X_{j}$ in a dialogue 
is causal discriminable, from the independent conditions: 
\begin{itemize}
  \item $\Sigma_{X_{i}} \perp \!\!\! \perp X_{j}, 
  \Sigma_{X_{j}} \not \! \perp \!\!\! \perp X_{i}
  \Rightarrow X_{j} \rightarrow X_{i}$
  \item $\Sigma_{X_{i}} \not \! \perp \!\!\! \perp X_{j}, 
  \Sigma_{X_{j}} \perp \!\!\! \perp X_{i}
  \Rightarrow X_{i} \rightarrow X_{j}$
  \item $\Sigma_{X_{i}} \not \! \perp \!\!\! \perp X_{j}, 
  \Sigma_{X_{j}} \not \! \perp \!\!\! \perp X_{i}
  \Rightarrow L \rightarrow X_{i}, L \rightarrow X_{j}$
  \item $\Sigma_{X_{i}} \perp \!\!\! \perp X_{j}, 
  \Sigma_{X_{j}} \perp \!\!\! \perp X_{i}
  \Rightarrow X_{i} \rightarrow L, X_{j} \rightarrow L$
\end{itemize}
\label{deficcfc}
\end{definition}

Definition~\ref{deficcfc} indicates that for any two variables 
that can be fitted, incorporating independent noise terms 
allows for the inclusion of knowledge from correlation to causation. 
Therefore, the focus of causal discrimination lies in how 
to incorporate the noise terms. Paper~\cite{chen2023learning} 
summarizes the application of variational inference in 
previous causal discovery. Specifically, treating the noise terms 
as latent variables $Z$ and computing the reconstruction loss 
of causal representation has emerged as a popular framework. 
The log-evidence of this framework is expressed as:

\begin{equation}
  \frac{1}{N} \sum_{n=1}^{N} \log p(X_{n})=\frac{1}{N} \sum_{n=1}^{N} \log\int p(X_{n}|E)p(E)dE
  \label{eqt8}
  \end{equation}

\subsection{Multi-value Perspectives}

The issues arising from multi-value data in the 
indefinite data paradigm still revolve around quantification 
inaccuracies and statistical feature deviations. 
However, considering the impact of multiple structures, 
we concretize the problems into two aspects: 
1) How to calculate the influence of latent confounders 
in the context of an unfixed causal model? 
2) How to ensure the causal consistency between the 
causal representation and the causal model? 

For 1), numerous studies~\cite{miao2018identifying,zhu2022mitigating,kallus2019interval,shimizu2014bayesian} 
have proposed approaches 
to handle confounders when it is not possible to make simplifying 
assumptions on the causal model. Instead of pinpointing 
the confounders, these approaches focus on disentangling the 
unattributed portion of the observed variables. 
Such methods represent $X$ as $\hat{X} + C$, 
where $C$ is represented in the form of a SCM:

\begin{align} 
  &c_{m,j}=\sum_{l_{m,k}\in Ec(x_{m,j})} g_{m,kj} l_{m,k} (c \in \mathbb{R}^{N \times D}, l\in \mathbb{R}^{K \times D}, 0\leq i \neq j,k < N)
  \label{eqt24}
 \end{align}

\begin{figure}
  \centering
  \subfigure[DAG $\mathcal{G}$]{
    \includegraphics[width=0.18\textwidth]{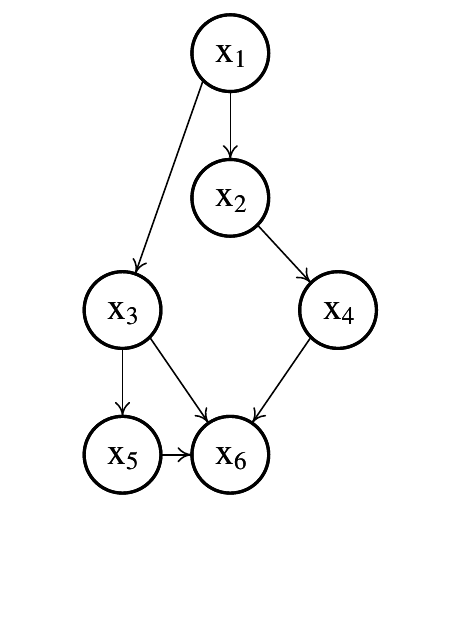}}
  \subfigure[DAG $\mathcal{H}$]{
    \includegraphics[width=0.18\textwidth]{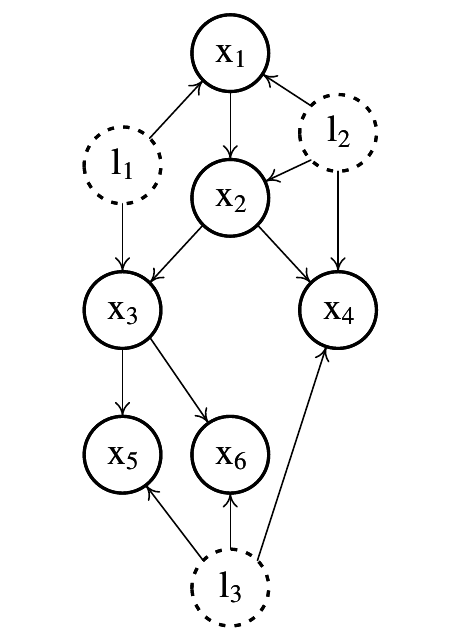}}
  \subfigure[DAG $\mathcal{O}$]{
    \includegraphics[width=0.18\textwidth]{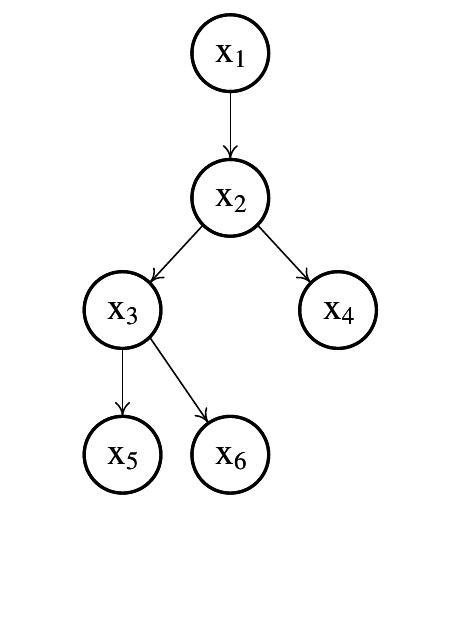}}
  \subfigure[DAG $\mathcal{C}$]{
    \includegraphics[width=0.18\textwidth]{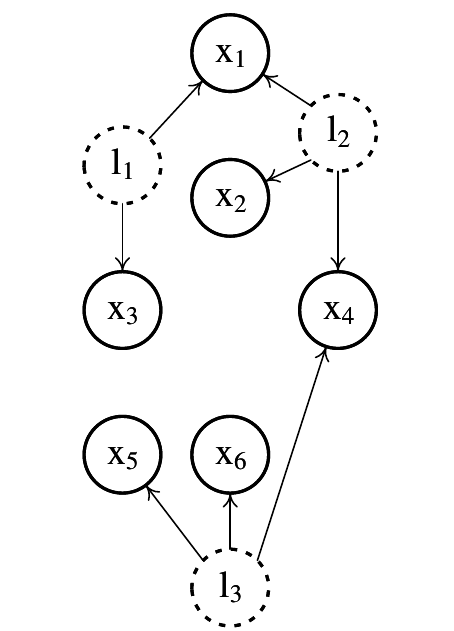}}
  \caption{Four causal DAGs discussed in~\cite{chen2023learning}.}
  \label{figdisentanglement}
\end{figure}

According to the setting of SCMs, the observed target $X_{i}$ 
consists of an endogenous term ($f_{i}Pa_{i}$) 
and two exogenous terms ($g_{i}l_{m} and U_{i}$). 
Its causal model can be decoupled into two subgraphs, 
representing the pure relationships between $X$ variables 
($f_{i}Pa_{i}+U_{i}$) and the remaining exogenous term 
coefficients ($g_{i}l_{m}$), as shown in Figure~\ref{figdisentanglement}.
~\cite{chen2023learning} analyzed the distributions 
of the causal model under different states: 
complete state, confounding-free state, 
pure relationship state among $X$ variables, 
and confounding effect state. In the case of 
single-value data, the confounding impact can be assumed to 
follow the distribution $P(X|L)$. However, this advantage 
does not hold for multi-value data. Hence, 
~\cite{chen2023learning} further proposes an approximate estimation  
for the confounding effect. 

\begin{definition}
  there exist an approximate estimation 
  \begin{equation}
    C_{j}=\frac{\mathbb{P}(x_{j}) \mathbb{P}(L|x_{j})}{\sum_{i}^{N} \mathbb{P}(x_{i}) \mathbb{P}(L|x_{i}) } x_{j} 
    \label{eqt26}
\end{equation}
  when the latent variables $l$ vastly outweigh independent noise $\epsilon_{x}$. 
  \label{def5}
\end{definition}

Regarding 2), we have addressed this issue in Section~\ref{secdd}. 
Specifically, assuming the parameters of the encoder as $\varphi$ 
and the parameters of the decoder as $\theta$, 
we can infer that in the context of indefinite data paradigms, 
the causal model can be expressed as $\mathcal{G} = F(X, \varphi)$, 
while for causal representation, it can be written as 
$Y = H(X, a, b)$. This leads to the indefinite data paradigm 
being the only data type where inconsistencies between 
the causal model and causal representation may exist 
(in single-structure models, the causal model $\mathcal{G}$ is fixed, 
and in single-value variables, $\varphi$ and $\theta$ do not exist). 
Furthermore, considering that most existing encoder-decoder structures
~\cite{rezende2015variational,wijmans1995solution,zhang2018advances,creswell2018generative}
lack constraints on $\theta$ w.r.t. $\varphi$ 
(i.e., $\theta = \omega(\varphi)$), the causal representation 
and causal model generated by the indefinite data paradigm 
often exhibit inconsistencies, resulting in less intelligent 
behavior when it comes to out-of-distribution generalization. 

Intervention (i.e., $do$ operator) is a well-recognized approach 
for testing causal consistency, but it is ineffective 
under the indefinite data paradigm when it comes to computing 
adjustment $Z$ formulas or controlling variable distributions. 
However, in essence, the indefinite data 
can achieve consistency with intervention by constraining the 
inputs of deep models to control the set of parent nodes involved 
in the computation, thus ensuring that the parent node set 
of the target variable $X$ is an empty set. Additionally, 
for the problem of consistency between two causal models, 
works such as exact transformation and causal abstraction 
provide a theoretical basis based on the intervention set
~\cite{rubenstein2017causal,geiger2021causal,beckers2019abstracting}.

\section{Datasets from Three Data Paradigms}

\subsection{Datasets for Definite Data}

The definite datasets are characterized by single-value 
and single-structure features, often manifested in the form of 
structured data. Each variable consists of single valued 
attributes, such as body temperature, age, wind strength, 
economic growth rate, population count, protein quantity, and more. 
Furthermore, all the data points are obtained from sampling 
conducted under the same $i.i.d.$ environment, 
suggesting that all samples mirror the same causal model. 
Table~\ref{tabdd} lists numerous popular definite datasets along with 
brief summaries of their characteristics. 

\begin{table}
  \caption{Definite Datasets}
  \resizebox{1\linewidth}{!}{
  \begin{tabular}{ccc}
    \toprule
    Datasets&$\#variables$&Highlights\\
    \midrule
    Erdos-Rényi Graphs & Customization & Number of probabilities based on custom 
    nodes $p = \frac{2e}{d^2}$ adding edges\\
    Twins~\cite{almond2005costs} &Customization&twin births in the United
    States between 1989-1991\\
    SynTReN~\citep{van2006syntren} & Customization & Synthetic gene expression data \\
    Census Income KDD~\citep{ristanoski2013discrimination} & 3 & The Census Income (KDD) 
    data set (US)\\
    ADNI~\cite{jack2008alzheimer}&3&three latent representation outputs about Alzheimer’s Disease\\
    CPRD~\cite{herrett2015data}&4&medical records from NHS general practice clinics in the UK\\
    Economic data~\citep{iacoviello2019foreign} & 4 & Quarterly data for the United States 1965 - 2017\\
    Arrhythmia~\citep{asuncion2007uci} & 4 & from the UCI Machine Learning 
    Repository concerns cardiac arrhythmia \\
    DWD climate~\citep{dietrich2019temporal} & 6 & Data from 349 weather stations in Germany\\
    IHDP~\cite{brooks1992effects}& 8 &originated from preterm infants with low birth weight\\
    Archaeology~\citep{huang2018generalized} & 8 & Archaeological data\\
    Soil Properties~\citep{solly2014factors} & 8 & Biological root decomposition data \\
    AutoMPG~\citep{asuncion2007uci} & 9 & Fuel consumption data for city car cycles\\
    Abalone~\citep{nash1994population} & 9 & Conch abalone size data\\
    HK Stock Market~\citep{huang2020causal} & 10 & Dividend-adjusted daily closing prices of 10 
    major stocks \\
    Sachs~\citep{sachs2005causal} & 11 & Proteins and phospholipids in human cells\\
    Breast Cancer Wisconsim~\citep{asuncion2007uci} & 11 & Breast cancer cell data\\
    TCGA~\cite{weinstein2013cancer}&12&a comprehensive and
    extensive database of genomic information\\
    CHESS~\cite{qian2020between}&14&ICU patients during the first wave of the COVID-19\\
    CogUSA study~\citep{tu2019causal} & 16 & Cognitive ageing data\\
    Jobs~\cite{lalonde1986evaluating} & 16 & assess the effect of vocational training on employment outcomes\\
    fMRI~\citep{wang2003training} & 25 & The different voxel data in fMRI were organised into ROIs\\
    
  \bottomrule
\end{tabular}}
\label{tabdd}
\end{table}

The definite datasets mainly originates from various 
real-world domains, with the most frequently represented being 
the fields of medicine and disease
~\cite{almond2005costs,van2006syntren,herrett2015data,jack2008alzheimer,
asuncion2007uci,sachs2005causal,asuncion2007uci,weinstein2013cancer,tu2019causal,wang2003training,qian2020between}, 
followed by finance
~\cite{ristanoski2013discrimination,iacoviello2019foreign,asuncion2007uci,
huang2020causal,lalonde1986evaluating} and nature
~\cite{dietrich2019temporal,brooks1992effects,huang2018generalized,solly2014factors,
nash1994population}. 
Such datasets typically exhibit two characteristics: 
1) a small number of observed variables, and 
2) a large sample size. When combined with the attributes 
of single-structure and single-value data, these datasets 
often facilitate the identification of quite accurate 
causal structures. Furthermore, they reveal the real-world 
meanings of these causal relations. 

A further consideration arises from realities of data collection, 
where errors and issues can result in incomplete data. 
This situation has fueled the emergence of causal completion tasks 
that aim at imputing missing values based on the observed 
causal associations. Another phenomenon borne out by the 
limited number of observed variables accounting for the 
whole system is the confounding bias problem. 
It has thus prompted corresponding research on confounding factors. 

With these datasets, researchers hardly need to worry about 
the structure of the causal relationship. 
A fully connected undirected graph, built upon the provided 
observed variables of the dataset, can serve to establish a 
complete causal structure. The primary focus of the statistical 
study then becomes how to utilize this data to derive 
the causal relation between pairs of variables. 
That is, determining which undirected edges should be 
retained and subsequently directed. 
This process embodies the cores of their domain-specific research.

\subsection{Datasets for Semi-definite Data}

Semi-definite datasets can be broadly categorized into two types: 
multi-structure $\&$ single-value data, and 
single-structure $\&$ multi-value data. The former type 
essentially only includes temporal datasets 
such as financial time series or blood oxygen level time series. 
Meanwhile, the latter type traverses a majority of 
disparate modalities including but not limited to image datasets 
and text datasets. To provide a grip on these diverse dataset types, 
we have articulated their respective examples and highlights  
in Table~\ref{tabds}.

\begin{table}
  \caption{Semi-Definite Datasets}
  \resizebox{1\linewidth}{!}{
  \begin{tabular}{ccc}
    \toprule
    Datasets&Types&Highlights\\
    \midrule
    CMU MoCap~\cite{CMUMoCap}&time series &containing data about joint angles, body position\\
    DREAM-3~\cite{prill2010towards}&time series &continuous gene expression and regulation dynamics\\
    BOLD~\cite{BOLD}&time series &different brain regions of interest in human subjects\\
    MemeTracker~\cite{MemeTracker}&time series &online articles’ website, publication time, and all the hyperlinks\\
    IPTV~\cite{luo2014you}&time series &the user’s viewing behavior, i.e., what program and when
    they watch in the IPTV systems\\
    G-7~\cite{demirer2018estimating}&time series &daily return volatility of sovereign bonds of countries in the
    Group of Seven\\
    \midrule
    CIFAR-10/100~\cite{krizhevsky2009learning}&image&each respectively containing fine-grained and coarse-grained labels representing the image.\\
    ImageNet~\cite{deng2009imagenet}&image&More than 20,000 categories of image recognition data\\
    ObjectNet~\citep{barbu2019objectnet} &iamge&objects are shown in different camera angles in a cluttered room\\
    YFCC-100M Flickr~\citep{karkkainen2021fairface}&image&Facial recognition data\\
    Casual Conversations~\citep{hazirbas2021towards}&image& Facial recognition dataset that can include age, gender, and skin color as classification tags.\\
    CUB200-2011~\citep{wah2011caltech}&image&Includes identification data for 200 different bird species.\\
    Aff-wild2~\citep{kollias2018aff} &image&a wide variation of subjects in terms of posture, age, lighting conditions, race and occupation\\
    Market1501~\citep{zheng2015person}&image&Pedestrian re-identification dataset\\
    Cityscapes~\citep{hu2020probabilistic}&image&Street scene recognition data from 50 different cities.\\
    \midrule
    Amazon reviews~\citep{mudambi2010research}&text&Amazon.com product reviews related to books, electronics, clothing and other 26 categories\\
    OntoNotes$5$~\citep{dror2017replicability}&text&Large news corpus, including 7 types\\
    CFPB compliant~\citep{tan2022causal}&text&U.S. Consumer Financial Protection Bureau (CFPB) Consumer Complaint Data\\
    SNIPS~\citep{larson2019evaluation} &text&The personal voice assistant collects data containing 7 domains\\
    CrisistextLine~\citep{gould2022crisis}&text&Session data of counselor and mentee\\
    Yelp~\cite{luca2016reviews}&text&Merchant review site user review data in Yelp\\
    CNC~\cite{tan2022causal}&text&Social News Data Corpus\\
    AG'news~\cite{zhang2015character}&text&Classified coverage from over 2,000 news sources\\
    Gender Citation Gap~\cite{maliniak2013gender}&text&Data from the TRIP database obtained from the IR literature\\
    Weiboscope~\citep{fu2013assessing}&text&Real-time posting data of Weibo social media users\\
    JDDC~\citep{chen2019jddc} &text&D customer service staff on the topic of after-sales conversations\\
    \midrule
    Spurious-Motif~\cite{ying2019gnnexplainer}&graph&Each graph is composed of one base and one motif\\
    MNIST-75sp~\cite{knyazev2019understanding}&graph&converts the MNIST images into 70000 superpixel graphs with at most 75 nodes each graph\\
    Graph-SST2~\cite{socher2013recursive}&graph&Each graph is labeled by its sentence sentiment \\
    Molhiv~\cite{wu2018moleculenet}&graph& nodes are atoms, and edges are chemical bonds\\
    BA-Shapes~\cite{barabasi1999emergence}&graph&80 five-node motifs are randomly attached to the base graph\\
    MUTAG~\cite{debnath1991structure}&graph&Every graph is labeled according to their mutagenic effect on the bacterium\\
    OGBG-Mol~\cite{wu2018moleculenet}&graph&Graphs represent molecules and labels are chemical properties of molecules\\
    \bottomrule
\end{tabular}}
\label{tabds}
\end{table}

Compared to defined datasets, semi-definite datasets span 
a more diverse range of areas, primarily due to the expansion 
from single-value data to multi-value data. 
This shift has enabled image, text, and graph data to 
be incorporated into causal discovery, and the development 
of multi-structure datasets introduces the novel data type 
of time series. For instance, ImageNet and Cifar-10 image datasets 
were not initially intended for causal discovery. 
However, with the convergence of causal discovery and 
deep learning, explaining the classification and identification 
of an entire image based on the semantics of its components  
has become a popular choice. The same pattern can be found 
in text datasets. For example, the CFPB compliant dataset 
contains records of consumer complaints and whether 
they were resolved timely, which can be used to build 
semi-definite causal models to identify the features 
of complaints that can be resolved. Graph datasets are most 
invloved with interpretability of GNN via analysing the causal 
relationships between shortcuts and causal features. 
Consequently, 
we did not list the number of causal variables, 
as depending on the high-level task, causal variables 
could represent a subset or a mapping of the observed variables. 
Similarly, we did not tackle the the number of causal structures  
in time-series datasets, as the number of structures is 
indeterminate depending on the causal variables involved. 

\subsection{Datasets for Indefinite Data}

Indefinite datasets contain multi-structure samples, 
and every sample incorporates multi-value variables. 
From a machine learning data resource perspective, 
indefinite datasets possess characteristics of both types 
of semi-definite datasets. In other words, 
they are more likely to be temporal multi-value datasets, 
such as dialog and video datasets. 

However, it's important to note that temporality is not a defined 
property of indefinite datasets. The association exists 
largely because, in the real world, most multi-structure data 
naturally occur as temporal sequences. Their variable lengths 
and partial non-replicability contribute to the multi-structure  
data type. But it should be stressed that multi-structure 
does not inherently imply temporality. 
(While this kind of raw data is challenging to collect 
in the real world, we can imagine its existence. 
For example, by disrupting and randomly reordering 
the sequence of interactions within a dialog dataset, 
the resulting dataset loses its temporal quality.) 
The same notion is also applicable to multi-structure datasets 
within the semi-definite data. 

When viewed from a causal discovery perspective, 
indefinite datasets share significant intersections with 
semi-definite datasets. Many datasets, while originally 
developed for semi-definite tasks, could transition to 
indefinite datasets by relaxing the constraints on structure 
or values. However, these indefinite tasks are often ill-posed, 
requiring additional prior knowledge or assumptions 
for their resolution. 

\begin{table}
  \caption{Datasets which can be regarded as indefinite datasets\protect\footnotemark[2]}
  \resizebox{1\linewidth}{!}{
  \begin{tabular}{cccc}
    \toprule
    Datasets&Types&Original tasks&Highlights\\
    \midrule
    DailyDialog~\cite{li2017dailydialog}&text&ER,AR,ECG&A daily conversation dataset annotated with 7 emotions, 4 actions and 10 themes\\
    CORR2CAUSE~\cite{jin2023large}&text&CI,CoT&A text datasets for large language model with causal labels\\
    MultiArith~\cite{roy2016solving}&text&QA,AR,CoT&an arithmetic datasets with multiple steps and operations\\
    CSQA~\cite{talmor2018commonsenseqa}&text&QA,CR,CoT&A commonsense datasets answering with piror knowledge\\
    Coin Flip~\cite{wei2022chain}&text&QA,SR,CoT&A symbolic reasoning datasets which can be used in large language model\\
    CAER~\cite{lee2019context}&video&ER,CI&a drama video datasets with 7 emotion labels\\
    50salads~\cite{stein2013combining}&video&AR,CI&a video datasets composed of action parts with 17 action labels\\
    Bach Chorales dataset~\citep{roman2019holistic}&audio&MG,PG&Bach’s four-part congregational hymn dataset\\
    MedleyDB~\citep{bittner2014medleydb} &audio&MG,PG&Polyphonic Pop Music Dataset\\
    IEMOCAP~\cite{busso2008iemocap}&multimodal&ERC,NLU,CI&A conversation script for 10 male and female actors with emotion labels\\
    
    \bottomrule
  \end{tabular}}
\label{tabdi}
\end{table}
\footnotetext[2]{ER is Emotional Recognition, 
~\citep{zhou2018emotional}, NLU is Natural Language Understanding
~\citep{liu2019multi}, AR is Action Recognition
~\citep{poppe2010survey}, ECG is Emotional Conversation Generation
~\citep{sun2018emotional}, ODGG is Open-Domain Dialogue Generation
~\citep{kann2022open}, ERG is Empathetic Response Generation
~\citep{majumder2020mime},CI is Causal Inference~\cite{jin2023large}, 
QA is Question Answering, AR is Arithmetic Reasoning~\cite{roy2016solving}, 
CoT is Chain of Thinking~\cite{zhang2022automatic}, 
CR is Commonsense Reasoning~\cite{talmor2018commonsenseqa}, 
SR is Symbolic Reasoning~\cite{wei2022chain}, 
MG is Melodic generation and PG is Polyphonic generation. }

Table~\ref{tabdi} showcases a collection of potential 
indefinite datasets we've compiled. While these datasets 
involve diverse modalities and tasks, under certain conditions, 
they all fulfill the characteristics of multi-structure and 
multi-value variables, with causal relations between 
variables being particularly significant. For instance, 
in the DailyDialog dataset, some studies have annotated the 
causes of emotions~\cite{poria2021recognizing,chen2023affective}, 
leading to the recognition of spans of causes. 
In this type of task, each dialog represents a unique causal 
structure, whereas each utterance acts as a multi-value variable. 
The 50salads dataset, meanwhile, investigates task segmentation 
in videos via automated action recognition. Each video 
is treated as a distinct causal model with various segments 
acting as multi-value variables. Understanding the relations  
between these segments through causal relationships 
among the multi-value variables is a crucial point of this task. 
Datasets associated with the CoT task explore the 
causal relationships between each step. Audio modality datasets 
also focus on the mutual influences between different 
polyphonic vioces. 

\section{Conclusion}

In this paper, we redefine existing causal data from two perspectives, 
causal structure and causal representation, to allow it to be 
embedded appropriately within the theories and methods of 
deep learning, forming an throughout and comprehensive domain. 

Firstly, we redefine fundamental concepts such as causal model, 
causal variable, and causal representation, introducing the 
differences in existing data regarding structure 
and representation. Specifically, we use the term 
``multi-structure data'' to denote cases where the causal 
structure is not unique, and ``single-structure data'' 
for those with a fixed causal structure. We employ 
``multi-value variables'' to refer to data that requires 
deep representation for computations in causal variables, 
and ``single-value variables'' to encompass statistical data 
that exists in numerical form without the need for 
deep representation. Based on structural and representational 
differences, we define three different data paradigms: 
definite data (single-structure and single-value), 
semi-definite data (single-structure and multi-value, 
or multi-structure and single-value), and indefinite data 
(multi-structure and multi-value). We exemplify their 
differences by their forms, analyze the distinct problems 
they face in terms of resolution pathways, and summarize 
their respective development directions in the evolution of researches.

Definite and semi-definite data, as mature research domains, 
have been discussed in relation to various tasks 
to highlight their current application scenarios. 
Definite data primarily involves causal discovery, 
causal discovery with latent confounders, 
and causal effect estimation. Semi-definite data 
mainly concerns causal data related to time series, 
images, text, other modalities, and deep representations. 
However, indefinite data is still in its infancy. 
We consider structure and representation to envision roadmaps 
for situations dealing with simultaneous 
multi-structure and multi-value data based on existing 
research problems. 

Lastly, we organize datasets commonly used in the 
three data paradigms, briefly introduce their attributes, 
and summarize their application fields. 

The three data paradigms proposed encompass 
nearly all causal models, especially upon including 
unstructured inputs such as images and texts. 
We hope these three data paradigms can provide readers 
with broader causal insights. When confronted with specific 
causal data, they can follow the categorization in this review 
to capture the key features and challenges of the data type, 
thereby identifying the baselines of deep models 
and causal theories.

\bibliographystyle{ACM-Reference-Format}
\bibliography{sample-base}


\begin{thebibliography}{292}


\ifx \showCODEN    \undefined \def \showCODEN     #1{\unskip}     \fi
\ifx \showDOI      \undefined \def \showDOI       #1{#1}\fi
\ifx \showISBNx    \undefined \def \showISBNx     #1{\unskip}     \fi
\ifx \showISBNxiii \undefined \def \showISBNxiii  #1{\unskip}     \fi
\ifx \showISSN     \undefined \def \showISSN      #1{\unskip}     \fi
\ifx \showLCCN     \undefined \def \showLCCN      #1{\unskip}     \fi
\ifx \shownote     \undefined \def \shownote      #1{#1}          \fi
\ifx \showarticletitle \undefined \def \showarticletitle #1{#1}   \fi
\ifx \showURL      \undefined \def \showURL       {\relax}        \fi
\providecommand\bibfield[2]{#2}
\providecommand\bibinfo[2]{#2}
\providecommand\natexlab[1]{#1}
\providecommand\showeprint[2][]{arXiv:#2}

\bibitem[\protect\citeauthoryear{Agrawal, Squires, Prasad, and Uhler}{Agrawal
  et~al\mbox{.}}{2021}]%
        {agrawal2021decamfounder}
\bibfield{author}{\bibinfo{person}{Raj Agrawal}, \bibinfo{person}{Chandler
  Squires}, \bibinfo{person}{Neha Prasad}, {and} \bibinfo{person}{Caroline
  Uhler}.} \bibinfo{year}{2021}\natexlab{}.
\newblock \showarticletitle{The DeCAMFounder: Non-linear causal discovery in
  the presence of hidden variables}.
\newblock \bibinfo{journal}{\emph{arXiv preprint arXiv:2102.07921}}
  (\bibinfo{year}{2021}).
\newblock


\bibitem[\protect\citeauthoryear{Ahuja, Hartford, and Bengio}{Ahuja
  et~al\mbox{.}}{2022}]%
        {ahuja2022weakly}
\bibfield{author}{\bibinfo{person}{Kartik Ahuja}, \bibinfo{person}{Jason~S
  Hartford}, {and} \bibinfo{person}{Yoshua Bengio}.}
  \bibinfo{year}{2022}\natexlab{}.
\newblock \showarticletitle{Weakly supervised representation learning with
  sparse perturbations}.
\newblock \bibinfo{journal}{\emph{Advances in Neural Information Processing
  Systems}}  \bibinfo{volume}{35} (\bibinfo{year}{2022}),
  \bibinfo{pages}{15516--15528}.
\newblock


\bibitem[\protect\citeauthoryear{Akbari, Mokhtarian, Ghassami, and
  Kiyavash}{Akbari et~al\mbox{.}}{2021}]%
        {akbari2021recursive}
\bibfield{author}{\bibinfo{person}{Sina Akbari}, \bibinfo{person}{Ehsan
  Mokhtarian}, \bibinfo{person}{AmirEmad Ghassami}, {and}
  \bibinfo{person}{Negar Kiyavash}.} \bibinfo{year}{2021}\natexlab{}.
\newblock \showarticletitle{Recursive causal structure learning in the presence
  of latent variables and selection bias}.
\newblock \bibinfo{journal}{\emph{Advances in Neural Information Processing
  Systems}}  \bibinfo{volume}{34} (\bibinfo{year}{2021}),
  \bibinfo{pages}{10119--10130}.
\newblock


\bibitem[\protect\citeauthoryear{Alaa, Weisz, and Van Der~Schaar}{Alaa
  et~al\mbox{.}}{2017}]%
        {alaa2017deep}
\bibfield{author}{\bibinfo{person}{Ahmed~M Alaa}, \bibinfo{person}{Michael
  Weisz}, {and} \bibinfo{person}{Mihaela Van Der~Schaar}.}
  \bibinfo{year}{2017}\natexlab{}.
\newblock \showarticletitle{Deep counterfactual networks with
  propensity-dropout}.
\newblock \bibinfo{journal}{\emph{arXiv preprint arXiv:1706.05966}}
  (\bibinfo{year}{2017}).
\newblock


\bibitem[\protect\citeauthoryear{Almond, Chay, and Lee}{Almond
  et~al\mbox{.}}{2005}]%
        {almond2005costs}
\bibfield{author}{\bibinfo{person}{Douglas Almond}, \bibinfo{person}{Kenneth~Y
  Chay}, {and} \bibinfo{person}{David~S Lee}.} \bibinfo{year}{2005}\natexlab{}.
\newblock \showarticletitle{The costs of low birth weight}.
\newblock \bibinfo{journal}{\emph{The Quarterly Journal of Economics}}
  \bibinfo{volume}{120}, \bibinfo{number}{3} (\bibinfo{year}{2005}),
  \bibinfo{pages}{1031--1083}.
\newblock


\bibitem[\protect\citeauthoryear{Anandkumar, Hsu, Javanmard, and
  Kakade}{Anandkumar et~al\mbox{.}}{2013}]%
        {anandkumar2013learning}
\bibfield{author}{\bibinfo{person}{Animashree Anandkumar},
  \bibinfo{person}{Daniel Hsu}, \bibinfo{person}{Adel Javanmard}, {and}
  \bibinfo{person}{Sham Kakade}.} \bibinfo{year}{2013}\natexlab{}.
\newblock \showarticletitle{Learning linear bayesian networks with latent
  variables}. In \bibinfo{booktitle}{\emph{International Conference on Machine
  Learning}}. PMLR, \bibinfo{pages}{249--257}.
\newblock


\bibitem[\protect\citeauthoryear{Anciukevicius, Fox-Roberts, Rosten, and
  Henderson}{Anciukevicius et~al\mbox{.}}{2022}]%
        {anciukevicius2022unsupervised}
\bibfield{author}{\bibinfo{person}{Titas Anciukevicius},
  \bibinfo{person}{Patrick Fox-Roberts}, \bibinfo{person}{Edward Rosten}, {and}
  \bibinfo{person}{Paul Henderson}.} \bibinfo{year}{2022}\natexlab{}.
\newblock \showarticletitle{Unsupervised Causal Generative Understanding of
  Images}.
\newblock \bibinfo{journal}{\emph{Advances in Neural Information Processing
  Systems}}  \bibinfo{volume}{35} (\bibinfo{year}{2022}),
  \bibinfo{pages}{37037--37054}.
\newblock


\bibitem[\protect\citeauthoryear{Ancona, Marinazzo, and Stramaglia}{Ancona
  et~al\mbox{.}}{2004}]%
        {ancona2004radial}
\bibfield{author}{\bibinfo{person}{Nicola Ancona}, \bibinfo{person}{Daniele
  Marinazzo}, {and} \bibinfo{person}{Sebastiano Stramaglia}.}
  \bibinfo{year}{2004}\natexlab{}.
\newblock \showarticletitle{Radial basis function approach to nonlinear Granger
  causality of time series}.
\newblock \bibinfo{journal}{\emph{Physical Review E}} \bibinfo{volume}{70},
  \bibinfo{number}{5} (\bibinfo{year}{2004}), \bibinfo{pages}{056221}.
\newblock


\bibitem[\protect\citeauthoryear{Assaad, Devijver, and Gaussier}{Assaad
  et~al\mbox{.}}{2022}]%
        {assaad2022survey}
\bibfield{author}{\bibinfo{person}{Charles~K Assaad}, \bibinfo{person}{Emilie
  Devijver}, {and} \bibinfo{person}{Eric Gaussier}.}
  \bibinfo{year}{2022}\natexlab{}.
\newblock \showarticletitle{Survey and evaluation of causal discovery methods
  for time series}.
\newblock \bibinfo{journal}{\emph{Journal of Artificial Intelligence Research}}
   \bibinfo{volume}{73} (\bibinfo{year}{2022}), \bibinfo{pages}{767--819}.
\newblock


\bibitem[\protect\citeauthoryear{Asuncion and Newman}{Asuncion and
  Newman}{2007}]%
        {asuncion2007uci}
\bibfield{author}{\bibinfo{person}{Arthur Asuncion} {and}
  \bibinfo{person}{David Newman}.} \bibinfo{year}{2007}\natexlab{}.
\newblock \bibinfo{title}{UCI machine learning repository}.
\newblock
\newblock


\bibitem[\protect\citeauthoryear{Athey and Imbens}{Athey and Imbens}{2016}]%
        {athey2016recursive}
\bibfield{author}{\bibinfo{person}{Susan Athey} {and} \bibinfo{person}{Guido
  Imbens}.} \bibinfo{year}{2016}\natexlab{}.
\newblock \showarticletitle{Recursive partitioning for heterogeneous causal
  effects}.
\newblock \bibinfo{journal}{\emph{Proceedings of the National Academy of
  Sciences}} \bibinfo{volume}{113}, \bibinfo{number}{27}
  (\bibinfo{year}{2016}), \bibinfo{pages}{7353--7360}.
\newblock


\bibitem[\protect\citeauthoryear{Athey and Imbens}{Athey and Imbens}{2015}]%
        {athey2015machine}
\bibfield{author}{\bibinfo{person}{Susan Athey} {and} \bibinfo{person}{Guido~W
  Imbens}.} \bibinfo{year}{2015}\natexlab{}.
\newblock \showarticletitle{Machine learning methods for estimating
  heterogeneous causal effects}.
\newblock \bibinfo{journal}{\emph{stat}} \bibinfo{volume}{1050},
  \bibinfo{number}{5} (\bibinfo{year}{2015}), \bibinfo{pages}{1--26}.
\newblock


\bibitem[\protect\citeauthoryear{Badura, Madarasova~Geckova, Sigmundova,
  Sigmund, van Dijk, and Reijneveld}{Badura et~al\mbox{.}}{2018}]%
        {badura2018can}
\bibfield{author}{\bibinfo{person}{Petr Badura}, \bibinfo{person}{Andrea
  Madarasova~Geckova}, \bibinfo{person}{Dagmar Sigmundova},
  \bibinfo{person}{Erik Sigmund}, \bibinfo{person}{Jitse~P van Dijk}, {and}
  \bibinfo{person}{Sijmen~A Reijneveld}.} \bibinfo{year}{2018}\natexlab{}.
\newblock \showarticletitle{Can organized leisure-time activities buffer the
  negative outcomes of unstructured activities for adolescents’ health?}
\newblock \bibinfo{journal}{\emph{International journal of public health}}
  \bibinfo{volume}{63} (\bibinfo{year}{2018}), \bibinfo{pages}{743--751}.
\newblock


\bibitem[\protect\citeauthoryear{Balashankar and Subramanian}{Balashankar and
  Subramanian}{2021}]%
        {balashankar2021learning}
\bibfield{author}{\bibinfo{person}{Ananth Balashankar} {and}
  \bibinfo{person}{Lakshminarayanan Subramanian}.}
  \bibinfo{year}{2021}\natexlab{}.
\newblock \showarticletitle{Learning faithful representations of causal
  graphs}. In \bibinfo{booktitle}{\emph{Proceedings of the 59th Annual Meeting
  of the Association for Computational Linguistics and the 11th International
  Joint Conference on Natural Language Processing (Volume 1: Long Papers)}}.
  \bibinfo{pages}{839--850}.
\newblock


\bibitem[\protect\citeauthoryear{Barab{\'a}si and Albert}{Barab{\'a}si and
  Albert}{1999}]%
        {barabasi1999emergence}
\bibfield{author}{\bibinfo{person}{Albert-L{\'a}szl{\'o} Barab{\'a}si} {and}
  \bibinfo{person}{R{\'e}ka Albert}.} \bibinfo{year}{1999}\natexlab{}.
\newblock \showarticletitle{Emergence of scaling in random networks}.
\newblock \bibinfo{journal}{\emph{science}} \bibinfo{volume}{286},
  \bibinfo{number}{5439} (\bibinfo{year}{1999}), \bibinfo{pages}{509--512}.
\newblock


\bibitem[\protect\citeauthoryear{Barbu, Mayo, Alverio, Luo, Wang, Gutfreund,
  Tenenbaum, and Katz}{Barbu et~al\mbox{.}}{2019}]%
        {barbu2019objectnet}
\bibfield{author}{\bibinfo{person}{Andrei Barbu}, \bibinfo{person}{David Mayo},
  \bibinfo{person}{Julian Alverio}, \bibinfo{person}{William Luo},
  \bibinfo{person}{Christopher Wang}, \bibinfo{person}{Dan Gutfreund},
  \bibinfo{person}{Josh Tenenbaum}, {and} \bibinfo{person}{Boris Katz}.}
  \bibinfo{year}{2019}\natexlab{}.
\newblock \showarticletitle{Objectnet: A large-scale bias-controlled dataset
  for pushing the limits of object recognition models}.
\newblock \bibinfo{journal}{\emph{Advances in neural information processing
  systems}}  \bibinfo{volume}{32} (\bibinfo{year}{2019}).
\newblock


\bibitem[\protect\citeauthoryear{Beamer, Rozovskaya, and Girju}{Beamer
  et~al\mbox{.}}{2008}]%
        {beamer2008automatic}
\bibfield{author}{\bibinfo{person}{Brandon Beamer}, \bibinfo{person}{Alla
  Rozovskaya}, {and} \bibinfo{person}{Roxana Girju}.}
  \bibinfo{year}{2008}\natexlab{}.
\newblock \showarticletitle{Automatic Semantic Relation Extraction with
  Multiple Boundary Generation.}. In \bibinfo{booktitle}{\emph{AAAI}}.
  \bibinfo{pages}{824--829}.
\newblock


\bibitem[\protect\citeauthoryear{Beckers and Halpern}{Beckers and
  Halpern}{2019}]%
        {beckers2019abstracting}
\bibfield{author}{\bibinfo{person}{Sander Beckers} {and}
  \bibinfo{person}{Joseph~Y Halpern}.} \bibinfo{year}{2019}\natexlab{}.
\newblock \showarticletitle{Abstracting causal models}. In
  \bibinfo{booktitle}{\emph{Proceedings of the aaai conference on artificial
  intelligence}}, Vol.~\bibinfo{volume}{33}. \bibinfo{pages}{2678--2685}.
\newblock


\bibitem[\protect\citeauthoryear{Belloni, Chernozhukov, Fern{\'a}ndez-Val, and
  Hansen}{Belloni et~al\mbox{.}}{2017}]%
        {belloni2017program}
\bibfield{author}{\bibinfo{person}{Alexandre Belloni}, \bibinfo{person}{Victor
  Chernozhukov}, \bibinfo{person}{Iv{\'a}n Fern{\'a}ndez-Val}, {and}
  \bibinfo{person}{Christian Hansen}.} \bibinfo{year}{2017}\natexlab{}.
\newblock \showarticletitle{Program evaluation and causal inference with
  high-dimensional data}.
\newblock \bibinfo{journal}{\emph{Econometrica}} \bibinfo{volume}{85},
  \bibinfo{number}{1} (\bibinfo{year}{2017}), \bibinfo{pages}{233--298}.
\newblock


\bibitem[\protect\citeauthoryear{Beltagy, Lo, and Cohan}{Beltagy
  et~al\mbox{.}}{2019}]%
        {beltagy2019scibert}
\bibfield{author}{\bibinfo{person}{Iz Beltagy}, \bibinfo{person}{Kyle Lo},
  {and} \bibinfo{person}{Arman Cohan}.} \bibinfo{year}{2019}\natexlab{}.
\newblock \showarticletitle{SciBERT: A pretrained language model for scientific
  text}.
\newblock \bibinfo{journal}{\emph{arXiv preprint arXiv:1903.10676}}
  (\bibinfo{year}{2019}).
\newblock


\bibitem[\protect\citeauthoryear{Bernstein, Saeed, Squires, and
  Uhler}{Bernstein et~al\mbox{.}}{2020}]%
        {bernstein2020ordering}
\bibfield{author}{\bibinfo{person}{Daniel Bernstein}, \bibinfo{person}{Basil
  Saeed}, \bibinfo{person}{Chandler Squires}, {and} \bibinfo{person}{Caroline
  Uhler}.} \bibinfo{year}{2020}\natexlab{}.
\newblock \showarticletitle{Ordering-based causal structure learning in the
  presence of latent variables}. In \bibinfo{booktitle}{\emph{International
  Conference on Artificial Intelligence and Statistics}}. PMLR,
  \bibinfo{pages}{4098--4108}.
\newblock


\bibitem[\protect\citeauthoryear{Bhattacharya, Nagarajan, Malinsky, and
  Shpitser}{Bhattacharya et~al\mbox{.}}{2021}]%
        {bhattacharya2021differentiable}
\bibfield{author}{\bibinfo{person}{Rohit Bhattacharya}, \bibinfo{person}{Tushar
  Nagarajan}, \bibinfo{person}{Daniel Malinsky}, {and} \bibinfo{person}{Ilya
  Shpitser}.} \bibinfo{year}{2021}\natexlab{}.
\newblock \showarticletitle{Differentiable causal discovery under unmeasured
  confounding}. In \bibinfo{booktitle}{\emph{International Conference on
  Artificial Intelligence and Statistics}}. PMLR, \bibinfo{pages}{2314--2322}.
\newblock


\bibitem[\protect\citeauthoryear{Bittner, Salamon, Tierney, Mauch, Cannam, and
  Bello}{Bittner et~al\mbox{.}}{2014}]%
        {bittner2014medleydb}
\bibfield{author}{\bibinfo{person}{Rachel~M Bittner}, \bibinfo{person}{Justin
  Salamon}, \bibinfo{person}{Mike Tierney}, \bibinfo{person}{Matthias Mauch},
  \bibinfo{person}{Chris Cannam}, {and} \bibinfo{person}{Juan~Pablo Bello}.}
  \bibinfo{year}{2014}\natexlab{}.
\newblock \showarticletitle{Medleydb: A multitrack dataset for
  annotation-intensive mir research.}. In \bibinfo{booktitle}{\emph{ISMIR}},
  Vol.~\bibinfo{volume}{14}. \bibinfo{pages}{155--160}.
\newblock


\bibitem[\protect\citeauthoryear{BOLD}{BOLD}{2012}]%
        {BOLD}
BOLD \bibinfo{year}{2012}\natexlab{}.
\newblock \bibinfo{title}{NetSim - Evaluation of Network Modelling Methods for
  FMRI}.
\newblock
\newblock
\newblock
\shownote{https://www.fmrib.ox.ac.uk/datasets/netsim/index.html}.


\bibitem[\protect\citeauthoryear{Brooks-Gunn, Liaw, and Klebanov}{Brooks-Gunn
  et~al\mbox{.}}{1992}]%
        {brooks1992effects}
\bibfield{author}{\bibinfo{person}{Jeanne Brooks-Gunn},
  \bibinfo{person}{Fong-ruey Liaw}, {and} \bibinfo{person}{Pamela~Kato
  Klebanov}.} \bibinfo{year}{1992}\natexlab{}.
\newblock \showarticletitle{Effects of early intervention on cognitive function
  of low birth weight preterm infants}.
\newblock \bibinfo{journal}{\emph{The Journal of pediatrics}}
  \bibinfo{volume}{120}, \bibinfo{number}{3} (\bibinfo{year}{1992}),
  \bibinfo{pages}{350--359}.
\newblock


\bibitem[\protect\citeauthoryear{Brown, Mann, Ryder, Subbiah, Kaplan, Dhariwal,
  Neelakantan, Shyam, Sastry, Askell, et~al\mbox{.}}{Brown
  et~al\mbox{.}}{2020}]%
        {brown2020language}
\bibfield{author}{\bibinfo{person}{Tom Brown}, \bibinfo{person}{Benjamin Mann},
  \bibinfo{person}{Nick Ryder}, \bibinfo{person}{Melanie Subbiah},
  \bibinfo{person}{Jared~D Kaplan}, \bibinfo{person}{Prafulla Dhariwal},
  \bibinfo{person}{Arvind Neelakantan}, \bibinfo{person}{Pranav Shyam},
  \bibinfo{person}{Girish Sastry}, \bibinfo{person}{Amanda Askell},
  {et~al\mbox{.}}} \bibinfo{year}{2020}\natexlab{}.
\newblock \showarticletitle{Language models are few-shot learners}.
\newblock \bibinfo{journal}{\emph{Advances in neural information processing
  systems}}  \bibinfo{volume}{33} (\bibinfo{year}{2020}),
  \bibinfo{pages}{1877--1901}.
\newblock


\bibitem[\protect\citeauthoryear{Busso, Bulut, Lee, Kazemzadeh, Mower, Kim,
  Chang, Lee, and Narayanan}{Busso et~al\mbox{.}}{2008}]%
        {busso2008iemocap}
\bibfield{author}{\bibinfo{person}{Carlos Busso}, \bibinfo{person}{Murtaza
  Bulut}, \bibinfo{person}{Chi-Chun Lee}, \bibinfo{person}{Abe Kazemzadeh},
  \bibinfo{person}{Emily Mower}, \bibinfo{person}{Samuel Kim},
  \bibinfo{person}{Jeannette~N Chang}, \bibinfo{person}{Sungbok Lee}, {and}
  \bibinfo{person}{Shrikanth~S Narayanan}.} \bibinfo{year}{2008}\natexlab{}.
\newblock \showarticletitle{IEMOCAP: Interactive emotional dyadic motion
  capture database}.
\newblock \bibinfo{journal}{\emph{Language resources and evaluation}}
  \bibinfo{volume}{42}, \bibinfo{number}{4} (\bibinfo{year}{2008}),
  \bibinfo{pages}{335--359}.
\newblock


\bibitem[\protect\citeauthoryear{Cai, Xie, Glymour, Hao, and Zhang}{Cai
  et~al\mbox{.}}{2019}]%
        {cai2019triad}
\bibfield{author}{\bibinfo{person}{Ruichu Cai}, \bibinfo{person}{Feng Xie},
  \bibinfo{person}{Clark Glymour}, \bibinfo{person}{Zhifeng Hao}, {and}
  \bibinfo{person}{Kun Zhang}.} \bibinfo{year}{2019}\natexlab{}.
\newblock \showarticletitle{Triad constraints for learning causal structure of
  latent variables}.
\newblock \bibinfo{journal}{\emph{Advances in neural information processing
  systems}}  \bibinfo{volume}{32} (\bibinfo{year}{2019}).
\newblock


\bibitem[\protect\citeauthoryear{Cai, Ye, Qiao, Fu, and Hao}{Cai
  et~al\mbox{.}}{2020}]%
        {cai2020fom}
\bibfield{author}{\bibinfo{person}{Ruichu Cai}, \bibinfo{person}{Jincheng Ye},
  \bibinfo{person}{Jie Qiao}, \bibinfo{person}{Huiyuan Fu}, {and}
  \bibinfo{person}{Zhifeng Hao}.} \bibinfo{year}{2020}\natexlab{}.
\newblock \showarticletitle{FOM: Fourth-order moment based causal direction
  identification on the heteroscedastic data}.
\newblock \bibinfo{journal}{\emph{Neural Networks}}  \bibinfo{volume}{124}
  (\bibinfo{year}{2020}), \bibinfo{pages}{193--201}.
\newblock


\bibitem[\protect\citeauthoryear{Chakraborty and Murphy}{Chakraborty and
  Murphy}{2014}]%
        {chakraborty2014dynamic}
\bibfield{author}{\bibinfo{person}{Bibhas Chakraborty} {and}
  \bibinfo{person}{Susan~A Murphy}.} \bibinfo{year}{2014}\natexlab{}.
\newblock \showarticletitle{Dynamic treatment regimes}.
\newblock \bibinfo{journal}{\emph{Annual review of statistics and its
  application}}  \bibinfo{volume}{1} (\bibinfo{year}{2014}),
  \bibinfo{pages}{447--464}.
\newblock


\bibitem[\protect\citeauthoryear{Chalupka, Eberhardt, and Perona}{Chalupka
  et~al\mbox{.}}{2017}]%
        {chalupka2017causal}
\bibfield{author}{\bibinfo{person}{Krzysztof Chalupka},
  \bibinfo{person}{Frederick Eberhardt}, {and} \bibinfo{person}{Pietro
  Perona}.} \bibinfo{year}{2017}\natexlab{}.
\newblock \showarticletitle{Causal feature learning: an overview}.
\newblock \bibinfo{journal}{\emph{Behaviormetrika}}  \bibinfo{volume}{44}
  (\bibinfo{year}{2017}), \bibinfo{pages}{137--164}.
\newblock


\bibitem[\protect\citeauthoryear{Chang and Dy}{Chang and Dy}{2017}]%
        {chang2017informative}
\bibfield{author}{\bibinfo{person}{Yale Chang} {and} \bibinfo{person}{Jennifer
  Dy}.} \bibinfo{year}{2017}\natexlab{}.
\newblock \showarticletitle{Informative subspace learning for counterfactual
  inference}. In \bibinfo{booktitle}{\emph{Proceedings of the AAAI Conference
  on Artificial Intelligence}}, Vol.~\bibinfo{volume}{31}.
\newblock


\bibitem[\protect\citeauthoryear{Chen, Liao, Luo, Zhu, and Yang}{Chen
  et~al\mbox{.}}{2023a}]%
        {chen2023learninga}
\bibfield{author}{\bibinfo{person}{Hang Chen}, \bibinfo{person}{Bingyu Liao},
  \bibinfo{person}{Jing Luo}, \bibinfo{person}{Wenjing Zhu}, {and}
  \bibinfo{person}{Xinyu Yang}.} \bibinfo{year}{2023}\natexlab{a}.
\newblock \bibinfo{title}{Learning a Structural Causal Model for Intuition
  Reasoning in Conversation}.
\newblock
\newblock
\showeprint[arxiv]{2305.17727}~[cs.CL]


\bibitem[\protect\citeauthoryear{Chen, Yang, and Li}{Chen
  et~al\mbox{.}}{2023b}]%
        {chen2022learning}
\bibfield{author}{\bibinfo{person}{Hang Chen}, \bibinfo{person}{Xinyu Yang},
  {and} \bibinfo{person}{Chenguang Li}.} \bibinfo{year}{2023}\natexlab{b}.
\newblock \bibinfo{title}{Learning a General Clause-to-Clause Relationships for
  Enhancing Emotion-Cause Pair Extraction}.
\newblock
\newblock
\showeprint[arxiv]{2208.13549}~[cs.CL]


\bibitem[\protect\citeauthoryear{Chen, Yang, Luo, and Zhu}{Chen
  et~al\mbox{.}}{2023d}]%
        {chen2023affective}
\bibfield{author}{\bibinfo{person}{Hang Chen}, \bibinfo{person}{Xinyu Yang},
  \bibinfo{person}{Jing Luo}, {and} \bibinfo{person}{Wenjing Zhu}.}
  \bibinfo{year}{2023}\natexlab{d}.
\newblock \showarticletitle{How to Enhance Causal Discrimination of Utterances:
  A Case on Affective Reasoning}.
\newblock \bibinfo{journal}{\emph{Proceedings of the 2023 Conference on
  Empirical Methods in Natural Language Processing (EMNLP)"}}
  (\bibinfo{year}{2023}).
\newblock


\bibitem[\protect\citeauthoryear{Chen, Yang, and Yang}{Chen
  et~al\mbox{.}}{2023c}]%
        {chen2023learning}
\bibfield{author}{\bibinfo{person}{Hang Chen}, \bibinfo{person}{Xinyu Yang},
  {and} \bibinfo{person}{Qing Yang}.} \bibinfo{year}{2023}\natexlab{c}.
\newblock \showarticletitle{Learning to Recover Causal Relationship from
  Indefinite Data in the Presence of Latent Confounders}.
\newblock \bibinfo{journal}{\emph{arXiv preprint arXiv:2305.02640}}
  (\bibinfo{year}{2023}).
\newblock


\bibitem[\protect\citeauthoryear{Chen, Zhang, Liu, Qiu, and Huang}{Chen
  et~al\mbox{.}}{2016}]%
        {chen2016implicit}
\bibfield{author}{\bibinfo{person}{Jifan Chen}, \bibinfo{person}{Qi Zhang},
  \bibinfo{person}{Pengfei Liu}, \bibinfo{person}{Xipeng Qiu}, {and}
  \bibinfo{person}{Xuan-Jing Huang}.} \bibinfo{year}{2016}\natexlab{}.
\newblock \showarticletitle{Implicit discourse relation detection via a deep
  architecture with gated relevance network}. In
  \bibinfo{booktitle}{\emph{Proceedings of the 54th Annual Meeting of the
  Association for Computational Linguistics (Volume 1: Long Papers)}}.
  \bibinfo{pages}{1726--1735}.
\newblock


\bibitem[\protect\citeauthoryear{Chen, Liu, Shen, Yuan, Zhou, Wu, He, and
  Zhou}{Chen et~al\mbox{.}}{2019}]%
        {chen2019jddc}
\bibfield{author}{\bibinfo{person}{Meng Chen}, \bibinfo{person}{Ruixue Liu},
  \bibinfo{person}{Lei Shen}, \bibinfo{person}{Shaozu Yuan},
  \bibinfo{person}{Jingyan Zhou}, \bibinfo{person}{Youzheng Wu},
  \bibinfo{person}{Xiaodong He}, {and} \bibinfo{person}{Bowen Zhou}.}
  \bibinfo{year}{2019}\natexlab{}.
\newblock \showarticletitle{The jddc corpus: A large-scale multi-turn chinese
  dialogue dataset for e-commerce customer service}.
\newblock \bibinfo{journal}{\emph{arXiv preprint arXiv:1911.09969}}
  (\bibinfo{year}{2019}).
\newblock


\bibitem[\protect\citeauthoryear{Chen and Chan}{Chen and Chan}{2013}]%
        {chen2013causality}
\bibfield{author}{\bibinfo{person}{Zhitang Chen} {and} \bibinfo{person}{Laiwan
  Chan}.} \bibinfo{year}{2013}\natexlab{}.
\newblock \showarticletitle{Causality in linear nongaussian acyclic models in
  the presence of latent gaussian confounders}.
\newblock \bibinfo{journal}{\emph{Neural Computation}} \bibinfo{volume}{25},
  \bibinfo{number}{6} (\bibinfo{year}{2013}), \bibinfo{pages}{1605--1641}.
\newblock


\bibitem[\protect\citeauthoryear{Chicco and Jurman}{Chicco and Jurman}{2020}]%
        {chicco2020advantages}
\bibfield{author}{\bibinfo{person}{Davide Chicco} {and}
  \bibinfo{person}{Giuseppe Jurman}.} \bibinfo{year}{2020}\natexlab{}.
\newblock \showarticletitle{The advantages of the Matthews correlation
  coefficient (MCC) over F1 score and accuracy in binary classification
  evaluation}.
\newblock \bibinfo{journal}{\emph{BMC genomics}} \bibinfo{volume}{21},
  \bibinfo{number}{1} (\bibinfo{year}{2020}), \bibinfo{pages}{1--13}.
\newblock


\bibitem[\protect\citeauthoryear{Chickering}{Chickering}{2002}]%
        {chickering2002optimal}
\bibfield{author}{\bibinfo{person}{David~Maxwell Chickering}.}
  \bibinfo{year}{2002}\natexlab{}.
\newblock \showarticletitle{Optimal structure identification with greedy
  search}.
\newblock \bibinfo{journal}{\emph{Journal of machine learning research}}
  \bibinfo{volume}{3}, \bibinfo{number}{Nov} (\bibinfo{year}{2002}),
  \bibinfo{pages}{507--554}.
\newblock


\bibitem[\protect\citeauthoryear{Chu, Glymour, and Ridgeway}{Chu
  et~al\mbox{.}}{2008}]%
        {chu2008search}
\bibfield{author}{\bibinfo{person}{Tianjiao Chu}, \bibinfo{person}{Clark
  Glymour}, {and} \bibinfo{person}{Greg Ridgeway}.}
  \bibinfo{year}{2008}\natexlab{}.
\newblock \showarticletitle{Search for Additive Nonlinear Time Series Causal
  Models.}
\newblock \bibinfo{journal}{\emph{Journal of Machine Learning Research}}
  \bibinfo{volume}{9}, \bibinfo{number}{5} (\bibinfo{year}{2008}).
\newblock


\bibitem[\protect\citeauthoryear{Chu, Rathbun, and Li}{Chu
  et~al\mbox{.}}{2020}]%
        {chu2020matching}
\bibfield{author}{\bibinfo{person}{Zhixuan Chu}, \bibinfo{person}{Stephen~L
  Rathbun}, {and} \bibinfo{person}{Sheng Li}.} \bibinfo{year}{2020}\natexlab{}.
\newblock \showarticletitle{Matching in selective and balanced representation
  space for treatment effects estimation}. In
  \bibinfo{booktitle}{\emph{Proceedings of the 29th ACM International
  Conference on Information \& Knowledge Management}}.
  \bibinfo{pages}{205--214}.
\newblock


\bibitem[\protect\citeauthoryear{Claggett and Karahanna}{Claggett and
  Karahanna}{2018}]%
        {claggett2018unpacking}
\bibfield{author}{\bibinfo{person}{Jennifer~L Claggett} {and}
  \bibinfo{person}{Elena Karahanna}.} \bibinfo{year}{2018}\natexlab{}.
\newblock \showarticletitle{Unpacking the structure of coordination mechanisms
  and the role of relational coordination in an era of digitally mediated work
  processes}.
\newblock \bibinfo{journal}{\emph{Academy of Management Review}}
  \bibinfo{volume}{43}, \bibinfo{number}{4} (\bibinfo{year}{2018}),
  \bibinfo{pages}{704--722}.
\newblock


\bibitem[\protect\citeauthoryear{Cleary and Witten}{Cleary and Witten}{1984}]%
        {cleary1984comparison}
\bibfield{author}{\bibinfo{person}{J Cleary} {and} \bibinfo{person}{I Witten}.}
  \bibinfo{year}{1984}\natexlab{}.
\newblock \showarticletitle{A comparison of enumerative and adaptive codes}.
\newblock \bibinfo{journal}{\emph{IEEE Transactions on Information Theory}}
  \bibinfo{volume}{30}, \bibinfo{number}{2} (\bibinfo{year}{1984}),
  \bibinfo{pages}{306--315}.
\newblock


\bibitem[\protect\citeauthoryear{CMUMoCap}{CMUMoCap}{2008}]%
        {CMUMoCap}
CMUMoCap \bibinfo{year}{2008}\natexlab{}.
\newblock \bibinfo{title}{CMU Graphics Lab Motion Capture Database}.
\newblock
\newblock
\newblock
\shownote{http://mocap.cs.cmu.edu/}.


\bibitem[\protect\citeauthoryear{Colombo, Maathuis, Kalisch, and
  Richardson}{Colombo et~al\mbox{.}}{2012}]%
        {colombo2012learning}
\bibfield{author}{\bibinfo{person}{Diego Colombo}, \bibinfo{person}{Marloes~H
  Maathuis}, \bibinfo{person}{Markus Kalisch}, {and} \bibinfo{person}{Thomas~S
  Richardson}.} \bibinfo{year}{2012}\natexlab{}.
\newblock \showarticletitle{Learning high-dimensional directed acyclic graphs
  with latent and selection variables}.
\newblock \bibinfo{journal}{\emph{The Annals of Statistics}}
  (\bibinfo{year}{2012}), \bibinfo{pages}{294--321}.
\newblock


\bibitem[\protect\citeauthoryear{Creswell, White, Dumoulin, Arulkumaran,
  Sengupta, and Bharath}{Creswell et~al\mbox{.}}{2018}]%
        {creswell2018generative}
\bibfield{author}{\bibinfo{person}{Antonia Creswell}, \bibinfo{person}{Tom
  White}, \bibinfo{person}{Vincent Dumoulin}, \bibinfo{person}{Kai
  Arulkumaran}, \bibinfo{person}{Biswa Sengupta}, {and} \bibinfo{person}{Anil~A
  Bharath}.} \bibinfo{year}{2018}\natexlab{}.
\newblock \showarticletitle{Generative adversarial networks: An overview}.
\newblock \bibinfo{journal}{\emph{IEEE signal processing magazine}}
  \bibinfo{volume}{35}, \bibinfo{number}{1} (\bibinfo{year}{2018}),
  \bibinfo{pages}{53--65}.
\newblock


\bibitem[\protect\citeauthoryear{Dai, Korb, Wallace, and Wu}{Dai
  et~al\mbox{.}}{1997}]%
        {dai1997study}
\bibfield{author}{\bibinfo{person}{Honghua Dai}, \bibinfo{person}{Kevin~B
  Korb}, \bibinfo{person}{Chris~S Wallace}, {and} \bibinfo{person}{Xindong
  Wu}.} \bibinfo{year}{1997}\natexlab{}.
\newblock \showarticletitle{A Study of Causal Discovery With Weak Links and
  Small Samples.}. In \bibinfo{booktitle}{\emph{IJCAI}}. Citeseer,
  \bibinfo{pages}{1304--1309}.
\newblock


\bibitem[\protect\citeauthoryear{De~Campos and Ji}{De~Campos and Ji}{2011}]%
        {de2011efficient}
\bibfield{author}{\bibinfo{person}{Cassio~P De~Campos} {and}
  \bibinfo{person}{Qiang Ji}.} \bibinfo{year}{2011}\natexlab{}.
\newblock \showarticletitle{Efficient structure learning of Bayesian networks
  using constraints}.
\newblock \bibinfo{journal}{\emph{The Journal of Machine Learning Research}}
  \bibinfo{volume}{12} (\bibinfo{year}{2011}), \bibinfo{pages}{663--689}.
\newblock


\bibitem[\protect\citeauthoryear{Debnath, Lopez~de Compadre, Debnath,
  Shusterman, and Hansch}{Debnath et~al\mbox{.}}{1991}]%
        {debnath1991structure}
\bibfield{author}{\bibinfo{person}{Asim~Kumar Debnath}, \bibinfo{person}{Rosa~L
  Lopez~de Compadre}, \bibinfo{person}{Gargi Debnath}, \bibinfo{person}{Alan~J
  Shusterman}, {and} \bibinfo{person}{Corwin Hansch}.}
  \bibinfo{year}{1991}\natexlab{}.
\newblock \showarticletitle{Structure-activity relationship of mutagenic
  aromatic and heteroaromatic nitro compounds. correlation with molecular
  orbital energies and hydrophobicity}.
\newblock \bibinfo{journal}{\emph{Journal of medicinal chemistry}}
  \bibinfo{volume}{34}, \bibinfo{number}{2} (\bibinfo{year}{1991}),
  \bibinfo{pages}{786--797}.
\newblock


\bibitem[\protect\citeauthoryear{Demirer, Diebold, Liu, and Yilmaz}{Demirer
  et~al\mbox{.}}{2018}]%
        {demirer2018estimating}
\bibfield{author}{\bibinfo{person}{Mert Demirer}, \bibinfo{person}{Francis~X
  Diebold}, \bibinfo{person}{Laura Liu}, {and} \bibinfo{person}{Kamil Yilmaz}.}
  \bibinfo{year}{2018}\natexlab{}.
\newblock \showarticletitle{Estimating global bank network connectedness}.
\newblock \bibinfo{journal}{\emph{Journal of Applied Econometrics}}
  \bibinfo{volume}{33}, \bibinfo{number}{1} (\bibinfo{year}{2018}),
  \bibinfo{pages}{1--15}.
\newblock


\bibitem[\protect\citeauthoryear{Deng, Dong, Socher, Li, Li, and Fei-Fei}{Deng
  et~al\mbox{.}}{2009}]%
        {deng2009imagenet}
\bibfield{author}{\bibinfo{person}{Jia Deng}, \bibinfo{person}{Wei Dong},
  \bibinfo{person}{Richard Socher}, \bibinfo{person}{Li-Jia Li},
  \bibinfo{person}{Kai Li}, {and} \bibinfo{person}{Li Fei-Fei}.}
  \bibinfo{year}{2009}\natexlab{}.
\newblock \showarticletitle{Imagenet: A large-scale hierarchical image
  database}. In \bibinfo{booktitle}{\emph{2009 IEEE conference on computer
  vision and pattern recognition}}. Ieee, \bibinfo{pages}{248--255}.
\newblock


\bibitem[\protect\citeauthoryear{Deng, Zheng, Tian, and Zeng}{Deng
  et~al\mbox{.}}{2022}]%
        {deng2022deep}
\bibfield{author}{\bibinfo{person}{Zizhen Deng}, \bibinfo{person}{Xiaolong
  Zheng}, \bibinfo{person}{Hu Tian}, {and} \bibinfo{person}{Daniel~Dajun
  Zeng}.} \bibinfo{year}{2022}\natexlab{}.
\newblock \showarticletitle{Deep Causal Learning: Representation, Discovery and
  Inference}.
\newblock \bibinfo{journal}{\emph{arXiv preprint arXiv:2211.03374}}
  (\bibinfo{year}{2022}).
\newblock


\bibitem[\protect\citeauthoryear{Devlin, Chang, Lee, and Toutanova}{Devlin
  et~al\mbox{.}}{2019}]%
        {devlin2019bert}
\bibfield{author}{\bibinfo{person}{Jacob Devlin}, \bibinfo{person}{Ming-Wei
  Chang}, \bibinfo{person}{Kenton Lee}, {and} \bibinfo{person}{Kristina
  Toutanova}.} \bibinfo{year}{2019}\natexlab{}.
\newblock \bibinfo{title}{BERT: Pre-training of Deep Bidirectional Transformers
  for Language Understanding}.
\newblock
\newblock
\showeprint[arxiv]{1810.04805}~[cs.CL]


\bibitem[\protect\citeauthoryear{Dietrich, Wolf, Kawohl, Wehberg, K{\"a}ndler,
  Mette, R{\"o}der, and B{\"o}hner}{Dietrich et~al\mbox{.}}{2019}]%
        {dietrich2019temporal}
\bibfield{author}{\bibinfo{person}{Helge Dietrich}, \bibinfo{person}{Thilo
  Wolf}, \bibinfo{person}{Tobias Kawohl}, \bibinfo{person}{Jan Wehberg},
  \bibinfo{person}{Gerald K{\"a}ndler}, \bibinfo{person}{Tobias Mette},
  \bibinfo{person}{Arno R{\"o}der}, {and} \bibinfo{person}{J{\"u}rgen
  B{\"o}hner}.} \bibinfo{year}{2019}\natexlab{}.
\newblock \showarticletitle{Temporal and spatial high-resolution climate data
  from 1961 to 2100 for the German National Forest Inventory (NFI)}.
\newblock \bibinfo{journal}{\emph{Annals of Forest Science}}
  \bibinfo{volume}{76}, \bibinfo{number}{1} (\bibinfo{year}{2019}),
  \bibinfo{pages}{1--14}.
\newblock


\bibitem[\protect\citeauthoryear{Ding, Xia, and Yu}{Ding et~al\mbox{.}}{2020}]%
        {ding-etal-2020-ecpe}
\bibfield{author}{\bibinfo{person}{Zixiang Ding}, \bibinfo{person}{Rui Xia},
  {and} \bibinfo{person}{Jianfei Yu}.} \bibinfo{year}{2020}\natexlab{}.
\newblock \showarticletitle{{ECPE}-2{D}: Emotion-Cause Pair Extraction based on
  Joint Two-Dimensional Representation, Interaction and Prediction}. In
  \bibinfo{booktitle}{\emph{Proceedings of the 58th Annual Meeting of the
  Association for Computational Linguistics}}. \bibinfo{publisher}{Association
  for Computational Linguistics}, \bibinfo{address}{Online},
  \bibinfo{pages}{3161--3170}.
\newblock
\urldef\tempurl%
\url{https://doi.org/10.18653/v1/2020.acl-main.288}
\showDOI{\tempurl}


\bibitem[\protect\citeauthoryear{Dittadi, Tr{\"a}uble, Locatello, W{\"u}thrich,
  Agrawal, Winther, Bauer, and Sch{\"o}lkopf}{Dittadi et~al\mbox{.}}{2020}]%
        {dittadi2020transfer}
\bibfield{author}{\bibinfo{person}{Andrea Dittadi}, \bibinfo{person}{Frederik
  Tr{\"a}uble}, \bibinfo{person}{Francesco Locatello}, \bibinfo{person}{Manuel
  W{\"u}thrich}, \bibinfo{person}{Vaibhav Agrawal}, \bibinfo{person}{Ole
  Winther}, \bibinfo{person}{Stefan Bauer}, {and} \bibinfo{person}{Bernhard
  Sch{\"o}lkopf}.} \bibinfo{year}{2020}\natexlab{}.
\newblock \showarticletitle{On the transfer of disentangled representations in
  realistic settings}.
\newblock \bibinfo{journal}{\emph{arXiv preprint arXiv:2010.14407}}
  (\bibinfo{year}{2020}).
\newblock


\bibitem[\protect\citeauthoryear{Dougherty, Bluedorn, and Keon}{Dougherty
  et~al\mbox{.}}{1985}]%
        {dougherty1985precursors}
\bibfield{author}{\bibinfo{person}{Thomas~W Dougherty},
  \bibinfo{person}{Allen~C Bluedorn}, {and} \bibinfo{person}{Thomas~L Keon}.}
  \bibinfo{year}{1985}\natexlab{}.
\newblock \showarticletitle{Precursors of employee turnover: A multiple-sample
  causal analysis}.
\newblock \bibinfo{journal}{\emph{Journal of Organizational Behavior}}
  \bibinfo{volume}{6}, \bibinfo{number}{4} (\bibinfo{year}{1985}),
  \bibinfo{pages}{259--271}.
\newblock


\bibitem[\protect\citeauthoryear{Dror, Baumer, Bogomolov, and Reichart}{Dror
  et~al\mbox{.}}{2017}]%
        {dror2017replicability}
\bibfield{author}{\bibinfo{person}{Rotem Dror}, \bibinfo{person}{Gili Baumer},
  \bibinfo{person}{Marina Bogomolov}, {and} \bibinfo{person}{Roi Reichart}.}
  \bibinfo{year}{2017}\natexlab{}.
\newblock \showarticletitle{Replicability analysis for natural language
  processing: Testing significance with multiple datasets}.
\newblock \bibinfo{journal}{\emph{Transactions of the Association for
  Computational Linguistics}}  \bibinfo{volume}{5} (\bibinfo{year}{2017}),
  \bibinfo{pages}{471--486}.
\newblock


\bibitem[\protect\citeauthoryear{Entner and Hoyer}{Entner and Hoyer}{2010}]%
        {entner2010causal}
\bibfield{author}{\bibinfo{person}{Doris Entner} {and}
  \bibinfo{person}{Patrik~O Hoyer}.} \bibinfo{year}{2010}\natexlab{}.
\newblock \showarticletitle{On causal discovery from time series data using
  FCI}.
\newblock \bibinfo{journal}{\emph{Probabilistic graphical models}}
  (\bibinfo{year}{2010}), \bibinfo{pages}{121--128}.
\newblock


\bibitem[\protect\citeauthoryear{Entner and Hoyer}{Entner and Hoyer}{2011}]%
        {entner2011discovering}
\bibfield{author}{\bibinfo{person}{Doris Entner} {and}
  \bibinfo{person}{Patrik~O Hoyer}.} \bibinfo{year}{2011}\natexlab{}.
\newblock \showarticletitle{Discovering unconfounded causal relationships using
  linear non-gaussian models}. In \bibinfo{booktitle}{\emph{New Frontiers in
  Artificial Intelligence: JSAI-isAI 2010 Workshops, LENLS, JURISIN, AMBN, ISS,
  Tokyo, Japan, November 18-19, 2010, Revised Selected Papers 2}}. Springer,
  \bibinfo{pages}{181--195}.
\newblock


\bibitem[\protect\citeauthoryear{Erdem, Haspalamutgil, Palaz, Patoglu, and
  Uras}{Erdem et~al\mbox{.}}{2011}]%
        {5980160}
\bibfield{author}{\bibinfo{person}{Esra Erdem}, \bibinfo{person}{Kadir
  Haspalamutgil}, \bibinfo{person}{Can Palaz}, \bibinfo{person}{Volkan
  Patoglu}, {and} \bibinfo{person}{Tansel Uras}.}
  \bibinfo{year}{2011}\natexlab{}.
\newblock \showarticletitle{Combining high-level causal reasoning with
  low-level geometric reasoning and motion planning for robotic manipulation}.
  In \bibinfo{booktitle}{\emph{2011 IEEE International Conference on Robotics
  and Automation}}. \bibinfo{pages}{4575--4581}.
\newblock
\urldef\tempurl%
\url{https://doi.org/10.1109/ICRA.2011.5980160}
\showDOI{\tempurl}


\bibitem[\protect\citeauthoryear{Fan, Wang, Mo, Shi, and Tang}{Fan
  et~al\mbox{.}}{2022}]%
        {fan2022debiasing}
\bibfield{author}{\bibinfo{person}{Shaohua Fan}, \bibinfo{person}{Xiao Wang},
  \bibinfo{person}{Yanhu Mo}, \bibinfo{person}{Chuan Shi}, {and}
  \bibinfo{person}{Jian Tang}.} \bibinfo{year}{2022}\natexlab{}.
\newblock \showarticletitle{Debiasing Graph Neural Networks via Learning
  Disentangled Causal Substructure}.
\newblock \bibinfo{journal}{\emph{arXiv preprint arXiv:2209.14107}}
  (\bibinfo{year}{2022}).
\newblock


\bibitem[\protect\citeauthoryear{Feng, Liu, Yang, Tang, Du, and Hu}{Feng
  et~al\mbox{.}}{2021}]%
        {feng2021degree}
\bibfield{author}{\bibinfo{person}{Qizhang Feng}, \bibinfo{person}{Ninghao
  Liu}, \bibinfo{person}{Fan Yang}, \bibinfo{person}{Ruixiang Tang},
  \bibinfo{person}{Mengnan Du}, {and} \bibinfo{person}{Xia Hu}.}
  \bibinfo{year}{2021}\natexlab{}.
\newblock \showarticletitle{DEGREE: Decomposition Based Explanation for Graph
  Neural Networks}. In \bibinfo{booktitle}{\emph{International Conference on
  Learning Representations}}.
\newblock


\bibitem[\protect\citeauthoryear{Fong, Hazlett, and Imai}{Fong
  et~al\mbox{.}}{2018}]%
        {fong2018covariate}
\bibfield{author}{\bibinfo{person}{Christian Fong}, \bibinfo{person}{Chad
  Hazlett}, {and} \bibinfo{person}{Kosuke Imai}.}
  \bibinfo{year}{2018}\natexlab{}.
\newblock \showarticletitle{Covariate balancing propensity score for a
  continuous treatment: Application to the efficacy of political
  advertisements}.
\newblock \bibinfo{journal}{\emph{The Annals of Applied Statistics}}
  \bibinfo{volume}{12}, \bibinfo{number}{1} (\bibinfo{year}{2018}),
  \bibinfo{pages}{156--177}.
\newblock


\bibitem[\protect\citeauthoryear{Frangakis and Rubin}{Frangakis and
  Rubin}{2002}]%
        {frangakis2002principal}
\bibfield{author}{\bibinfo{person}{Constantine~E Frangakis} {and}
  \bibinfo{person}{Donald~B Rubin}.} \bibinfo{year}{2002}\natexlab{}.
\newblock \showarticletitle{Principal stratification in causal inference}.
\newblock \bibinfo{journal}{\emph{Biometrics}} \bibinfo{volume}{58},
  \bibinfo{number}{1} (\bibinfo{year}{2002}), \bibinfo{pages}{21--29}.
\newblock


\bibitem[\protect\citeauthoryear{Friedman, Murphy, and Russell}{Friedman
  et~al\mbox{.}}{2013}]%
        {friedman2013learning}
\bibfield{author}{\bibinfo{person}{Nir Friedman}, \bibinfo{person}{Kevin
  Murphy}, {and} \bibinfo{person}{Stuart Russell}.}
  \bibinfo{year}{2013}\natexlab{}.
\newblock \showarticletitle{Learning the structure of dynamic probabilistic
  networks}.
\newblock \bibinfo{journal}{\emph{arXiv preprint arXiv:1301.7374}}
  (\bibinfo{year}{2013}).
\newblock


\bibitem[\protect\citeauthoryear{Fu, Chan, and Chau}{Fu et~al\mbox{.}}{2013}]%
        {fu2013assessing}
\bibfield{author}{\bibinfo{person}{King-wa Fu}, \bibinfo{person}{Chung-hong
  Chan}, {and} \bibinfo{person}{Michael Chau}.}
  \bibinfo{year}{2013}\natexlab{}.
\newblock \showarticletitle{Assessing censorship on microblogs in China:
  Discriminatory keyword analysis and the real-name registration policy}.
\newblock \bibinfo{journal}{\emph{IEEE internet computing}}
  \bibinfo{volume}{17}, \bibinfo{number}{3} (\bibinfo{year}{2013}),
  \bibinfo{pages}{42--50}.
\newblock


\bibitem[\protect\citeauthoryear{Gabler, Geiger, Schuppler, and Kern}{Gabler
  et~al\mbox{.}}{2023}]%
        {gabler2023reconsidering}
\bibfield{author}{\bibinfo{person}{Philipp Gabler}, \bibinfo{person}{Bernhard~C
  Geiger}, \bibinfo{person}{Barbara Schuppler}, {and} \bibinfo{person}{Roman
  Kern}.} \bibinfo{year}{2023}\natexlab{}.
\newblock \showarticletitle{Reconsidering Read and Spontaneous Speech: Causal
  Perspectives on the Generation of Training Data for Automatic Speech
  Recognition}.
\newblock \bibinfo{journal}{\emph{Information}} \bibinfo{volume}{14},
  \bibinfo{number}{2} (\bibinfo{year}{2023}), \bibinfo{pages}{137}.
\newblock


\bibitem[\protect\citeauthoryear{Galanti, Nabati, and Wolf}{Galanti
  et~al\mbox{.}}{2020}]%
        {galanti2020critical}
\bibfield{author}{\bibinfo{person}{Tomer Galanti}, \bibinfo{person}{Ofir
  Nabati}, {and} \bibinfo{person}{Lior Wolf}.} \bibinfo{year}{2020}\natexlab{}.
\newblock \showarticletitle{A critical view of the structural causal model}.
\newblock \bibinfo{journal}{\emph{arXiv preprint arXiv:2002.10007}}
  (\bibinfo{year}{2020}).
\newblock


\bibitem[\protect\citeauthoryear{Gao, Zheng, Wang, Feng, He, and Li}{Gao
  et~al\mbox{.}}{2022}]%
        {gao2022causal}
\bibfield{author}{\bibinfo{person}{Chen Gao}, \bibinfo{person}{Yu Zheng},
  \bibinfo{person}{Wenjie Wang}, \bibinfo{person}{Fuli Feng},
  \bibinfo{person}{Xiangnan He}, {and} \bibinfo{person}{Yong Li}.}
  \bibinfo{year}{2022}\natexlab{}.
\newblock \showarticletitle{Causal inference in recommender systems: A survey
  and future directions}.
\newblock \bibinfo{journal}{\emph{arXiv preprint arXiv:2208.12397}}
  (\bibinfo{year}{2022}).
\newblock


\bibitem[\protect\citeauthoryear{Geiger, Lu, Icard, and Potts}{Geiger
  et~al\mbox{.}}{2021}]%
        {geiger2021causal}
\bibfield{author}{\bibinfo{person}{Atticus Geiger}, \bibinfo{person}{Hanson
  Lu}, \bibinfo{person}{Thomas Icard}, {and} \bibinfo{person}{Christopher
  Potts}.} \bibinfo{year}{2021}\natexlab{}.
\newblock \showarticletitle{Causal abstractions of neural networks}.
\newblock \bibinfo{journal}{\emph{Advances in Neural Information Processing
  Systems}}  \bibinfo{volume}{34} (\bibinfo{year}{2021}),
  \bibinfo{pages}{9574--9586}.
\newblock


\bibitem[\protect\citeauthoryear{Geiger, Wu, Potts, Icard, and Goodman}{Geiger
  et~al\mbox{.}}{2023}]%
        {geiger2023finding}
\bibfield{author}{\bibinfo{person}{Atticus Geiger}, \bibinfo{person}{Zhengxuan
  Wu}, \bibinfo{person}{Christopher Potts}, \bibinfo{person}{Thomas Icard},
  {and} \bibinfo{person}{Noah~D Goodman}.} \bibinfo{year}{2023}\natexlab{}.
\newblock \showarticletitle{Finding alignments between interpretable causal
  variables and distributed neural representations}.
\newblock \bibinfo{journal}{\emph{arXiv preprint arXiv:2303.02536}}
  (\bibinfo{year}{2023}).
\newblock


\bibitem[\protect\citeauthoryear{Girju}{Girju}{2003}]%
        {girju2003automatic}
\bibfield{author}{\bibinfo{person}{Roxana Girju}.}
  \bibinfo{year}{2003}\natexlab{}.
\newblock \showarticletitle{Automatic detection of causal relations for
  question answering}. In \bibinfo{booktitle}{\emph{Proceedings of the ACL 2003
  workshop on Multilingual summarization and question answering}}.
  \bibinfo{pages}{76--83}.
\newblock


\bibitem[\protect\citeauthoryear{Girju, Nakov, Nastase, Szpakowicz, Turney, and
  Yuret}{Girju et~al\mbox{.}}{2009}]%
        {girju2009classification}
\bibfield{author}{\bibinfo{person}{Roxana Girju}, \bibinfo{person}{Preslav
  Nakov}, \bibinfo{person}{Vivi Nastase}, \bibinfo{person}{Stan Szpakowicz},
  \bibinfo{person}{Peter Turney}, {and} \bibinfo{person}{Deniz Yuret}.}
  \bibinfo{year}{2009}\natexlab{}.
\newblock \showarticletitle{Classification of semantic relations between
  nominals}.
\newblock \bibinfo{journal}{\emph{Language Resources and Evaluation}}
  \bibinfo{volume}{43} (\bibinfo{year}{2009}), \bibinfo{pages}{105--121}.
\newblock


\bibitem[\protect\citeauthoryear{Glymour, Zhang, and Spirtes}{Glymour
  et~al\mbox{.}}{2019}]%
        {glymour2019review}
\bibfield{author}{\bibinfo{person}{Clark Glymour}, \bibinfo{person}{Kun Zhang},
  {and} \bibinfo{person}{Peter Spirtes}.} \bibinfo{year}{2019}\natexlab{}.
\newblock \showarticletitle{Review of causal discovery methods based on
  graphical models}.
\newblock \bibinfo{journal}{\emph{Frontiers in genetics}}  \bibinfo{volume}{10}
  (\bibinfo{year}{2019}), \bibinfo{pages}{524}.
\newblock


\bibitem[\protect\citeauthoryear{Gong, Yao, Zhang, Li, and Bi}{Gong
  et~al\mbox{.}}{2023}]%
        {gong2023causal}
\bibfield{author}{\bibinfo{person}{Chang Gong}, \bibinfo{person}{Di Yao},
  \bibinfo{person}{Chuzhe Zhang}, \bibinfo{person}{Wenbin Li}, {and}
  \bibinfo{person}{Jingping Bi}.} \bibinfo{year}{2023}\natexlab{}.
\newblock \showarticletitle{Causal Discovery from Temporal Data: An Overview
  and New Perspectives}.
\newblock \bibinfo{journal}{\emph{arXiv preprint arXiv:2303.10112}}
  (\bibinfo{year}{2023}).
\newblock


\bibitem[\protect\citeauthoryear{Goudet, Kalainathan, Caillou, Guyon,
  Lopez-Paz, and Sebag}{Goudet et~al\mbox{.}}{2017}]%
        {goudet2017causal}
\bibfield{author}{\bibinfo{person}{Olivier Goudet}, \bibinfo{person}{Diviyan
  Kalainathan}, \bibinfo{person}{Philippe Caillou}, \bibinfo{person}{Isabelle
  Guyon}, \bibinfo{person}{David Lopez-Paz}, {and} \bibinfo{person}{Mich{\`e}le
  Sebag}.} \bibinfo{year}{2017}\natexlab{}.
\newblock \showarticletitle{Causal generative neural networks}.
\newblock \bibinfo{journal}{\emph{arXiv preprint arXiv:1711.08936}}
  (\bibinfo{year}{2017}).
\newblock


\bibitem[\protect\citeauthoryear{Goudet, Kalainathan, Caillou, Guyon,
  Lopez-Paz, and Sebag}{Goudet et~al\mbox{.}}{2018}]%
        {goudet2018learning}
\bibfield{author}{\bibinfo{person}{Olivier Goudet}, \bibinfo{person}{Diviyan
  Kalainathan}, \bibinfo{person}{Philippe Caillou}, \bibinfo{person}{Isabelle
  Guyon}, \bibinfo{person}{David Lopez-Paz}, {and} \bibinfo{person}{Michele
  Sebag}.} \bibinfo{year}{2018}\natexlab{}.
\newblock \showarticletitle{Learning functional causal models with generative
  neural networks}.
\newblock \bibinfo{journal}{\emph{Explainable and interpretable models in
  computer vision and machine learning}} (\bibinfo{year}{2018}),
  \bibinfo{pages}{39--80}.
\newblock


\bibitem[\protect\citeauthoryear{Gould, Pisani, Gallo, Ertefaie, Harrington,
  Kelberman, and Green}{Gould et~al\mbox{.}}{2022}]%
        {gould2022crisis}
\bibfield{author}{\bibinfo{person}{Madelyn~S Gould}, \bibinfo{person}{Anthony
  Pisani}, \bibinfo{person}{Carlos Gallo}, \bibinfo{person}{Ashkan Ertefaie},
  \bibinfo{person}{Donald Harrington}, \bibinfo{person}{Caroline Kelberman},
  {and} \bibinfo{person}{Shannon Green}.} \bibinfo{year}{2022}\natexlab{}.
\newblock \showarticletitle{Crisis text-line interventions: Evaluation of
  texters' perceptions of effectiveness}.
\newblock \bibinfo{journal}{\emph{Suicide and Life-Threatening Behavior}}
  (\bibinfo{year}{2022}).
\newblock


\bibitem[\protect\citeauthoryear{Goyal, Lamb, Gampa, Beaudoin, Levine,
  Blundell, Bengio, and Mozer}{Goyal et~al\mbox{.}}{2020}]%
        {goyal2020object}
\bibfield{author}{\bibinfo{person}{Anirudh Goyal}, \bibinfo{person}{Alex Lamb},
  \bibinfo{person}{Phanideep Gampa}, \bibinfo{person}{Philippe Beaudoin},
  \bibinfo{person}{Sergey Levine}, \bibinfo{person}{Charles Blundell},
  \bibinfo{person}{Yoshua Bengio}, {and} \bibinfo{person}{Michael Mozer}.}
  \bibinfo{year}{2020}\natexlab{}.
\newblock \showarticletitle{Object files and schemata: Factorizing declarative
  and procedural knowledge in dynamical systems}.
\newblock \bibinfo{journal}{\emph{arXiv preprint arXiv:2006.16225}}
  (\bibinfo{year}{2020}).
\newblock


\bibitem[\protect\citeauthoryear{Granger}{Granger}{1969}]%
        {granger1969investigating}
\bibfield{author}{\bibinfo{person}{Clive~WJ Granger}.}
  \bibinfo{year}{1969}\natexlab{}.
\newblock \showarticletitle{Investigating causal relations by econometric
  models and cross-spectral methods}.
\newblock \bibinfo{journal}{\emph{Econometrica: journal of the Econometric
  Society}} (\bibinfo{year}{1969}), \bibinfo{pages}{424--438}.
\newblock


\bibitem[\protect\citeauthoryear{Guo, Cheng, Li, Hahn, and Liu}{Guo
  et~al\mbox{.}}{2020}]%
        {guo2020survey}
\bibfield{author}{\bibinfo{person}{Ruocheng Guo}, \bibinfo{person}{Lu Cheng},
  \bibinfo{person}{Jundong Li}, \bibinfo{person}{P~Richard Hahn}, {and}
  \bibinfo{person}{Huan Liu}.} \bibinfo{year}{2020}\natexlab{}.
\newblock \showarticletitle{A survey of learning causality with data: Problems
  and methods}.
\newblock \bibinfo{journal}{\emph{ACM Computing Surveys (CSUR)}}
  \bibinfo{volume}{53}, \bibinfo{number}{4} (\bibinfo{year}{2020}),
  \bibinfo{pages}{1--37}.
\newblock


\bibitem[\protect\citeauthoryear{Guo, Zhang, Feng, and Chen}{Guo
  et~al\mbox{.}}{2021}]%
        {guo2021causal}
\bibfield{author}{\bibinfo{person}{Zongyu Guo}, \bibinfo{person}{Zhizheng
  Zhang}, \bibinfo{person}{Runsen Feng}, {and} \bibinfo{person}{Zhibo Chen}.}
  \bibinfo{year}{2021}\natexlab{}.
\newblock \showarticletitle{Causal contextual prediction for learned image
  compression}.
\newblock \bibinfo{journal}{\emph{IEEE Transactions on Circuits and Systems for
  Video Technology}} \bibinfo{volume}{32}, \bibinfo{number}{4}
  (\bibinfo{year}{2021}), \bibinfo{pages}{2329--2341}.
\newblock


\bibitem[\protect\citeauthoryear{Guvenir, Acar, Demiroz, and Cekin}{Guvenir
  et~al\mbox{.}}{1997}]%
        {647926}
\bibfield{author}{\bibinfo{person}{H.A. Guvenir}, \bibinfo{person}{B. Acar},
  \bibinfo{person}{G. Demiroz}, {and} \bibinfo{person}{A. Cekin}.}
  \bibinfo{year}{1997}\natexlab{}.
\newblock \showarticletitle{A supervised machine learning algorithm for
  arrhythmia analysis}. In \bibinfo{booktitle}{\emph{Computers in Cardiology
  1997}}. \bibinfo{pages}{433--436}.
\newblock
\urldef\tempurl%
\url{https://doi.org/10.1109/CIC.1997.647926}
\showDOI{\tempurl}


\bibitem[\protect\citeauthoryear{Hafiani, Jaouad, and Mahi}{Hafiani
  et~al\mbox{.}}{2023}]%
        {hafiani2023rare}
\bibfield{author}{\bibinfo{person}{H Hafiani}, \bibinfo{person}{MR~Cherkaoui
  Jaouad}, {and} \bibinfo{person}{M Mahi}.} \bibinfo{year}{2023}\natexlab{}.
\newblock \showarticletitle{Rare images of renal mass and lithiasis in native
  kidney after renal transplantation in a patient with hematuria: Casual or
  correlated?}
\newblock \bibinfo{journal}{\emph{Urology Case Reports}}
  (\bibinfo{year}{2023}), \bibinfo{pages}{102475}.
\newblock


\bibitem[\protect\citeauthoryear{Hahn, Murray, and Carvalho}{Hahn
  et~al\mbox{.}}{2020}]%
        {hahn2020bayesian}
\bibfield{author}{\bibinfo{person}{P~Richard Hahn}, \bibinfo{person}{Jared~S
  Murray}, {and} \bibinfo{person}{Carlos~M Carvalho}.}
  \bibinfo{year}{2020}\natexlab{}.
\newblock \showarticletitle{Bayesian regression tree models for causal
  inference: Regularization, confounding, and heterogeneous effects (with
  discussion)}.
\newblock \bibinfo{journal}{\emph{Bayesian Analysis}} \bibinfo{volume}{15},
  \bibinfo{number}{3} (\bibinfo{year}{2020}), \bibinfo{pages}{965--1056}.
\newblock


\bibitem[\protect\citeauthoryear{Hauser and B{\"u}hlmann}{Hauser and
  B{\"u}hlmann}{2012}]%
        {hauser2012characterization}
\bibfield{author}{\bibinfo{person}{Alain Hauser} {and} \bibinfo{person}{Peter
  B{\"u}hlmann}.} \bibinfo{year}{2012}\natexlab{}.
\newblock \showarticletitle{Characterization and greedy learning of
  interventional Markov equivalence classes of directed acyclic graphs}.
\newblock \bibinfo{journal}{\emph{The Journal of Machine Learning Research}}
  \bibinfo{volume}{13}, \bibinfo{number}{1} (\bibinfo{year}{2012}),
  \bibinfo{pages}{2409--2464}.
\newblock


\bibitem[\protect\citeauthoryear{Hazan, Brossier, Marxer, and Purwins}{Hazan
  et~al\mbox{.}}{2007}]%
        {hazan2007causal}
\bibfield{author}{\bibinfo{person}{Amaury Hazan}, \bibinfo{person}{Paul
  Brossier}, \bibinfo{person}{Ricard Marxer}, {and} \bibinfo{person}{Hendrik
  Purwins}.} \bibinfo{year}{2007}\natexlab{}.
\newblock \showarticletitle{What/when causal expectation modelling in
  monophonic pitched and percussive audio}. In \bibinfo{booktitle}{\emph{NIPS
  music, brain and cognition workshop. Whistler, CA}}.
\newblock


\bibitem[\protect\citeauthoryear{Hazirbas, Bitton, Dolhansky, Pan, Gordo, and
  Ferrer}{Hazirbas et~al\mbox{.}}{2021}]%
        {hazirbas2021towards}
\bibfield{author}{\bibinfo{person}{Caner Hazirbas}, \bibinfo{person}{Joanna
  Bitton}, \bibinfo{person}{Brian Dolhansky}, \bibinfo{person}{Jacqueline Pan},
  \bibinfo{person}{Albert Gordo}, {and} \bibinfo{person}{Cristian~Canton
  Ferrer}.} \bibinfo{year}{2021}\natexlab{}.
\newblock \showarticletitle{Towards measuring fairness in ai: the casual
  conversations dataset}.
\newblock \bibinfo{journal}{\emph{IEEE Transactions on Biometrics, Behavior,
  and Identity Science}} (\bibinfo{year}{2021}).
\newblock


\bibitem[\protect\citeauthoryear{He, Liu, Gao, and Chen}{He
  et~al\mbox{.}}{2020}]%
        {he2020deberta}
\bibfield{author}{\bibinfo{person}{Pengcheng He}, \bibinfo{person}{Xiaodong
  Liu}, \bibinfo{person}{Jianfeng Gao}, {and} \bibinfo{person}{Weizhu Chen}.}
  \bibinfo{year}{2020}\natexlab{}.
\newblock \showarticletitle{Deberta: Decoding-enhanced bert with disentangled
  attention}.
\newblock \bibinfo{journal}{\emph{arXiv preprint arXiv:2006.03654}}
  (\bibinfo{year}{2020}).
\newblock


\bibitem[\protect\citeauthoryear{Hearst, Dumais, Osuna, Platt, and
  Scholkopf}{Hearst et~al\mbox{.}}{1998}]%
        {hearst1998support}
\bibfield{author}{\bibinfo{person}{Marti~A. Hearst}, \bibinfo{person}{Susan~T
  Dumais}, \bibinfo{person}{Edgar Osuna}, \bibinfo{person}{John Platt}, {and}
  \bibinfo{person}{Bernhard Scholkopf}.} \bibinfo{year}{1998}\natexlab{}.
\newblock \showarticletitle{Support vector machines}.
\newblock \bibinfo{journal}{\emph{IEEE Intelligent Systems and their
  applications}} \bibinfo{volume}{13}, \bibinfo{number}{4}
  (\bibinfo{year}{1998}), \bibinfo{pages}{18--28}.
\newblock


\bibitem[\protect\citeauthoryear{Herrett, Gallagher, Bhaskaran, Forbes, Mathur,
  Van~Staa, and Smeeth}{Herrett et~al\mbox{.}}{2015}]%
        {herrett2015data}
\bibfield{author}{\bibinfo{person}{Emily Herrett}, \bibinfo{person}{Arlene~M
  Gallagher}, \bibinfo{person}{Krishnan Bhaskaran}, \bibinfo{person}{Harriet
  Forbes}, \bibinfo{person}{Rohini Mathur}, \bibinfo{person}{Tjeerd Van~Staa},
  {and} \bibinfo{person}{Liam Smeeth}.} \bibinfo{year}{2015}\natexlab{}.
\newblock \showarticletitle{Data resource profile: clinical practice research
  datalink (CPRD)}.
\newblock \bibinfo{journal}{\emph{International journal of epidemiology}}
  \bibinfo{volume}{44}, \bibinfo{number}{3} (\bibinfo{year}{2015}),
  \bibinfo{pages}{827--836}.
\newblock


\bibitem[\protect\citeauthoryear{Hill}{Hill}{2011}]%
        {hill2011bayesian}
\bibfield{author}{\bibinfo{person}{Jennifer~L Hill}.}
  \bibinfo{year}{2011}\natexlab{}.
\newblock \showarticletitle{Bayesian nonparametric modeling for causal
  inference}.
\newblock \bibinfo{journal}{\emph{Journal of Computational and Graphical
  Statistics}} \bibinfo{volume}{20}, \bibinfo{number}{1}
  (\bibinfo{year}{2011}), \bibinfo{pages}{217--240}.
\newblock


\bibitem[\protect\citeauthoryear{Hoyer, Janzing, Mooij, Peters, and
  Sch{\"o}lkopf}{Hoyer et~al\mbox{.}}{2008a}]%
        {hoyer2008nonlinear}
\bibfield{author}{\bibinfo{person}{Patrik Hoyer}, \bibinfo{person}{Dominik
  Janzing}, \bibinfo{person}{Joris~M Mooij}, \bibinfo{person}{Jonas Peters},
  {and} \bibinfo{person}{Bernhard Sch{\"o}lkopf}.}
  \bibinfo{year}{2008}\natexlab{a}.
\newblock \showarticletitle{Nonlinear causal discovery with additive noise
  models}.
\newblock \bibinfo{journal}{\emph{Advances in neural information processing
  systems}}  \bibinfo{volume}{21} (\bibinfo{year}{2008}).
\newblock


\bibitem[\protect\citeauthoryear{Hoyer, Shimizu, Kerminen, and
  Palviainen}{Hoyer et~al\mbox{.}}{2008b}]%
        {hoyer2008estimation}
\bibfield{author}{\bibinfo{person}{Patrik~O Hoyer}, \bibinfo{person}{Shohei
  Shimizu}, \bibinfo{person}{Antti~J Kerminen}, {and} \bibinfo{person}{Markus
  Palviainen}.} \bibinfo{year}{2008}\natexlab{b}.
\newblock \showarticletitle{Estimation of causal effects using linear
  non-Gaussian causal models with hidden variables}.
\newblock \bibinfo{journal}{\emph{International Journal of Approximate
  Reasoning}} \bibinfo{volume}{49}, \bibinfo{number}{2} (\bibinfo{year}{2008}),
  \bibinfo{pages}{362--378}.
\newblock


\bibitem[\protect\citeauthoryear{Hu, Cotter, Mohan, Gurau, and Kendall}{Hu
  et~al\mbox{.}}{2020}]%
        {hu2020probabilistic}
\bibfield{author}{\bibinfo{person}{Anthony Hu}, \bibinfo{person}{Fergal
  Cotter}, \bibinfo{person}{Nikhil Mohan}, \bibinfo{person}{Corina Gurau},
  {and} \bibinfo{person}{Alex Kendall}.} \bibinfo{year}{2020}\natexlab{}.
\newblock \showarticletitle{Probabilistic future prediction for video scene
  understanding}. In \bibinfo{booktitle}{\emph{European Conference on Computer
  Vision}}. Springer, \bibinfo{pages}{767--785}.
\newblock


\bibitem[\protect\citeauthoryear{Hu and Li}{Hu and Li}{2021}]%
        {hu2021causal}
\bibfield{author}{\bibinfo{person}{Zhiting Hu} {and} \bibinfo{person}{Li~Erran
  Li}.} \bibinfo{year}{2021}\natexlab{}.
\newblock \showarticletitle{A causal lens for controllable text generation}.
\newblock \bibinfo{journal}{\emph{Advances in Neural Information Processing
  Systems}}  \bibinfo{volume}{34} (\bibinfo{year}{2021}),
  \bibinfo{pages}{24941--24955}.
\newblock


\bibitem[\protect\citeauthoryear{Huang, Zhang, Lin, Sch{\"o}lkopf, and
  Glymour}{Huang et~al\mbox{.}}{2018}]%
        {huang2018generalized}
\bibfield{author}{\bibinfo{person}{Biwei Huang}, \bibinfo{person}{Kun Zhang},
  \bibinfo{person}{Yizhu Lin}, \bibinfo{person}{Bernhard Sch{\"o}lkopf}, {and}
  \bibinfo{person}{Clark Glymour}.} \bibinfo{year}{2018}\natexlab{}.
\newblock \showarticletitle{Generalized score functions for causal discovery}.
  In \bibinfo{booktitle}{\emph{Proceedings of the 24th ACM SIGKDD international
  conference on knowledge discovery \& data mining}}.
  \bibinfo{pages}{1551--1560}.
\newblock


\bibitem[\protect\citeauthoryear{Huang, Zhang, Zhang, Ramsey, Sanchez-Romero,
  Glymour, and Sch{\"o}lkopf}{Huang et~al\mbox{.}}{2020}]%
        {huang2020causal}
\bibfield{author}{\bibinfo{person}{Biwei Huang}, \bibinfo{person}{Kun Zhang},
  \bibinfo{person}{Jiji Zhang}, \bibinfo{person}{Joseph~D Ramsey},
  \bibinfo{person}{Ruben Sanchez-Romero}, \bibinfo{person}{Clark Glymour},
  {and} \bibinfo{person}{Bernhard Sch{\"o}lkopf}.}
  \bibinfo{year}{2020}\natexlab{}.
\newblock \showarticletitle{Causal Discovery from Heterogeneous/Nonstationary
  Data.}
\newblock \bibinfo{journal}{\emph{J. Mach. Learn. Res.}} \bibinfo{volume}{21},
  \bibinfo{number}{89} (\bibinfo{year}{2020}), \bibinfo{pages}{1--53}.
\newblock


\bibitem[\protect\citeauthoryear{Hullsiek and Louis}{Hullsiek and
  Louis}{2002}]%
        {hullsiek2002propensity}
\bibfield{author}{\bibinfo{person}{Katherine~Huppler Hullsiek} {and}
  \bibinfo{person}{Thomas~A Louis}.} \bibinfo{year}{2002}\natexlab{}.
\newblock \showarticletitle{Propensity score modeling strategies for the causal
  analysis of observational data}.
\newblock \bibinfo{journal}{\emph{Biostatistics}} \bibinfo{volume}{3},
  \bibinfo{number}{2} (\bibinfo{year}{2002}), \bibinfo{pages}{179--193}.
\newblock


\bibitem[\protect\citeauthoryear{Hyv{\"a}rinen, Zhang, Shimizu, and
  Hoyer}{Hyv{\"a}rinen et~al\mbox{.}}{2010}]%
        {hyvarinen2010estimation}
\bibfield{author}{\bibinfo{person}{Aapo Hyv{\"a}rinen}, \bibinfo{person}{Kun
  Zhang}, \bibinfo{person}{Shohei Shimizu}, {and} \bibinfo{person}{Patrik~O
  Hoyer}.} \bibinfo{year}{2010}\natexlab{}.
\newblock \showarticletitle{Estimation of a structural vector autoregression
  model using non-gaussianity.}
\newblock \bibinfo{journal}{\emph{Journal of Machine Learning Research}}
  \bibinfo{volume}{11}, \bibinfo{number}{5} (\bibinfo{year}{2010}).
\newblock


\bibitem[\protect\citeauthoryear{Iacoviello and Navarro}{Iacoviello and
  Navarro}{2019}]%
        {iacoviello2019foreign}
\bibfield{author}{\bibinfo{person}{Matteo Iacoviello} {and}
  \bibinfo{person}{Gaston Navarro}.} \bibinfo{year}{2019}\natexlab{}.
\newblock \showarticletitle{Foreign effects of higher US interest rates}.
\newblock \bibinfo{journal}{\emph{Journal of International Money and Finance}}
  \bibinfo{volume}{95} (\bibinfo{year}{2019}), \bibinfo{pages}{232--250}.
\newblock


\bibitem[\protect\citeauthoryear{Iacus, King, and Porro}{Iacus
  et~al\mbox{.}}{2012}]%
        {iacus2012causal}
\bibfield{author}{\bibinfo{person}{Stefano~M Iacus}, \bibinfo{person}{Gary
  King}, {and} \bibinfo{person}{Giuseppe Porro}.}
  \bibinfo{year}{2012}\natexlab{}.
\newblock \showarticletitle{Causal inference without balance checking:
  Coarsened exact matching}.
\newblock \bibinfo{journal}{\emph{Political analysis}} \bibinfo{volume}{20},
  \bibinfo{number}{1} (\bibinfo{year}{2012}), \bibinfo{pages}{1--24}.
\newblock


\bibitem[\protect\citeauthoryear{Imai and Ratkovic}{Imai and Ratkovic}{2014}]%
        {imai2014covariate}
\bibfield{author}{\bibinfo{person}{Kosuke Imai} {and} \bibinfo{person}{Marc
  Ratkovic}.} \bibinfo{year}{2014}\natexlab{}.
\newblock \showarticletitle{Covariate balancing propensity score}.
\newblock \bibinfo{journal}{\emph{Journal of the Royal Statistical Society
  Series B: Statistical Methodology}} \bibinfo{volume}{76}, \bibinfo{number}{1}
  (\bibinfo{year}{2014}), \bibinfo{pages}{243--263}.
\newblock


\bibitem[\protect\citeauthoryear{Jack~Jr, Bernstein, Fox, Thompson, Alexander,
  Harvey, Borowski, Britson, L.~Whitwell, Ward, et~al\mbox{.}}{Jack~Jr
  et~al\mbox{.}}{2008}]%
        {jack2008alzheimer}
\bibfield{author}{\bibinfo{person}{Clifford~R Jack~Jr}, \bibinfo{person}{Matt~A
  Bernstein}, \bibinfo{person}{Nick~C Fox}, \bibinfo{person}{Paul Thompson},
  \bibinfo{person}{Gene Alexander}, \bibinfo{person}{Danielle Harvey},
  \bibinfo{person}{Bret Borowski}, \bibinfo{person}{Paula~J Britson},
  \bibinfo{person}{Jennifer L.~Whitwell}, \bibinfo{person}{Chadwick Ward},
  {et~al\mbox{.}}} \bibinfo{year}{2008}\natexlab{}.
\newblock \showarticletitle{The Alzheimer's disease neuroimaging initiative
  (ADNI): MRI methods}.
\newblock \bibinfo{journal}{\emph{Journal of Magnetic Resonance Imaging: An
  Official Journal of the International Society for Magnetic Resonance in
  Medicine}} \bibinfo{volume}{27}, \bibinfo{number}{4} (\bibinfo{year}{2008}),
  \bibinfo{pages}{685--691}.
\newblock


\bibitem[\protect\citeauthoryear{Janzing, Mooij, Zhang, Lemeire, Zscheischler,
  Daniu{\v{s}}is, Steudel, and Sch{\"o}lkopf}{Janzing et~al\mbox{.}}{2012}]%
        {janzing2012information}
\bibfield{author}{\bibinfo{person}{Dominik Janzing}, \bibinfo{person}{Joris
  Mooij}, \bibinfo{person}{Kun Zhang}, \bibinfo{person}{Jan Lemeire},
  \bibinfo{person}{Jakob Zscheischler}, \bibinfo{person}{Povilas
  Daniu{\v{s}}is}, \bibinfo{person}{Bastian Steudel}, {and}
  \bibinfo{person}{Bernhard Sch{\"o}lkopf}.} \bibinfo{year}{2012}\natexlab{}.
\newblock \showarticletitle{Information-geometric approach to inferring causal
  directions}.
\newblock \bibinfo{journal}{\emph{Artificial Intelligence}}
  \bibinfo{volume}{182} (\bibinfo{year}{2012}), \bibinfo{pages}{1--31}.
\newblock


\bibitem[\protect\citeauthoryear{Jiang and Aragam}{Jiang and Aragam}{2023}]%
        {jiang2023learning}
\bibfield{author}{\bibinfo{person}{Yibo Jiang} {and} \bibinfo{person}{Bryon
  Aragam}.} \bibinfo{year}{2023}\natexlab{}.
\newblock \showarticletitle{Learning nonparametric latent causal graphs with
  unknown interventions}.
\newblock \bibinfo{journal}{\emph{arXiv preprint arXiv:2306.02899}}
  (\bibinfo{year}{2023}).
\newblock


\bibitem[\protect\citeauthoryear{Jin, Liu, Lyu, Poff, Sachan, Mihalcea, Diab,
  and Schölkopf}{Jin et~al\mbox{.}}{2023}]%
        {jin2023large}
\bibfield{author}{\bibinfo{person}{Zhijing Jin}, \bibinfo{person}{Jiarui Liu},
  \bibinfo{person}{Zhiheng Lyu}, \bibinfo{person}{Spencer Poff},
  \bibinfo{person}{Mrinmaya Sachan}, \bibinfo{person}{Rada Mihalcea},
  \bibinfo{person}{Mona Diab}, {and} \bibinfo{person}{Bernhard Schölkopf}.}
  \bibinfo{year}{2023}\natexlab{}.
\newblock \bibinfo{title}{Can Large Language Models Infer Causation from
  Correlation?}
\newblock
\newblock
\showeprint[arxiv]{2306.05836}~[cs.CL]


\bibitem[\protect\citeauthoryear{Kaddour, Lynch, Liu, Kusner, and
  Silva}{Kaddour et~al\mbox{.}}{2022}]%
        {kaddour2022causal}
\bibfield{author}{\bibinfo{person}{Jean Kaddour}, \bibinfo{person}{Aengus
  Lynch}, \bibinfo{person}{Qi Liu}, \bibinfo{person}{Matt~J Kusner}, {and}
  \bibinfo{person}{Ricardo Silva}.} \bibinfo{year}{2022}\natexlab{}.
\newblock \showarticletitle{Causal machine learning: A survey and open
  problems}.
\newblock \bibinfo{journal}{\emph{arXiv preprint arXiv:2206.15475}}
  (\bibinfo{year}{2022}).
\newblock


\bibitem[\protect\citeauthoryear{Kalainathan}{Kalainathan}{2019}]%
        {kalainathan2019generative}
\bibfield{author}{\bibinfo{person}{Diviyan Kalainathan}.}
  \bibinfo{year}{2019}\natexlab{}.
\newblock \emph{\bibinfo{title}{Generative neural networks to infer causal
  mechanisms: algorithms and applications}}.
\newblock \bibinfo{thesistype}{Ph.\,D. Dissertation}.
  \bibinfo{school}{Universit{\'e} Paris Saclay (COmUE)}.
\newblock


\bibitem[\protect\citeauthoryear{Kalisch and B{\"u}hlman}{Kalisch and
  B{\"u}hlman}{2007}]%
        {kalisch2007estimating}
\bibfield{author}{\bibinfo{person}{Markus Kalisch} {and} \bibinfo{person}{Peter
  B{\"u}hlman}.} \bibinfo{year}{2007}\natexlab{}.
\newblock \showarticletitle{Estimating high-dimensional directed acyclic graphs
  with the PC-algorithm.}
\newblock \bibinfo{journal}{\emph{Journal of Machine Learning Research}}
  \bibinfo{volume}{8}, \bibinfo{number}{3} (\bibinfo{year}{2007}).
\newblock


\bibitem[\protect\citeauthoryear{Kallus, Mao, and Zhou}{Kallus
  et~al\mbox{.}}{2019}]%
        {kallus2019interval}
\bibfield{author}{\bibinfo{person}{Nathan Kallus}, \bibinfo{person}{Xiaojie
  Mao}, {and} \bibinfo{person}{Angela Zhou}.} \bibinfo{year}{2019}\natexlab{}.
\newblock \showarticletitle{Interval estimation of individual-level causal
  effects under unobserved confounding}. In \bibinfo{booktitle}{\emph{The 22nd
  international conference on artificial intelligence and statistics}}. PMLR,
  \bibinfo{pages}{2281--2290}.
\newblock


\bibitem[\protect\citeauthoryear{Kang and Tian}{Kang and Tian}{2009}]%
        {kang2009markov}
\bibfield{author}{\bibinfo{person}{Changsung Kang} {and} \bibinfo{person}{Jin
  Tian}.} \bibinfo{year}{2009}\natexlab{}.
\newblock \showarticletitle{Markov Properties for Linear Causal Models with
  Correlated Errors.}
\newblock \bibinfo{journal}{\emph{Journal of Machine Learning Research}}
  \bibinfo{volume}{10}, \bibinfo{number}{1} (\bibinfo{year}{2009}).
\newblock


\bibitem[\protect\citeauthoryear{Kann, Ebrahimi, Koh, Dudy, and Roncone}{Kann
  et~al\mbox{.}}{2022}]%
        {kann2022open}
\bibfield{author}{\bibinfo{person}{Katharina Kann}, \bibinfo{person}{Abteen
  Ebrahimi}, \bibinfo{person}{Joewie Koh}, \bibinfo{person}{Shiran Dudy}, {and}
  \bibinfo{person}{Alessandro Roncone}.} \bibinfo{year}{2022}\natexlab{}.
\newblock \showarticletitle{Open-domain Dialogue Generation: What We Can Do,
  Cannot Do, And Should Do Next}. In \bibinfo{booktitle}{\emph{Proceedings of
  the 4th Workshop on NLP for Conversational AI}}. \bibinfo{pages}{148--165}.
\newblock


\bibitem[\protect\citeauthoryear{Karkkainen and Joo}{Karkkainen and
  Joo}{2021}]%
        {karkkainen2021fairface}
\bibfield{author}{\bibinfo{person}{Kimmo Karkkainen} {and}
  \bibinfo{person}{Jungseock Joo}.} \bibinfo{year}{2021}\natexlab{}.
\newblock \showarticletitle{Fairface: Face attribute dataset for balanced race,
  gender, and age for bias measurement and mitigation}. In
  \bibinfo{booktitle}{\emph{Proceedings of the IEEE/CVF Winter Conference on
  Applications of Computer Vision}}. \bibinfo{pages}{1548--1558}.
\newblock


\bibitem[\protect\citeauthoryear{Ke, Wang, Mitrovic, Szummer, Rezende,
  et~al\mbox{.}}{Ke et~al\mbox{.}}{2020}]%
        {ke2020amortized}
\bibfield{author}{\bibinfo{person}{Nan~Rosemary Ke}, \bibinfo{person}{Jane
  Wang}, \bibinfo{person}{Jovana Mitrovic}, \bibinfo{person}{Martin Szummer},
  \bibinfo{person}{Danilo~J Rezende}, {et~al\mbox{.}}}
  \bibinfo{year}{2020}\natexlab{}.
\newblock \showarticletitle{Amortized learning of neural causal
  representations}.
\newblock \bibinfo{journal}{\emph{arXiv preprint arXiv:2008.09301}}
  (\bibinfo{year}{2020}).
\newblock


\bibitem[\protect\citeauthoryear{Kew and Hodkinson}{Kew and Hodkinson}{2006}]%
        {kew2006membranous}
\bibfield{author}{\bibinfo{person}{MC Kew} {and} \bibinfo{person}{HJ
  Hodkinson}.} \bibinfo{year}{2006}\natexlab{}.
\newblock \showarticletitle{Membranous obstruction of the inferior vena cava
  and its causal relation to hepatocellular carcinoma}.
\newblock \bibinfo{journal}{\emph{Liver International}} \bibinfo{volume}{26},
  \bibinfo{number}{1} (\bibinfo{year}{2006}), \bibinfo{pages}{1--7}.
\newblock


\bibitem[\protect\citeauthoryear{Khoo, Chan, and Niu}{Khoo
  et~al\mbox{.}}{2000}]%
        {khoo2000extracting}
\bibfield{author}{\bibinfo{person}{Christopher~SG Khoo}, \bibinfo{person}{Syin
  Chan}, {and} \bibinfo{person}{Yun Niu}.} \bibinfo{year}{2000}\natexlab{}.
\newblock \showarticletitle{Extracting causal knowledge from a medical database
  using graphical patterns}. In \bibinfo{booktitle}{\emph{Proceedings of the
  38th annual meeting of the association for computational linguistics}}.
  \bibinfo{pages}{336--343}.
\newblock


\bibitem[\protect\citeauthoryear{Knyazev, Taylor, and Amer}{Knyazev
  et~al\mbox{.}}{2019}]%
        {knyazev2019understanding}
\bibfield{author}{\bibinfo{person}{Boris Knyazev}, \bibinfo{person}{Graham~W
  Taylor}, {and} \bibinfo{person}{Mohamed Amer}.}
  \bibinfo{year}{2019}\natexlab{}.
\newblock \showarticletitle{Understanding attention and generalization in graph
  neural networks}.
\newblock \bibinfo{journal}{\emph{Advances in neural information processing
  systems}}  \bibinfo{volume}{32} (\bibinfo{year}{2019}).
\newblock


\bibitem[\protect\citeauthoryear{Kocaoglu, Jaber, Shanmugam, and
  Bareinboim}{Kocaoglu et~al\mbox{.}}{2019}]%
        {kocaoglu2019characterization}
\bibfield{author}{\bibinfo{person}{Murat Kocaoglu}, \bibinfo{person}{Amin
  Jaber}, \bibinfo{person}{Karthikeyan Shanmugam}, {and} \bibinfo{person}{Elias
  Bareinboim}.} \bibinfo{year}{2019}\natexlab{}.
\newblock \showarticletitle{Characterization and learning of causal graphs with
  latent variables from soft interventions}.
\newblock \bibinfo{journal}{\emph{Advances in Neural Information Processing
  Systems}}  \bibinfo{volume}{32} (\bibinfo{year}{2019}).
\newblock


\bibitem[\protect\citeauthoryear{Kollias and Zafeiriou}{Kollias and
  Zafeiriou}{2018}]%
        {kollias2018aff}
\bibfield{author}{\bibinfo{person}{Dimitrios Kollias} {and}
  \bibinfo{person}{Stefanos Zafeiriou}.} \bibinfo{year}{2018}\natexlab{}.
\newblock \showarticletitle{Aff-wild2: Extending the aff-wild database for
  affect recognition}.
\newblock \bibinfo{journal}{\emph{arXiv preprint arXiv:1811.07770}}
  (\bibinfo{year}{2018}).
\newblock


\bibitem[\protect\citeauthoryear{Kpotufe, Sgouritsa, Janzing, and
  Sch{\"o}lkopf}{Kpotufe et~al\mbox{.}}{2014}]%
        {kpotufe2014consistency}
\bibfield{author}{\bibinfo{person}{Samory Kpotufe}, \bibinfo{person}{Eleni
  Sgouritsa}, \bibinfo{person}{Dominik Janzing}, {and}
  \bibinfo{person}{Bernhard Sch{\"o}lkopf}.} \bibinfo{year}{2014}\natexlab{}.
\newblock \showarticletitle{Consistency of causal inference under the additive
  noise model}. In \bibinfo{booktitle}{\emph{International Conference on
  Machine Learning}}. PMLR, \bibinfo{pages}{478--486}.
\newblock


\bibitem[\protect\citeauthoryear{Krizhevsky, Hinton, et~al\mbox{.}}{Krizhevsky
  et~al\mbox{.}}{2009}]%
        {krizhevsky2009learning}
\bibfield{author}{\bibinfo{person}{Alex Krizhevsky}, \bibinfo{person}{Geoffrey
  Hinton}, {et~al\mbox{.}}} \bibinfo{year}{2009}\natexlab{}.
\newblock \showarticletitle{Learning multiple layers of features from tiny
  images}.
\newblock  (\bibinfo{year}{2009}).
\newblock


\bibitem[\protect\citeauthoryear{Kuang, Cui, Li, Jiang, Yang, and Wang}{Kuang
  et~al\mbox{.}}{2017}]%
        {kuang2017treatment}
\bibfield{author}{\bibinfo{person}{Kun Kuang}, \bibinfo{person}{Peng Cui},
  \bibinfo{person}{Bo Li}, \bibinfo{person}{Meng Jiang},
  \bibinfo{person}{Shiqiang Yang}, {and} \bibinfo{person}{Fei Wang}.}
  \bibinfo{year}{2017}\natexlab{}.
\newblock \showarticletitle{Treatment effect estimation with data-driven
  variable decomposition}. In \bibinfo{booktitle}{\emph{Proceedings of the AAAI
  Conference on Artificial Intelligence}}, Vol.~\bibinfo{volume}{31}.
\newblock


\bibitem[\protect\citeauthoryear{Kuhn}{Kuhn}{2012}]%
        {kuhn2012development}
\bibfield{author}{\bibinfo{person}{Deanna Kuhn}.}
  \bibinfo{year}{2012}\natexlab{}.
\newblock \showarticletitle{The development of causal reasoning}.
\newblock \bibinfo{journal}{\emph{Wiley Interdisciplinary Reviews: Cognitive
  Science}} \bibinfo{volume}{3}, \bibinfo{number}{3} (\bibinfo{year}{2012}),
  \bibinfo{pages}{327--335}.
\newblock


\bibitem[\protect\citeauthoryear{Kulkarni, Gupta, Ionescu, Borgeaud, Reynolds,
  Zisserman, and Mnih}{Kulkarni et~al\mbox{.}}{2019}]%
        {kulkarni2019unsupervised}
\bibfield{author}{\bibinfo{person}{Tejas~D Kulkarni}, \bibinfo{person}{Ankush
  Gupta}, \bibinfo{person}{Catalin Ionescu}, \bibinfo{person}{Sebastian
  Borgeaud}, \bibinfo{person}{Malcolm Reynolds}, \bibinfo{person}{Andrew
  Zisserman}, {and} \bibinfo{person}{Volodymyr Mnih}.}
  \bibinfo{year}{2019}\natexlab{}.
\newblock \showarticletitle{Unsupervised learning of object keypoints for
  perception and control}.
\newblock \bibinfo{journal}{\emph{Advances in neural information processing
  systems}}  \bibinfo{volume}{32} (\bibinfo{year}{2019}).
\newblock


\bibitem[\protect\citeauthoryear{Kummerfeld and Ramsey}{Kummerfeld and
  Ramsey}{2016}]%
        {kummerfeld2016causal}
\bibfield{author}{\bibinfo{person}{Erich Kummerfeld} {and}
  \bibinfo{person}{Joseph Ramsey}.} \bibinfo{year}{2016}\natexlab{}.
\newblock \showarticletitle{Causal clustering for 1-factor measurement models}.
  In \bibinfo{booktitle}{\emph{Proceedings of the 22nd ACM SIGKDD international
  conference on knowledge discovery and data mining}}.
  \bibinfo{pages}{1655--1664}.
\newblock


\bibitem[\protect\citeauthoryear{K{\"u}nzel, Sekhon, Bickel, and Yu}{K{\"u}nzel
  et~al\mbox{.}}{2019}]%
        {kunzel2019metalearners}
\bibfield{author}{\bibinfo{person}{S{\"o}ren~R K{\"u}nzel},
  \bibinfo{person}{Jasjeet~S Sekhon}, \bibinfo{person}{Peter~J Bickel}, {and}
  \bibinfo{person}{Bin Yu}.} \bibinfo{year}{2019}\natexlab{}.
\newblock \showarticletitle{Metalearners for estimating heterogeneous treatment
  effects using machine learning}.
\newblock \bibinfo{journal}{\emph{Proceedings of the national academy of
  sciences}} \bibinfo{volume}{116}, \bibinfo{number}{10}
  (\bibinfo{year}{2019}), \bibinfo{pages}{4156--4165}.
\newblock


\bibitem[\protect\citeauthoryear{Kyono, Zhang, and van~der Schaar}{Kyono
  et~al\mbox{.}}{2020}]%
        {kyono2020castle}
\bibfield{author}{\bibinfo{person}{Trent Kyono}, \bibinfo{person}{Yao Zhang},
  {and} \bibinfo{person}{Mihaela van~der Schaar}.}
  \bibinfo{year}{2020}\natexlab{}.
\newblock \showarticletitle{Castle: Regularization via auxiliary causal graph
  discovery}.
\newblock \bibinfo{journal}{\emph{Advances in Neural Information Processing
  Systems}}  \bibinfo{volume}{33} (\bibinfo{year}{2020}),
  \bibinfo{pages}{1501--1512}.
\newblock


\bibitem[\protect\citeauthoryear{Lachapelle, Brouillard, Deleu, and
  Lacoste-Julien}{Lachapelle et~al\mbox{.}}{2019}]%
        {lachapelle2019gradient}
\bibfield{author}{\bibinfo{person}{S{\'e}bastien Lachapelle},
  \bibinfo{person}{Philippe Brouillard}, \bibinfo{person}{Tristan Deleu}, {and}
  \bibinfo{person}{Simon Lacoste-Julien}.} \bibinfo{year}{2019}\natexlab{}.
\newblock \showarticletitle{Gradient-based neural dag learning}.
\newblock \bibinfo{journal}{\emph{arXiv preprint arXiv:1906.02226}}
  (\bibinfo{year}{2019}).
\newblock


\bibitem[\protect\citeauthoryear{Lally, Bagchi, Barborak, Buchanan,
  Chu-Carroll, Ferrucci, Glass, Kalyanpur, Mueller, Murdock,
  et~al\mbox{.}}{Lally et~al\mbox{.}}{2017}]%
        {lally2017watsonpaths}
\bibfield{author}{\bibinfo{person}{Adam Lally}, \bibinfo{person}{Sugato
  Bagchi}, \bibinfo{person}{Michael~A Barborak}, \bibinfo{person}{David~W
  Buchanan}, \bibinfo{person}{Jennifer Chu-Carroll}, \bibinfo{person}{David~A
  Ferrucci}, \bibinfo{person}{Michael~R Glass}, \bibinfo{person}{Aditya
  Kalyanpur}, \bibinfo{person}{Erik~T Mueller}, \bibinfo{person}{J~William
  Murdock}, {et~al\mbox{.}}} \bibinfo{year}{2017}\natexlab{}.
\newblock \showarticletitle{WatsonPaths: scenario-based question answering and
  inference over unstructured information}.
\newblock \bibinfo{journal}{\emph{AI magazine}} \bibinfo{volume}{38},
  \bibinfo{number}{2} (\bibinfo{year}{2017}), \bibinfo{pages}{59--76}.
\newblock


\bibitem[\protect\citeauthoryear{LaLonde}{LaLonde}{1986}]%
        {lalonde1986evaluating}
\bibfield{author}{\bibinfo{person}{Robert~J LaLonde}.}
  \bibinfo{year}{1986}\natexlab{}.
\newblock \showarticletitle{Evaluating the econometric evaluations of training
  programs with experimental data}.
\newblock \bibinfo{journal}{\emph{The American economic review}}
  (\bibinfo{year}{1986}), \bibinfo{pages}{604--620}.
\newblock


\bibitem[\protect\citeauthoryear{Lanne, Meitz, and Saikkonen}{Lanne
  et~al\mbox{.}}{2017}]%
        {lanne2017identification}
\bibfield{author}{\bibinfo{person}{Markku Lanne}, \bibinfo{person}{Mika Meitz},
  {and} \bibinfo{person}{Pentti Saikkonen}.} \bibinfo{year}{2017}\natexlab{}.
\newblock \showarticletitle{Identification and estimation of non-Gaussian
  structural vector autoregressions}.
\newblock \bibinfo{journal}{\emph{Journal of Econometrics}}
  \bibinfo{volume}{196}, \bibinfo{number}{2} (\bibinfo{year}{2017}),
  \bibinfo{pages}{288--304}.
\newblock


\bibitem[\protect\citeauthoryear{Larson, Mahendran, Peper, Clarke, Lee, Hill,
  Kummerfeld, Leach, Laurenzano, Tang, et~al\mbox{.}}{Larson
  et~al\mbox{.}}{2019}]%
        {larson2019evaluation}
\bibfield{author}{\bibinfo{person}{Stefan Larson}, \bibinfo{person}{Anish
  Mahendran}, \bibinfo{person}{Joseph~J Peper}, \bibinfo{person}{Christopher
  Clarke}, \bibinfo{person}{Andrew Lee}, \bibinfo{person}{Parker Hill},
  \bibinfo{person}{Jonathan~K Kummerfeld}, \bibinfo{person}{Kevin Leach},
  \bibinfo{person}{Michael~A Laurenzano}, \bibinfo{person}{Lingjia Tang},
  {et~al\mbox{.}}} \bibinfo{year}{2019}\natexlab{}.
\newblock \showarticletitle{An evaluation dataset for intent classification and
  out-of-scope prediction}.
\newblock \bibinfo{journal}{\emph{arXiv preprint arXiv:1909.02027}}
  (\bibinfo{year}{2019}).
\newblock


\bibitem[\protect\citeauthoryear{LeCun, Bengio, and Hinton}{LeCun
  et~al\mbox{.}}{2015}]%
        {lecun2015deep}
\bibfield{author}{\bibinfo{person}{Yann LeCun}, \bibinfo{person}{Yoshua
  Bengio}, {and} \bibinfo{person}{Geoffrey Hinton}.}
  \bibinfo{year}{2015}\natexlab{}.
\newblock \showarticletitle{Deep learning}.
\newblock \bibinfo{journal}{\emph{nature}} \bibinfo{volume}{521},
  \bibinfo{number}{7553} (\bibinfo{year}{2015}), \bibinfo{pages}{436--444}.
\newblock


\bibitem[\protect\citeauthoryear{Lee, Kim, Kim, Park, and Sohn}{Lee
  et~al\mbox{.}}{2019}]%
        {lee2019context}
\bibfield{author}{\bibinfo{person}{Jiyoung Lee}, \bibinfo{person}{Seungryong
  Kim}, \bibinfo{person}{Sunok Kim}, \bibinfo{person}{Jungin Park}, {and}
  \bibinfo{person}{Kwanghoon Sohn}.} \bibinfo{year}{2019}\natexlab{}.
\newblock \showarticletitle{Context-aware emotion recognition networks}. In
  \bibinfo{booktitle}{\emph{Proceedings of the IEEE/CVF international
  conference on computer vision}}. \bibinfo{pages}{10143--10152}.
\newblock


\bibitem[\protect\citeauthoryear{Lewis, Liu, Goyal, Ghazvininejad, Mohamed,
  Levy, Stoyanov, and Zettlemoyer}{Lewis et~al\mbox{.}}{2019}]%
        {lewis2019bart}
\bibfield{author}{\bibinfo{person}{Mike Lewis}, \bibinfo{person}{Yinhan Liu},
  \bibinfo{person}{Naman Goyal}, \bibinfo{person}{Marjan Ghazvininejad},
  \bibinfo{person}{Abdelrahman Mohamed}, \bibinfo{person}{Omer Levy},
  \bibinfo{person}{Ves Stoyanov}, {and} \bibinfo{person}{Luke Zettlemoyer}.}
  \bibinfo{year}{2019}\natexlab{}.
\newblock \showarticletitle{Bart: Denoising sequence-to-sequence pre-training
  for natural language generation, translation, and comprehension}.
\newblock \bibinfo{journal}{\emph{arXiv preprint arXiv:1910.13461}}
  (\bibinfo{year}{2019}).
\newblock


\bibitem[\protect\citeauthoryear{Li and Pearl}{Li and Pearl}{2022}]%
        {li2022bounds}
\bibfield{author}{\bibinfo{person}{Ang Li} {and} \bibinfo{person}{Judea
  Pearl}.} \bibinfo{year}{2022}\natexlab{}.
\newblock \showarticletitle{Bounds on causal effects and application to high
  dimensional data}. In \bibinfo{booktitle}{\emph{Proceedings of the AAAI
  Conference on Artificial Intelligence}}, Vol.~\bibinfo{volume}{36}.
  \bibinfo{pages}{5773--5780}.
\newblock


\bibitem[\protect\citeauthoryear{Li, Morgan, and Zaslavsky}{Li
  et~al\mbox{.}}{2018}]%
        {li2018balancing}
\bibfield{author}{\bibinfo{person}{Fan Li}, \bibinfo{person}{Kari~Lock Morgan},
  {and} \bibinfo{person}{Alan~M Zaslavsky}.} \bibinfo{year}{2018}\natexlab{}.
\newblock \showarticletitle{Balancing covariates via propensity score
  weighting}.
\newblock \bibinfo{journal}{\emph{J. Amer. Statist. Assoc.}}
  \bibinfo{volume}{113}, \bibinfo{number}{521} (\bibinfo{year}{2018}),
  \bibinfo{pages}{390--400}.
\newblock


\bibitem[\protect\citeauthoryear{Li, Su, Shen, Li, Cao, and Niu}{Li
  et~al\mbox{.}}{2017}]%
        {li2017dailydialog}
\bibfield{author}{\bibinfo{person}{Yanran Li}, \bibinfo{person}{Hui Su},
  \bibinfo{person}{Xiaoyu Shen}, \bibinfo{person}{Wenjie Li},
  \bibinfo{person}{Ziqiang Cao}, {and} \bibinfo{person}{Shuzi Niu}.}
  \bibinfo{year}{2017}\natexlab{}.
\newblock \showarticletitle{Dailydialog: A manually labelled multi-turn
  dialogue dataset}.
\newblock \bibinfo{journal}{\emph{arXiv preprint arXiv:1710.03957}}
  (\bibinfo{year}{2017}).
\newblock


\bibitem[\protect\citeauthoryear{Li, Li, Zou, and Ren}{Li
  et~al\mbox{.}}{2021}]%
        {li2021causality}
\bibfield{author}{\bibinfo{person}{Zhaoning Li}, \bibinfo{person}{Qi Li},
  \bibinfo{person}{Xiaotian Zou}, {and} \bibinfo{person}{Jiangtao Ren}.}
  \bibinfo{year}{2021}\natexlab{}.
\newblock \showarticletitle{Causality extraction based on self-attentive
  BiLSTM-CRF with transferred embeddings}.
\newblock \bibinfo{journal}{\emph{Neurocomputing}}  \bibinfo{volume}{423}
  (\bibinfo{year}{2021}), \bibinfo{pages}{207--219}.
\newblock


\bibitem[\protect\citeauthoryear{Lin, Yuan, Peng, and Tzeng}{Lin
  et~al\mbox{.}}{2022}]%
        {lin2022cascade}
\bibfield{author}{\bibinfo{person}{Fudong Lin}, \bibinfo{person}{Xu Yuan},
  \bibinfo{person}{Lu Peng}, {and} \bibinfo{person}{Nian-Feng Tzeng}.}
  \bibinfo{year}{2022}\natexlab{}.
\newblock \showarticletitle{Cascade Variational Auto-Encoder for Hierarchical
  Disentanglement}. In \bibinfo{booktitle}{\emph{Proceedings of the 31st ACM
  International Conference on Information \& Knowledge Management}}.
  \bibinfo{pages}{1248--1257}.
\newblock


\bibitem[\protect\citeauthoryear{Liu, He, Chen, and Gao}{Liu
  et~al\mbox{.}}{2019a}]%
        {liu2019multi}
\bibfield{author}{\bibinfo{person}{Xiaodong Liu}, \bibinfo{person}{Pengcheng
  He}, \bibinfo{person}{Weizhu Chen}, {and} \bibinfo{person}{Jianfeng Gao}.}
  \bibinfo{year}{2019}\natexlab{a}.
\newblock \showarticletitle{Multi-task deep neural networks for natural
  language understanding}.
\newblock \bibinfo{journal}{\emph{arXiv preprint arXiv:1901.11504}}
  (\bibinfo{year}{2019}).
\newblock


\bibitem[\protect\citeauthoryear{Liu, Ott, Goyal, Du, Joshi, Chen, Levy, Lewis,
  Zettlemoyer, and Stoyanov}{Liu et~al\mbox{.}}{2019b}]%
        {liu2019roberta}
\bibfield{author}{\bibinfo{person}{Yinhan Liu}, \bibinfo{person}{Myle Ott},
  \bibinfo{person}{Naman Goyal}, \bibinfo{person}{Jingfei Du},
  \bibinfo{person}{Mandar Joshi}, \bibinfo{person}{Danqi Chen},
  \bibinfo{person}{Omer Levy}, \bibinfo{person}{Mike Lewis},
  \bibinfo{person}{Luke Zettlemoyer}, {and} \bibinfo{person}{Veselin
  Stoyanov}.} \bibinfo{year}{2019}\natexlab{b}.
\newblock \showarticletitle{Roberta: A robustly optimized bert pretraining
  approach}.
\newblock \bibinfo{journal}{\emph{arXiv preprint arXiv:1907.11692}}
  (\bibinfo{year}{2019}).
\newblock


\bibitem[\protect\citeauthoryear{Liu, Zhang, Gong, Gong, Huang, Hengel, Zhang,
  and Shi}{Liu et~al\mbox{.}}{2022}]%
        {liu2022identifying}
\bibfield{author}{\bibinfo{person}{Yuhang Liu}, \bibinfo{person}{Zhen Zhang},
  \bibinfo{person}{Dong Gong}, \bibinfo{person}{Mingming Gong},
  \bibinfo{person}{Biwei Huang}, \bibinfo{person}{Anton van~den Hengel},
  \bibinfo{person}{Kun Zhang}, {and} \bibinfo{person}{Javen~Qinfeng Shi}.}
  \bibinfo{year}{2022}\natexlab{}.
\newblock \showarticletitle{Identifying weight-variant latent causal models}.
\newblock \bibinfo{journal}{\emph{arXiv preprint arXiv:2208.14153}}
  (\bibinfo{year}{2022}).
\newblock


\bibitem[\protect\citeauthoryear{Locatello, Abbati, Rainforth, Bauer,
  Sch{\"o}lkopf, and Bachem}{Locatello et~al\mbox{.}}{2019}]%
        {locatello2019fairness}
\bibfield{author}{\bibinfo{person}{Francesco Locatello},
  \bibinfo{person}{Gabriele Abbati}, \bibinfo{person}{Thomas Rainforth},
  \bibinfo{person}{Stefan Bauer}, \bibinfo{person}{Bernhard Sch{\"o}lkopf},
  {and} \bibinfo{person}{Olivier Bachem}.} \bibinfo{year}{2019}\natexlab{}.
\newblock \showarticletitle{On the fairness of disentangled representations}.
\newblock \bibinfo{journal}{\emph{Advances in neural information processing
  systems}}  \bibinfo{volume}{32} (\bibinfo{year}{2019}).
\newblock


\bibitem[\protect\citeauthoryear{Loh and B{\"u}hlmann}{Loh and
  B{\"u}hlmann}{2014}]%
        {loh2014high}
\bibfield{author}{\bibinfo{person}{Po-Ling Loh} {and} \bibinfo{person}{Peter
  B{\"u}hlmann}.} \bibinfo{year}{2014}\natexlab{}.
\newblock \showarticletitle{High-dimensional learning of linear causal networks
  via inverse covariance estimation}.
\newblock \bibinfo{journal}{\emph{The Journal of Machine Learning Research}}
  \bibinfo{volume}{15}, \bibinfo{number}{1} (\bibinfo{year}{2014}),
  \bibinfo{pages}{3065--3105}.
\newblock


\bibitem[\protect\citeauthoryear{Lopez-Paz, Muandet, Sch{\"o}lkopf, and
  Tolstikhin}{Lopez-Paz et~al\mbox{.}}{2015}]%
        {lopez2015towards}
\bibfield{author}{\bibinfo{person}{David Lopez-Paz}, \bibinfo{person}{Krikamol
  Muandet}, \bibinfo{person}{Bernhard Sch{\"o}lkopf}, {and}
  \bibinfo{person}{Iliya Tolstikhin}.} \bibinfo{year}{2015}\natexlab{}.
\newblock \showarticletitle{Towards a learning theory of cause-effect
  inference}. In \bibinfo{booktitle}{\emph{International Conference on Machine
  Learning}}. PMLR, \bibinfo{pages}{1452--1461}.
\newblock


\bibitem[\protect\citeauthoryear{Lopez-Paz, Nishihara, Chintala, Scholkopf, and
  Bottou}{Lopez-Paz et~al\mbox{.}}{2017}]%
        {lopez2017discovering}
\bibfield{author}{\bibinfo{person}{David Lopez-Paz}, \bibinfo{person}{Robert
  Nishihara}, \bibinfo{person}{Soumith Chintala}, \bibinfo{person}{Bernhard
  Scholkopf}, {and} \bibinfo{person}{L{\'e}on Bottou}.}
  \bibinfo{year}{2017}\natexlab{}.
\newblock \showarticletitle{Discovering causal signals in images}. In
  \bibinfo{booktitle}{\emph{Proceedings of the IEEE conference on computer
  vision and pattern recognition}}. \bibinfo{pages}{6979--6987}.
\newblock


\bibitem[\protect\citeauthoryear{L{\"o}we, Madras, Zemel, and Welling}{L{\"o}we
  et~al\mbox{.}}{2022}]%
        {lowe2022amortized}
\bibfield{author}{\bibinfo{person}{Sindy L{\"o}we}, \bibinfo{person}{David
  Madras}, \bibinfo{person}{Richard Zemel}, {and} \bibinfo{person}{Max
  Welling}.} \bibinfo{year}{2022}\natexlab{}.
\newblock \showarticletitle{Amortized causal discovery: Learning to infer
  causal graphs from time-series data}. In \bibinfo{booktitle}{\emph{Conference
  on Causal Learning and Reasoning}}. PMLR, \bibinfo{pages}{509--525}.
\newblock


\bibitem[\protect\citeauthoryear{Luca}{Luca}{2016}]%
        {luca2016reviews}
\bibfield{author}{\bibinfo{person}{Michael Luca}.}
  \bibinfo{year}{2016}\natexlab{}.
\newblock \showarticletitle{Reviews, reputation, and revenue: The case of Yelp.
  com}.
\newblock \bibinfo{journal}{\emph{Com (March 15, 2016). Harvard Business School
  NOM Unit Working Paper}} \bibinfo{number}{12-016} (\bibinfo{year}{2016}).
\newblock


\bibitem[\protect\citeauthoryear{Luo, Xu, Zha, Du, Xie, Yang, and Zhang}{Luo
  et~al\mbox{.}}{2014}]%
        {luo2014you}
\bibfield{author}{\bibinfo{person}{Dixin Luo}, \bibinfo{person}{Hongteng Xu},
  \bibinfo{person}{Hongyuan Zha}, \bibinfo{person}{Jun Du},
  \bibinfo{person}{Rong Xie}, \bibinfo{person}{Xiaokang Yang}, {and}
  \bibinfo{person}{Wenjun Zhang}.} \bibinfo{year}{2014}\natexlab{}.
\newblock \showarticletitle{You are what you watch and when you watch:
  Inferring household structures from IPTV viewing data}.
\newblock \bibinfo{journal}{\emph{IEEE Transactions on Broadcasting}}
  \bibinfo{volume}{60}, \bibinfo{number}{1} (\bibinfo{year}{2014}),
  \bibinfo{pages}{61--72}.
\newblock


\bibitem[\protect\citeauthoryear{Ma and Wang}{Ma and Wang}{2020}]%
        {ma2020robust}
\bibfield{author}{\bibinfo{person}{Xinwei Ma} {and} \bibinfo{person}{Jingshen
  Wang}.} \bibinfo{year}{2020}\natexlab{}.
\newblock \showarticletitle{Robust inference using inverse probability
  weighting}.
\newblock \bibinfo{journal}{\emph{J. Amer. Statist. Assoc.}}
  \bibinfo{volume}{115}, \bibinfo{number}{532} (\bibinfo{year}{2020}),
  \bibinfo{pages}{1851--1860}.
\newblock


\bibitem[\protect\citeauthoryear{Majumder, Hong, Peng, Lu, Ghosal, Gelbukh,
  Mihalcea, and Poria}{Majumder et~al\mbox{.}}{2020}]%
        {majumder2020mime}
\bibfield{author}{\bibinfo{person}{Navonil Majumder}, \bibinfo{person}{Pengfei
  Hong}, \bibinfo{person}{Shanshan Peng}, \bibinfo{person}{Jiankun Lu},
  \bibinfo{person}{Deepanway Ghosal}, \bibinfo{person}{Alexander Gelbukh},
  \bibinfo{person}{Rada Mihalcea}, {and} \bibinfo{person}{Soujanya Poria}.}
  \bibinfo{year}{2020}\natexlab{}.
\newblock \showarticletitle{MIME: MIMicking emotions for empathetic response
  generation}.
\newblock \bibinfo{journal}{\emph{arXiv preprint arXiv:2010.01454}}
  (\bibinfo{year}{2020}).
\newblock


\bibitem[\protect\citeauthoryear{Maliniak, Powers, and Walter}{Maliniak
  et~al\mbox{.}}{2013}]%
        {maliniak2013gender}
\bibfield{author}{\bibinfo{person}{Daniel Maliniak}, \bibinfo{person}{Ryan
  Powers}, {and} \bibinfo{person}{Barbara~F Walter}.}
  \bibinfo{year}{2013}\natexlab{}.
\newblock \showarticletitle{The gender citation gap in international
  relations}.
\newblock \bibinfo{journal}{\emph{International Organization}}
  \bibinfo{volume}{67}, \bibinfo{number}{4} (\bibinfo{year}{2013}),
  \bibinfo{pages}{889--922}.
\newblock


\bibitem[\protect\citeauthoryear{Mao, Xia, Wang, Wang, Yang, Bareinboim, and
  Vondrick}{Mao et~al\mbox{.}}{2022}]%
        {mao2022causal}
\bibfield{author}{\bibinfo{person}{Chengzhi Mao}, \bibinfo{person}{Kevin Xia},
  \bibinfo{person}{James Wang}, \bibinfo{person}{Hao Wang},
  \bibinfo{person}{Junfeng Yang}, \bibinfo{person}{Elias Bareinboim}, {and}
  \bibinfo{person}{Carl Vondrick}.} \bibinfo{year}{2022}\natexlab{}.
\newblock \showarticletitle{Causal transportability for visual recognition}. In
  \bibinfo{booktitle}{\emph{Proceedings of the IEEE/CVF Conference on Computer
  Vision and Pattern Recognition}}. \bibinfo{pages}{7521--7531}.
\newblock


\bibitem[\protect\citeauthoryear{Marinazzo, Pellicoro, and
  Stramaglia}{Marinazzo et~al\mbox{.}}{2008}]%
        {marinazzo2008kernel}
\bibfield{author}{\bibinfo{person}{Daniele Marinazzo}, \bibinfo{person}{Mario
  Pellicoro}, {and} \bibinfo{person}{Sebastiano Stramaglia}.}
  \bibinfo{year}{2008}\natexlab{}.
\newblock \showarticletitle{Kernel method for nonlinear Granger causality}.
\newblock \bibinfo{journal}{\emph{Physical review letters}}
  \bibinfo{volume}{100}, \bibinfo{number}{14} (\bibinfo{year}{2008}),
  \bibinfo{pages}{144103}.
\newblock


\bibitem[\protect\citeauthoryear{Mei and Moura}{Mei and Moura}{2016}]%
        {mei2016signal}
\bibfield{author}{\bibinfo{person}{Jonathan Mei} {and}
  \bibinfo{person}{Jos{\'e}~MF Moura}.} \bibinfo{year}{2016}\natexlab{}.
\newblock \showarticletitle{Signal processing on graphs: Causal modeling of
  unstructured data}.
\newblock \bibinfo{journal}{\emph{IEEE Transactions on Signal Processing}}
  \bibinfo{volume}{65}, \bibinfo{number}{8} (\bibinfo{year}{2016}),
  \bibinfo{pages}{2077--2092}.
\newblock


\bibitem[\protect\citeauthoryear{MemeTracker}{MemeTracker}{2018}]%
        {MemeTracker}
MemeTracker \bibinfo{year}{2018}\natexlab{}.
\newblock \bibinfo{title}{Stanford Network Analysis Project}.
\newblock
\newblock
\newblock
\shownote{http://memetracker.org/}.


\bibitem[\protect\citeauthoryear{Miao, Geng, and Tchetgen~Tchetgen}{Miao
  et~al\mbox{.}}{2018}]%
        {miao2018identifying}
\bibfield{author}{\bibinfo{person}{Wang Miao}, \bibinfo{person}{Zhi Geng},
  {and} \bibinfo{person}{Eric~J Tchetgen~Tchetgen}.}
  \bibinfo{year}{2018}\natexlab{}.
\newblock \showarticletitle{Identifying causal effects with proxy variables of
  an unmeasured confounder}.
\newblock \bibinfo{journal}{\emph{Biometrika}} \bibinfo{volume}{105},
  \bibinfo{number}{4} (\bibinfo{year}{2018}), \bibinfo{pages}{987--993}.
\newblock


\bibitem[\protect\citeauthoryear{Mnih, Kavukcuoglu, Silver, Rusu, Veness,
  Bellemare, Graves, Riedmiller, Fidjeland, Ostrovski, et~al\mbox{.}}{Mnih
  et~al\mbox{.}}{2015}]%
        {mnih2015human}
\bibfield{author}{\bibinfo{person}{Volodymyr Mnih}, \bibinfo{person}{Koray
  Kavukcuoglu}, \bibinfo{person}{David Silver}, \bibinfo{person}{Andrei~A
  Rusu}, \bibinfo{person}{Joel Veness}, \bibinfo{person}{Marc~G Bellemare},
  \bibinfo{person}{Alex Graves}, \bibinfo{person}{Martin Riedmiller},
  \bibinfo{person}{Andreas~K Fidjeland}, \bibinfo{person}{Georg Ostrovski},
  {et~al\mbox{.}}} \bibinfo{year}{2015}\natexlab{}.
\newblock \showarticletitle{Human-level control through deep reinforcement
  learning}.
\newblock \bibinfo{journal}{\emph{nature}} \bibinfo{volume}{518},
  \bibinfo{number}{7540} (\bibinfo{year}{2015}), \bibinfo{pages}{529--533}.
\newblock


\bibitem[\protect\citeauthoryear{Mooij, Peters, Janzing, Zscheischler, and
  Sch{\"o}lkopf}{Mooij et~al\mbox{.}}{2016}]%
        {mooij2016distinguishing}
\bibfield{author}{\bibinfo{person}{Joris~M Mooij}, \bibinfo{person}{Jonas
  Peters}, \bibinfo{person}{Dominik Janzing}, \bibinfo{person}{Jakob
  Zscheischler}, {and} \bibinfo{person}{Bernhard Sch{\"o}lkopf}.}
  \bibinfo{year}{2016}\natexlab{}.
\newblock \showarticletitle{Distinguishing cause from effect using
  observational data: methods and benchmarks}.
\newblock \bibinfo{journal}{\emph{The Journal of Machine Learning Research}}
  \bibinfo{volume}{17}, \bibinfo{number}{1} (\bibinfo{year}{2016}),
  \bibinfo{pages}{1103--1204}.
\newblock


\bibitem[\protect\citeauthoryear{Moraffah, Moraffah, Karami, Raglin, and
  Liu}{Moraffah et~al\mbox{.}}{2020}]%
        {moraffah2020causal}
\bibfield{author}{\bibinfo{person}{Raha Moraffah}, \bibinfo{person}{Bahman
  Moraffah}, \bibinfo{person}{Mansooreh Karami}, \bibinfo{person}{Adrienne
  Raglin}, {and} \bibinfo{person}{Huan Liu}.} \bibinfo{year}{2020}\natexlab{}.
\newblock \showarticletitle{Causal adversarial network for learning conditional
  and interventional distributions}.
\newblock \bibinfo{journal}{\emph{arXiv preprint arXiv:2008.11376}}
  (\bibinfo{year}{2020}).
\newblock


\bibitem[\protect\citeauthoryear{Moraffah, Sheth, Karami, Bhattacharya, Wang,
  Tahir, Raglin, and Liu}{Moraffah et~al\mbox{.}}{2021}]%
        {moraffah2021causal}
\bibfield{author}{\bibinfo{person}{Raha Moraffah}, \bibinfo{person}{Paras
  Sheth}, \bibinfo{person}{Mansooreh Karami}, \bibinfo{person}{Anchit
  Bhattacharya}, \bibinfo{person}{Qianru Wang}, \bibinfo{person}{Anique Tahir},
  \bibinfo{person}{Adrienne Raglin}, {and} \bibinfo{person}{Huan Liu}.}
  \bibinfo{year}{2021}\natexlab{}.
\newblock \showarticletitle{Causal inference for time series analysis:
  Problems, methods and evaluation}.
\newblock \bibinfo{journal}{\emph{Knowledge and Information Systems}}
  \bibinfo{volume}{63} (\bibinfo{year}{2021}), \bibinfo{pages}{3041--3085}.
\newblock


\bibitem[\protect\citeauthoryear{Mudambi and Schuff}{Mudambi and
  Schuff}{2010}]%
        {mudambi2010research}
\bibfield{author}{\bibinfo{person}{Susan~M Mudambi} {and}
  \bibinfo{person}{David Schuff}.} \bibinfo{year}{2010}\natexlab{}.
\newblock \showarticletitle{Research note: What makes a helpful online review?
  A study of customer reviews on Amazon. com}.
\newblock \bibinfo{journal}{\emph{MIS quarterly}} (\bibinfo{year}{2010}),
  \bibinfo{pages}{185--200}.
\newblock


\bibitem[\protect\citeauthoryear{Nash, Sellers, Talbot, Cawthorn, and
  Ford}{Nash et~al\mbox{.}}{1994}]%
        {nash1994population}
\bibfield{author}{\bibinfo{person}{Warwick~J Nash}, \bibinfo{person}{Tracy~L
  Sellers}, \bibinfo{person}{Simon~R Talbot}, \bibinfo{person}{Andrew~J
  Cawthorn}, {and} \bibinfo{person}{Wes~B Ford}.}
  \bibinfo{year}{1994}\natexlab{}.
\newblock \showarticletitle{The population biology of abalone (haliotis
  species) in tasmania. i. blacklip abalone (h. rubra) from the north coast and
  islands of bass strait}.
\newblock \bibinfo{journal}{\emph{Sea Fisheries Division, Technical Report}}
  \bibinfo{volume}{48} (\bibinfo{year}{1994}), \bibinfo{pages}{p411}.
\newblock


\bibitem[\protect\citeauthoryear{Ng, Zhu, Chen, and Fang}{Ng
  et~al\mbox{.}}{2019}]%
        {ng2019graph}
\bibfield{author}{\bibinfo{person}{Ignavier Ng}, \bibinfo{person}{Shengyu Zhu},
  \bibinfo{person}{Zhitang Chen}, {and} \bibinfo{person}{Zhuangyan Fang}.}
  \bibinfo{year}{2019}\natexlab{}.
\newblock \showarticletitle{A graph autoencoder approach to causal structure
  learning}.
\newblock \bibinfo{journal}{\emph{arXiv preprint arXiv:1911.07420}}
  (\bibinfo{year}{2019}).
\newblock


\bibitem[\protect\citeauthoryear{Nicolson and Paliwal}{Nicolson and
  Paliwal}{2020}]%
        {nicolson2020masked}
\bibfield{author}{\bibinfo{person}{Aaron Nicolson} {and}
  \bibinfo{person}{Kuldip~K Paliwal}.} \bibinfo{year}{2020}\natexlab{}.
\newblock \showarticletitle{Masked multi-head self-attention for causal speech
  enhancement}.
\newblock \bibinfo{journal}{\emph{Speech Communication}}  \bibinfo{volume}{125}
  (\bibinfo{year}{2020}), \bibinfo{pages}{80--96}.
\newblock


\bibitem[\protect\citeauthoryear{Oh, Jeong, and Lim}{Oh et~al\mbox{.}}{2021}]%
        {oh2021causal}
\bibfield{author}{\bibinfo{person}{Geesung Oh}, \bibinfo{person}{Euiseok
  Jeong}, {and} \bibinfo{person}{Sejoon Lim}.} \bibinfo{year}{2021}\natexlab{}.
\newblock \showarticletitle{Causal affect prediction model using a past facial
  image sequence}. In \bibinfo{booktitle}{\emph{Proceedings of the IEEE/CVF
  International Conference on Computer Vision}}. \bibinfo{pages}{3550--3556}.
\newblock


\bibitem[\protect\citeauthoryear{OpenAI}{OpenAI}{2023}]%
        {openai2023gpt4}
\bibfield{author}{\bibinfo{person}{OpenAI}.} \bibinfo{year}{2023}\natexlab{}.
\newblock \bibinfo{title}{GPT-4 Technical Report}.
\newblock
\newblock
\showeprint[arxiv]{2303.08774}~[cs.CL]


\bibitem[\protect\citeauthoryear{Ouyang, Wu, Jiang, Almeida, Wainwright,
  Mishkin, Zhang, Agarwal, Slama, Ray, et~al\mbox{.}}{Ouyang
  et~al\mbox{.}}{2022}]%
        {ouyang2022training}
\bibfield{author}{\bibinfo{person}{Long Ouyang}, \bibinfo{person}{Jeffrey Wu},
  \bibinfo{person}{Xu Jiang}, \bibinfo{person}{Diogo Almeida},
  \bibinfo{person}{Carroll Wainwright}, \bibinfo{person}{Pamela Mishkin},
  \bibinfo{person}{Chong Zhang}, \bibinfo{person}{Sandhini Agarwal},
  \bibinfo{person}{Katarina Slama}, \bibinfo{person}{Alex Ray},
  {et~al\mbox{.}}} \bibinfo{year}{2022}\natexlab{}.
\newblock \showarticletitle{Training language models to follow instructions
  with human feedback}.
\newblock \bibinfo{journal}{\emph{Advances in Neural Information Processing
  Systems}}  \bibinfo{volume}{35} (\bibinfo{year}{2022}),
  \bibinfo{pages}{27730--27744}.
\newblock


\bibitem[\protect\citeauthoryear{Pamfil, Sriwattanaworachai, Desai,
  Pilgerstorfer, Georgatzis, Beaumont, and Aragam}{Pamfil
  et~al\mbox{.}}{2020}]%
        {pamfil2020dynotears}
\bibfield{author}{\bibinfo{person}{Roxana Pamfil}, \bibinfo{person}{Nisara
  Sriwattanaworachai}, \bibinfo{person}{Shaan Desai}, \bibinfo{person}{Philip
  Pilgerstorfer}, \bibinfo{person}{Konstantinos Georgatzis},
  \bibinfo{person}{Paul Beaumont}, {and} \bibinfo{person}{Bryon Aragam}.}
  \bibinfo{year}{2020}\natexlab{}.
\newblock \showarticletitle{Dynotears: Structure learning from time-series
  data}. In \bibinfo{booktitle}{\emph{International Conference on Artificial
  Intelligence and Statistics}}. PMLR, \bibinfo{pages}{1595--1605}.
\newblock


\bibitem[\protect\citeauthoryear{Pearl}{Pearl}{2010}]%
        {pearl2010causal}
\bibfield{author}{\bibinfo{person}{Judea Pearl}.}
  \bibinfo{year}{2010}\natexlab{}.
\newblock \showarticletitle{Causal inference}.
\newblock \bibinfo{journal}{\emph{Causality: objectives and assessment}}
  (\bibinfo{year}{2010}), \bibinfo{pages}{39--58}.
\newblock


\bibitem[\protect\citeauthoryear{Pearl et~al\mbox{.}}{Pearl
  et~al\mbox{.}}{2000}]%
        {pearl2000models}
\bibfield{author}{\bibinfo{person}{Judea Pearl} {et~al\mbox{.}}}
  \bibinfo{year}{2000}\natexlab{}.
\newblock \showarticletitle{Models, reasoning and inference}.
\newblock \bibinfo{journal}{\emph{Cambridge, UK: CambridgeUniversityPress}}
  \bibinfo{volume}{19}, \bibinfo{number}{2} (\bibinfo{year}{2000}).
\newblock


\bibitem[\protect\citeauthoryear{Pena, Bj{\"o}rkegren, and Tegn{\'e}r}{Pena
  et~al\mbox{.}}{2005}]%
        {pena2005learning}
\bibfield{author}{\bibinfo{person}{Jose~M Pena}, \bibinfo{person}{Johan
  Bj{\"o}rkegren}, {and} \bibinfo{person}{Jesper Tegn{\'e}r}.}
  \bibinfo{year}{2005}\natexlab{}.
\newblock \showarticletitle{Learning dynamic Bayesian network models via
  cross-validation}.
\newblock \bibinfo{journal}{\emph{Pattern Recognition Letters}}
  \bibinfo{volume}{26}, \bibinfo{number}{14} (\bibinfo{year}{2005}),
  \bibinfo{pages}{2295--2308}.
\newblock


\bibitem[\protect\citeauthoryear{Peters, Janzing, and Sch{\"o}lkopf}{Peters
  et~al\mbox{.}}{2013}]%
        {peters2013causal}
\bibfield{author}{\bibinfo{person}{Jonas Peters}, \bibinfo{person}{Dominik
  Janzing}, {and} \bibinfo{person}{Bernhard Sch{\"o}lkopf}.}
  \bibinfo{year}{2013}\natexlab{}.
\newblock \showarticletitle{Causal inference on time series using restricted
  structural equation models}.
\newblock \bibinfo{journal}{\emph{Advances in neural information processing
  systems}}  \bibinfo{volume}{26} (\bibinfo{year}{2013}).
\newblock


\bibitem[\protect\citeauthoryear{Peters, Janzing, and Sch{\"o}lkopf}{Peters
  et~al\mbox{.}}{2017}]%
        {peters2017elements}
\bibfield{author}{\bibinfo{person}{Jonas Peters}, \bibinfo{person}{Dominik
  Janzing}, {and} \bibinfo{person}{Bernhard Sch{\"o}lkopf}.}
  \bibinfo{year}{2017}\natexlab{}.
\newblock \bibinfo{booktitle}{\emph{Elements of causal inference: foundations
  and learning algorithms}}.
\newblock \bibinfo{publisher}{The MIT Press}.
\newblock


\bibitem[\protect\citeauthoryear{Peters, Mooij, Janzing, and
  Sch{\"o}lkopf}{Peters et~al\mbox{.}}{2012}]%
        {peters2012identifiability}
\bibfield{author}{\bibinfo{person}{Jonas Peters}, \bibinfo{person}{Joris
  Mooij}, \bibinfo{person}{Dominik Janzing}, {and} \bibinfo{person}{Bernhard
  Sch{\"o}lkopf}.} \bibinfo{year}{2012}\natexlab{}.
\newblock \showarticletitle{Identifiability of causal graphs using functional
  models}.
\newblock \bibinfo{journal}{\emph{arXiv preprint arXiv:1202.3757}}
  (\bibinfo{year}{2012}).
\newblock


\bibitem[\protect\citeauthoryear{Peters, Mooij, Janzing, and
  Sch{\"o}lkopf}{Peters et~al\mbox{.}}{2014}]%
        {peters2014causal}
\bibfield{author}{\bibinfo{person}{Jonas Peters}, \bibinfo{person}{Joris~M
  Mooij}, \bibinfo{person}{Dominik Janzing}, {and} \bibinfo{person}{Bernhard
  Sch{\"o}lkopf}.} \bibinfo{year}{2014}\natexlab{}.
\newblock \showarticletitle{Causal discovery with continuous additive noise
  models}.
\newblock  (\bibinfo{year}{2014}).
\newblock


\bibitem[\protect\citeauthoryear{Poppe}{Poppe}{2010}]%
        {poppe2010survey}
\bibfield{author}{\bibinfo{person}{Ronald Poppe}.}
  \bibinfo{year}{2010}\natexlab{}.
\newblock \showarticletitle{A survey on vision-based human action recognition}.
\newblock \bibinfo{journal}{\emph{Image and vision computing}}
  \bibinfo{volume}{28}, \bibinfo{number}{6} (\bibinfo{year}{2010}),
  \bibinfo{pages}{976--990}.
\newblock


\bibitem[\protect\citeauthoryear{Poria, Majumder, Hazarika, Ghosal, Bhardwaj,
  Jian, Hong, Ghosh, Roy, Chhaya, et~al\mbox{.}}{Poria et~al\mbox{.}}{2021}]%
        {poria2021recognizing}
\bibfield{author}{\bibinfo{person}{Soujanya Poria}, \bibinfo{person}{Navonil
  Majumder}, \bibinfo{person}{Devamanyu Hazarika}, \bibinfo{person}{Deepanway
  Ghosal}, \bibinfo{person}{Rishabh Bhardwaj}, \bibinfo{person}{Samson Yu~Bai
  Jian}, \bibinfo{person}{Pengfei Hong}, \bibinfo{person}{Romila Ghosh},
  \bibinfo{person}{Abhinaba Roy}, \bibinfo{person}{Niyati Chhaya},
  {et~al\mbox{.}}} \bibinfo{year}{2021}\natexlab{}.
\newblock \showarticletitle{Recognizing emotion cause in conversations}.
\newblock \bibinfo{journal}{\emph{Cognitive Computation}}  \bibinfo{volume}{13}
  (\bibinfo{year}{2021}), \bibinfo{pages}{1317--1332}.
\newblock


\bibitem[\protect\citeauthoryear{Prill, Marbach, Saez-Rodriguez, Sorger,
  Alexopoulos, Xue, Clarke, Altan-Bonnet, and Stolovitzky}{Prill
  et~al\mbox{.}}{2010}]%
        {prill2010towards}
\bibfield{author}{\bibinfo{person}{Robert~J Prill}, \bibinfo{person}{Daniel
  Marbach}, \bibinfo{person}{Julio Saez-Rodriguez}, \bibinfo{person}{Peter~K
  Sorger}, \bibinfo{person}{Leonidas~G Alexopoulos}, \bibinfo{person}{Xiaowei
  Xue}, \bibinfo{person}{Neil~D Clarke}, \bibinfo{person}{Gregoire
  Altan-Bonnet}, {and} \bibinfo{person}{Gustavo Stolovitzky}.}
  \bibinfo{year}{2010}\natexlab{}.
\newblock \showarticletitle{Towards a rigorous assessment of systems biology
  models: the DREAM3 challenges}.
\newblock \bibinfo{journal}{\emph{PloS one}} \bibinfo{volume}{5},
  \bibinfo{number}{2} (\bibinfo{year}{2010}), \bibinfo{pages}{e9202}.
\newblock


\bibitem[\protect\citeauthoryear{Qian, Alaa, van~der Schaar, and Ercole}{Qian
  et~al\mbox{.}}{2020}]%
        {qian2020between}
\bibfield{author}{\bibinfo{person}{Zhaozhi Qian}, \bibinfo{person}{Ahmed~M
  Alaa}, \bibinfo{person}{Mihaela van~der Schaar}, {and} \bibinfo{person}{Ari
  Ercole}.} \bibinfo{year}{2020}\natexlab{}.
\newblock \showarticletitle{Between-centre differences for COVID-19 ICU
  mortality from early data in England}.
\newblock \bibinfo{journal}{\emph{Intensive care medicine}}
  \bibinfo{volume}{46} (\bibinfo{year}{2020}), \bibinfo{pages}{1779--1780}.
\newblock


\bibitem[\protect\citeauthoryear{Radford, Wu, Child, Luan, Amodei, Sutskever,
  et~al\mbox{.}}{Radford et~al\mbox{.}}{2019}]%
        {radford2019language}
\bibfield{author}{\bibinfo{person}{Alec Radford}, \bibinfo{person}{Jeffrey Wu},
  \bibinfo{person}{Rewon Child}, \bibinfo{person}{David Luan},
  \bibinfo{person}{Dario Amodei}, \bibinfo{person}{Ilya Sutskever},
  {et~al\mbox{.}}} \bibinfo{year}{2019}\natexlab{}.
\newblock \showarticletitle{Language models are unsupervised multitask
  learners}.
\newblock \bibinfo{journal}{\emph{OpenAI blog}} \bibinfo{volume}{1},
  \bibinfo{number}{8} (\bibinfo{year}{2019}), \bibinfo{pages}{9}.
\newblock


\bibitem[\protect\citeauthoryear{Ramsey, Glymour, Sanchez-Romero, and
  Glymour}{Ramsey et~al\mbox{.}}{2017}]%
        {ramsey2017million}
\bibfield{author}{\bibinfo{person}{Joseph Ramsey}, \bibinfo{person}{Madelyn
  Glymour}, \bibinfo{person}{Ruben Sanchez-Romero}, {and}
  \bibinfo{person}{Clark Glymour}.} \bibinfo{year}{2017}\natexlab{}.
\newblock \showarticletitle{A million variables and more: the fast greedy
  equivalence search algorithm for learning high-dimensional graphical causal
  models, with an application to functional magnetic resonance images}.
\newblock \bibinfo{journal}{\emph{International journal of data science and
  analytics}}  \bibinfo{volume}{3} (\bibinfo{year}{2017}),
  \bibinfo{pages}{121--129}.
\newblock


\bibitem[\protect\citeauthoryear{Rezende and Mohamed}{Rezende and
  Mohamed}{2015}]%
        {rezende2015variational}
\bibfield{author}{\bibinfo{person}{Danilo Rezende} {and}
  \bibinfo{person}{Shakir Mohamed}.} \bibinfo{year}{2015}\natexlab{}.
\newblock \showarticletitle{Variational inference with normalizing flows}. In
  \bibinfo{booktitle}{\emph{International conference on machine learning}}.
  PMLR, \bibinfo{pages}{1530--1538}.
\newblock


\bibitem[\protect\citeauthoryear{Ristanoski, Liu, and Bailey}{Ristanoski
  et~al\mbox{.}}{2013}]%
        {ristanoski2013discrimination}
\bibfield{author}{\bibinfo{person}{Goce Ristanoski}, \bibinfo{person}{Wei Liu},
  {and} \bibinfo{person}{James Bailey}.} \bibinfo{year}{2013}\natexlab{}.
\newblock \showarticletitle{Discrimination aware classification for imbalanced
  datasets}. In \bibinfo{booktitle}{\emph{Proceedings of the 22nd ACM
  international conference on Information \& Knowledge Management}}.
  \bibinfo{pages}{1529--1532}.
\newblock


\bibitem[\protect\citeauthoryear{Rom{\'a}n, Pertusa, and
  Calvo-Zaragoza}{Rom{\'a}n et~al\mbox{.}}{2019}]%
        {roman2019holistic}
\bibfield{author}{\bibinfo{person}{Miguel~A Rom{\'a}n},
  \bibinfo{person}{Antonio Pertusa}, {and} \bibinfo{person}{Jorge
  Calvo-Zaragoza}.} \bibinfo{year}{2019}\natexlab{}.
\newblock \showarticletitle{A holistic approach to polyphonic music
  transcription with neural networks}.
\newblock \bibinfo{journal}{\emph{arXiv preprint arXiv:1910.12086}}
  (\bibinfo{year}{2019}).
\newblock


\bibitem[\protect\citeauthoryear{Rosenbaum and Rubin}{Rosenbaum and
  Rubin}{1983}]%
        {rosenbaum1983central}
\bibfield{author}{\bibinfo{person}{Paul~R Rosenbaum} {and}
  \bibinfo{person}{Donald~B Rubin}.} \bibinfo{year}{1983}\natexlab{}.
\newblock \showarticletitle{The central role of the propensity score in
  observational studies for causal effects}.
\newblock \bibinfo{journal}{\emph{Biometrika}} \bibinfo{volume}{70},
  \bibinfo{number}{1} (\bibinfo{year}{1983}), \bibinfo{pages}{41--55}.
\newblock


\bibitem[\protect\citeauthoryear{Rosenbaum and Rubin}{Rosenbaum and
  Rubin}{1985}]%
        {rosenbaum1985constructing}
\bibfield{author}{\bibinfo{person}{Paul~R Rosenbaum} {and}
  \bibinfo{person}{Donald~B Rubin}.} \bibinfo{year}{1985}\natexlab{}.
\newblock \showarticletitle{Constructing a control group using multivariate
  matched sampling methods that incorporate the propensity score}.
\newblock \bibinfo{journal}{\emph{The American Statistician}}
  \bibinfo{volume}{39}, \bibinfo{number}{1} (\bibinfo{year}{1985}),
  \bibinfo{pages}{33--38}.
\newblock


\bibitem[\protect\citeauthoryear{Roy and Roth}{Roy and Roth}{2016}]%
        {roy2016solving}
\bibfield{author}{\bibinfo{person}{Subhro Roy} {and} \bibinfo{person}{Dan
  Roth}.} \bibinfo{year}{2016}\natexlab{}.
\newblock \showarticletitle{Solving general arithmetic word problems}.
\newblock \bibinfo{journal}{\emph{arXiv preprint arXiv:1608.01413}}
  (\bibinfo{year}{2016}).
\newblock


\bibitem[\protect\citeauthoryear{Rubenstein, Weichwald, Bongers, Mooij,
  Janzing, Grosse-Wentrup, and Sch{\"o}lkopf}{Rubenstein et~al\mbox{.}}{2017}]%
        {rubenstein2017causal}
\bibfield{author}{\bibinfo{person}{Paul~K Rubenstein},
  \bibinfo{person}{Sebastian Weichwald}, \bibinfo{person}{Stephan Bongers},
  \bibinfo{person}{Joris~M Mooij}, \bibinfo{person}{Dominik Janzing},
  \bibinfo{person}{Moritz Grosse-Wentrup}, {and} \bibinfo{person}{Bernhard
  Sch{\"o}lkopf}.} \bibinfo{year}{2017}\natexlab{}.
\newblock \showarticletitle{Causal consistency of structural equation models}.
\newblock \bibinfo{journal}{\emph{arXiv preprint arXiv:1707.00819}}
  (\bibinfo{year}{2017}).
\newblock


\bibitem[\protect\citeauthoryear{Runge}{Runge}{2020}]%
        {runge2020discovering}
\bibfield{author}{\bibinfo{person}{Jakob Runge}.}
  \bibinfo{year}{2020}\natexlab{}.
\newblock \showarticletitle{Discovering contemporaneous and lagged causal
  relations in autocorrelated nonlinear time series datasets}. In
  \bibinfo{booktitle}{\emph{Conference on Uncertainty in Artificial
  Intelligence}}. PMLR, \bibinfo{pages}{1388--1397}.
\newblock


\bibitem[\protect\citeauthoryear{Runge, Nowack, Kretschmer, Flaxman, and
  Sejdinovic}{Runge et~al\mbox{.}}{2019}]%
        {runge2019detecting}
\bibfield{author}{\bibinfo{person}{Jakob Runge}, \bibinfo{person}{Peer Nowack},
  \bibinfo{person}{Marlene Kretschmer}, \bibinfo{person}{Seth Flaxman}, {and}
  \bibinfo{person}{Dino Sejdinovic}.} \bibinfo{year}{2019}\natexlab{}.
\newblock \showarticletitle{Detecting and quantifying causal associations in
  large nonlinear time series datasets}.
\newblock \bibinfo{journal}{\emph{Science advances}} \bibinfo{volume}{5},
  \bibinfo{number}{11} (\bibinfo{year}{2019}), \bibinfo{pages}{eaau4996}.
\newblock


\bibitem[\protect\citeauthoryear{Sachs, Perez, Pe'er, Lauffenburger, and
  Nolan}{Sachs et~al\mbox{.}}{2005}]%
        {sachs2005causal}
\bibfield{author}{\bibinfo{person}{Karen Sachs}, \bibinfo{person}{Omar Perez},
  \bibinfo{person}{Dana Pe'er}, \bibinfo{person}{Douglas~A Lauffenburger},
  {and} \bibinfo{person}{Garry~P Nolan}.} \bibinfo{year}{2005}\natexlab{}.
\newblock \showarticletitle{Causal protein-signaling networks derived from
  multiparameter single-cell data}.
\newblock \bibinfo{journal}{\emph{Science}} \bibinfo{volume}{308},
  \bibinfo{number}{5721} (\bibinfo{year}{2005}), \bibinfo{pages}{523--529}.
\newblock


\bibitem[\protect\citeauthoryear{Salehkaleybar, Ghassami, Kiyavash, and
  Zhang}{Salehkaleybar et~al\mbox{.}}{2020}]%
        {salehkaleybar2020learning}
\bibfield{author}{\bibinfo{person}{Saber Salehkaleybar},
  \bibinfo{person}{AmirEmad Ghassami}, \bibinfo{person}{Negar Kiyavash}, {and}
  \bibinfo{person}{Kun Zhang}.} \bibinfo{year}{2020}\natexlab{}.
\newblock \showarticletitle{Learning linear non-Gaussian causal models in the
  presence of latent variables}.
\newblock \bibinfo{journal}{\emph{The Journal of Machine Learning Research}}
  \bibinfo{volume}{21}, \bibinfo{number}{1} (\bibinfo{year}{2020}),
  \bibinfo{pages}{1436--1459}.
\newblock


\bibitem[\protect\citeauthoryear{Sanchez, Voisey, Xia, Watson, O’Neil, and
  Tsaftaris}{Sanchez et~al\mbox{.}}{2022}]%
        {sanchez2022causal}
\bibfield{author}{\bibinfo{person}{Pedro Sanchez}, \bibinfo{person}{Jeremy~P
  Voisey}, \bibinfo{person}{Tian Xia}, \bibinfo{person}{Hannah~I Watson},
  \bibinfo{person}{Alison~Q O’Neil}, {and} \bibinfo{person}{Sotirios~A
  Tsaftaris}.} \bibinfo{year}{2022}\natexlab{}.
\newblock \showarticletitle{Causal machine learning for healthcare and
  precision medicine}.
\newblock \bibinfo{journal}{\emph{Royal Society Open Science}}
  \bibinfo{volume}{9}, \bibinfo{number}{8} (\bibinfo{year}{2022}),
  \bibinfo{pages}{220638}.
\newblock


\bibitem[\protect\citeauthoryear{Sanchez-Gonzalez, Godwin, Pfaff, Ying,
  Leskovec, and Battaglia}{Sanchez-Gonzalez et~al\mbox{.}}{2020}]%
        {sanchez2020learning}
\bibfield{author}{\bibinfo{person}{Alvaro Sanchez-Gonzalez},
  \bibinfo{person}{Jonathan Godwin}, \bibinfo{person}{Tobias Pfaff},
  \bibinfo{person}{Rex Ying}, \bibinfo{person}{Jure Leskovec}, {and}
  \bibinfo{person}{Peter Battaglia}.} \bibinfo{year}{2020}\natexlab{}.
\newblock \showarticletitle{Learning to simulate complex physics with graph
  networks}. In \bibinfo{booktitle}{\emph{International conference on machine
  learning}}. PMLR, \bibinfo{pages}{8459--8468}.
\newblock


\bibitem[\protect\citeauthoryear{Sanh, Debut, Chaumond, and Wolf}{Sanh
  et~al\mbox{.}}{2019}]%
        {sanh2019distilbert}
\bibfield{author}{\bibinfo{person}{Victor Sanh}, \bibinfo{person}{Lysandre
  Debut}, \bibinfo{person}{Julien Chaumond}, {and} \bibinfo{person}{Thomas
  Wolf}.} \bibinfo{year}{2019}\natexlab{}.
\newblock \showarticletitle{DistilBERT, a distilled version of BERT: smaller,
  faster, cheaper and lighter}.
\newblock \bibinfo{journal}{\emph{arXiv preprint arXiv:1910.01108}}
  (\bibinfo{year}{2019}).
\newblock


\bibitem[\protect\citeauthoryear{Sch{\"o}lkopf}{Sch{\"o}lkopf}{2022}]%
        {scholkopf2022causality}
\bibfield{author}{\bibinfo{person}{Bernhard Sch{\"o}lkopf}.}
  \bibinfo{year}{2022}\natexlab{}.
\newblock \showarticletitle{Causality for machine learning}.
\newblock In \bibinfo{booktitle}{\emph{Probabilistic and Causal Inference: The
  Works of Judea Pearl}}. \bibinfo{pages}{765--804}.
\newblock


\bibitem[\protect\citeauthoryear{Sch{\"o}lkopf, Janzing, Peters, Sgouritsa,
  Zhang, and Mooij}{Sch{\"o}lkopf et~al\mbox{.}}{2012}]%
        {scholkopf2012causal}
\bibfield{author}{\bibinfo{person}{Bernhard Sch{\"o}lkopf},
  \bibinfo{person}{Dominik Janzing}, \bibinfo{person}{Jonas Peters},
  \bibinfo{person}{Eleni Sgouritsa}, \bibinfo{person}{Kun Zhang}, {and}
  \bibinfo{person}{Joris Mooij}.} \bibinfo{year}{2012}\natexlab{}.
\newblock \showarticletitle{On causal and anticausal learning}.
\newblock \bibinfo{journal}{\emph{arXiv preprint arXiv:1206.6471}}
  (\bibinfo{year}{2012}).
\newblock


\bibitem[\protect\citeauthoryear{Sch{\"o}lkopf, Locatello, Bauer, Ke,
  Kalchbrenner, Goyal, and Bengio}{Sch{\"o}lkopf et~al\mbox{.}}{2021}]%
        {scholkopf2021toward}
\bibfield{author}{\bibinfo{person}{Bernhard Sch{\"o}lkopf},
  \bibinfo{person}{Francesco Locatello}, \bibinfo{person}{Stefan Bauer},
  \bibinfo{person}{Nan~Rosemary Ke}, \bibinfo{person}{Nal Kalchbrenner},
  \bibinfo{person}{Anirudh Goyal}, {and} \bibinfo{person}{Yoshua Bengio}.}
  \bibinfo{year}{2021}\natexlab{}.
\newblock \showarticletitle{Toward causal representation learning}.
\newblock \bibinfo{journal}{\emph{Proc. IEEE}} \bibinfo{volume}{109},
  \bibinfo{number}{5} (\bibinfo{year}{2021}), \bibinfo{pages}{612--634}.
\newblock


\bibitem[\protect\citeauthoryear{Schrittwieser, Antonoglou, Hubert, Simonyan,
  Sifre, Schmitt, Guez, Lockhart, Hassabis, Graepel,
  et~al\mbox{.}}{Schrittwieser et~al\mbox{.}}{2020}]%
        {schrittwieser2020mastering}
\bibfield{author}{\bibinfo{person}{Julian Schrittwieser},
  \bibinfo{person}{Ioannis Antonoglou}, \bibinfo{person}{Thomas Hubert},
  \bibinfo{person}{Karen Simonyan}, \bibinfo{person}{Laurent Sifre},
  \bibinfo{person}{Simon Schmitt}, \bibinfo{person}{Arthur Guez},
  \bibinfo{person}{Edward Lockhart}, \bibinfo{person}{Demis Hassabis},
  \bibinfo{person}{Thore Graepel}, {et~al\mbox{.}}}
  \bibinfo{year}{2020}\natexlab{}.
\newblock \showarticletitle{Mastering atari, go, chess and shogi by planning
  with a learned model}.
\newblock \bibinfo{journal}{\emph{Nature}} \bibinfo{volume}{588},
  \bibinfo{number}{7839} (\bibinfo{year}{2020}), \bibinfo{pages}{604--609}.
\newblock


\bibitem[\protect\citeauthoryear{Schwab, Linhardt, Bauer, Buhmann, and
  Karlen}{Schwab et~al\mbox{.}}{2020}]%
        {schwab2020learning}
\bibfield{author}{\bibinfo{person}{Patrick Schwab}, \bibinfo{person}{Lorenz
  Linhardt}, \bibinfo{person}{Stefan Bauer}, \bibinfo{person}{Joachim~M
  Buhmann}, {and} \bibinfo{person}{Walter Karlen}.}
  \bibinfo{year}{2020}\natexlab{}.
\newblock \showarticletitle{Learning counterfactual representations for
  estimating individual dose-response curves}. In
  \bibinfo{booktitle}{\emph{Proceedings of the AAAI Conference on Artificial
  Intelligence}}, Vol.~\bibinfo{volume}{34}. \bibinfo{pages}{5612--5619}.
\newblock


\bibitem[\protect\citeauthoryear{Seigal, Squires, and Uhler}{Seigal
  et~al\mbox{.}}{2022}]%
        {seigal2022linear}
\bibfield{author}{\bibinfo{person}{Anna Seigal}, \bibinfo{person}{Chandler
  Squires}, {and} \bibinfo{person}{Caroline Uhler}.}
  \bibinfo{year}{2022}\natexlab{}.
\newblock \showarticletitle{Linear causal disentanglement via interventions}.
\newblock \bibinfo{journal}{\emph{arXiv preprint arXiv:2211.16467}}
  (\bibinfo{year}{2022}).
\newblock


\bibitem[\protect\citeauthoryear{Shadaydeh, M{\"u}ller, Schneider, Th{\"u}mmel,
  Kessler, and Denzler}{Shadaydeh et~al\mbox{.}}{2021}]%
        {shadaydeh2021analyzing}
\bibfield{author}{\bibinfo{person}{Maha Shadaydeh}, \bibinfo{person}{Lea
  M{\"u}ller}, \bibinfo{person}{Dana Schneider}, \bibinfo{person}{Martin
  Th{\"u}mmel}, \bibinfo{person}{Thomas Kessler}, {and}
  \bibinfo{person}{Joachim Denzler}.} \bibinfo{year}{2021}\natexlab{}.
\newblock \showarticletitle{Analyzing the direction of emotional influence in
  nonverbal dyadic communication: A facial-expression study}.
\newblock \bibinfo{journal}{\emph{IEEE Access}}  \bibinfo{volume}{9}
  (\bibinfo{year}{2021}), \bibinfo{pages}{73780--73790}.
\newblock


\bibitem[\protect\citeauthoryear{Shalit, Johansson, and Sontag}{Shalit
  et~al\mbox{.}}{2017}]%
        {shalit2017estimating}
\bibfield{author}{\bibinfo{person}{Uri Shalit}, \bibinfo{person}{Fredrik~D
  Johansson}, {and} \bibinfo{person}{David Sontag}.}
  \bibinfo{year}{2017}\natexlab{}.
\newblock \showarticletitle{Estimating individual treatment effect:
  generalization bounds and algorithms}. In
  \bibinfo{booktitle}{\emph{International conference on machine learning}}.
  PMLR, \bibinfo{pages}{3076--3085}.
\newblock


\bibitem[\protect\citeauthoryear{Shimizu and Bollen}{Shimizu and
  Bollen}{2014}]%
        {shimizu2014bayesian}
\bibfield{author}{\bibinfo{person}{Shohei Shimizu} {and}
  \bibinfo{person}{Kenneth Bollen}.} \bibinfo{year}{2014}\natexlab{}.
\newblock \showarticletitle{Bayesian estimation of causal direction in acyclic
  structural equation models with individual-specific confounder variables and
  non-Gaussian distributions.}
\newblock \bibinfo{journal}{\emph{J. Mach. Learn. Res.}} \bibinfo{volume}{15},
  \bibinfo{number}{1} (\bibinfo{year}{2014}), \bibinfo{pages}{2629--2652}.
\newblock


\bibitem[\protect\citeauthoryear{Shimizu, Hoyer, Hyv{\"a}rinen, Kerminen, and
  Jordan}{Shimizu et~al\mbox{.}}{2006}]%
        {shimizu2006linear}
\bibfield{author}{\bibinfo{person}{Shohei Shimizu}, \bibinfo{person}{Patrik~O
  Hoyer}, \bibinfo{person}{Aapo Hyv{\"a}rinen}, \bibinfo{person}{Antti
  Kerminen}, {and} \bibinfo{person}{Michael Jordan}.}
  \bibinfo{year}{2006}\natexlab{}.
\newblock \showarticletitle{A linear non-Gaussian acyclic model for causal
  discovery.}
\newblock \bibinfo{journal}{\emph{Journal of Machine Learning Research}}
  \bibinfo{volume}{7}, \bibinfo{number}{10} (\bibinfo{year}{2006}).
\newblock


\bibitem[\protect\citeauthoryear{Shleifer and Rush}{Shleifer and Rush}{2020}]%
        {shleifer2020pre}
\bibfield{author}{\bibinfo{person}{Sam Shleifer} {and}
  \bibinfo{person}{Alexander~M Rush}.} \bibinfo{year}{2020}\natexlab{}.
\newblock \showarticletitle{Pre-trained summarization distillation}.
\newblock \bibinfo{journal}{\emph{arXiv preprint arXiv:2010.13002}}
  (\bibinfo{year}{2020}).
\newblock


\bibitem[\protect\citeauthoryear{Shojaie and Fox}{Shojaie and Fox}{2022}]%
        {shojaie2022granger}
\bibfield{author}{\bibinfo{person}{Ali Shojaie} {and} \bibinfo{person}{Emily~B
  Fox}.} \bibinfo{year}{2022}\natexlab{}.
\newblock \showarticletitle{Granger causality: A review and recent advances}.
\newblock \bibinfo{journal}{\emph{Annual Review of Statistics and Its
  Application}}  \bibinfo{volume}{9} (\bibinfo{year}{2022}),
  \bibinfo{pages}{289--319}.
\newblock


\bibitem[\protect\citeauthoryear{Silva, Scheines, Glymour, Spirtes, and
  Chickering}{Silva et~al\mbox{.}}{2006}]%
        {silva2006learning}
\bibfield{author}{\bibinfo{person}{Ricardo Silva}, \bibinfo{person}{Richard
  Scheines}, \bibinfo{person}{Clark Glymour}, \bibinfo{person}{Peter Spirtes},
  {and} \bibinfo{person}{David~Maxwell Chickering}.}
  \bibinfo{year}{2006}\natexlab{}.
\newblock \showarticletitle{Learning the Structure of Linear Latent Variable
  Models.}
\newblock \bibinfo{journal}{\emph{Journal of Machine Learning Research}}
  \bibinfo{volume}{7}, \bibinfo{number}{2} (\bibinfo{year}{2006}).
\newblock


\bibitem[\protect\citeauthoryear{Sindhwani, Quang, and Lozano}{Sindhwani
  et~al\mbox{.}}{2012}]%
        {sindhwani2012scalable}
\bibfield{author}{\bibinfo{person}{Vikas Sindhwani}, \bibinfo{person}{Minh~Ha
  Quang}, {and} \bibinfo{person}{Aur{\'e}lie~C Lozano}.}
  \bibinfo{year}{2012}\natexlab{}.
\newblock \showarticletitle{Scalable matrix-valued kernel learning for
  high-dimensional nonlinear multivariate regression and granger causality}.
\newblock \bibinfo{journal}{\emph{arXiv preprint arXiv:1210.4792}}
  (\bibinfo{year}{2012}).
\newblock


\bibitem[\protect\citeauthoryear{Smith, Miller, Salimi-Khorshidi, Webster,
  Beckmann, Nichols, Ramsey, and Woolrich}{Smith et~al\mbox{.}}{2011}]%
        {smith2011network}
\bibfield{author}{\bibinfo{person}{Stephen~M Smith}, \bibinfo{person}{Karla~L
  Miller}, \bibinfo{person}{Gholamreza Salimi-Khorshidi},
  \bibinfo{person}{Matthew Webster}, \bibinfo{person}{Christian~F Beckmann},
  \bibinfo{person}{Thomas~E Nichols}, \bibinfo{person}{Joseph~D Ramsey}, {and}
  \bibinfo{person}{Mark~W Woolrich}.} \bibinfo{year}{2011}\natexlab{}.
\newblock \showarticletitle{Network modelling methods for FMRI}.
\newblock \bibinfo{journal}{\emph{Neuroimage}} \bibinfo{volume}{54},
  \bibinfo{number}{2} (\bibinfo{year}{2011}), \bibinfo{pages}{875--891}.
\newblock


\bibitem[\protect\citeauthoryear{Smola and Sch{\"o}lkopf}{Smola and
  Sch{\"o}lkopf}{1998}]%
        {smola1998learning}
\bibfield{author}{\bibinfo{person}{Alex~J Smola} {and}
  \bibinfo{person}{Bernhard Sch{\"o}lkopf}.} \bibinfo{year}{1998}\natexlab{}.
\newblock \bibinfo{booktitle}{\emph{Learning with kernels}}.
  Vol.~\bibinfo{volume}{4}.
\newblock \bibinfo{publisher}{Citeseer}.
\newblock


\bibitem[\protect\citeauthoryear{Socher, Perelygin, Wu, Chuang, Manning, Ng,
  and Potts}{Socher et~al\mbox{.}}{2013}]%
        {socher2013recursive}
\bibfield{author}{\bibinfo{person}{Richard Socher}, \bibinfo{person}{Alex
  Perelygin}, \bibinfo{person}{Jean Wu}, \bibinfo{person}{Jason Chuang},
  \bibinfo{person}{Christopher~D Manning}, \bibinfo{person}{Andrew~Y Ng}, {and}
  \bibinfo{person}{Christopher Potts}.} \bibinfo{year}{2013}\natexlab{}.
\newblock \showarticletitle{Recursive deep models for semantic compositionality
  over a sentiment treebank}. In \bibinfo{booktitle}{\emph{Proceedings of the
  2013 conference on empirical methods in natural language processing}}.
  \bibinfo{pages}{1631--1642}.
\newblock


\bibitem[\protect\citeauthoryear{Solly, Sch{\"o}ning, Boch, Kandeler, Marhan,
  Michalzik, M{\"u}ller, Zscheischler, Trumbore, and Schrumpf}{Solly
  et~al\mbox{.}}{2014}]%
        {solly2014factors}
\bibfield{author}{\bibinfo{person}{Emily~F Solly}, \bibinfo{person}{Ingo
  Sch{\"o}ning}, \bibinfo{person}{Steffen Boch}, \bibinfo{person}{Ellen
  Kandeler}, \bibinfo{person}{Sven Marhan}, \bibinfo{person}{Beate Michalzik},
  \bibinfo{person}{J{\"o}rg M{\"u}ller}, \bibinfo{person}{Jakob Zscheischler},
  \bibinfo{person}{Susan~E Trumbore}, {and} \bibinfo{person}{Marion Schrumpf}.}
  \bibinfo{year}{2014}\natexlab{}.
\newblock \showarticletitle{Factors controlling decomposition rates of fine
  root litter in temperate forests and grasslands}.
\newblock \bibinfo{journal}{\emph{Plant and Soil}} \bibinfo{volume}{382},
  \bibinfo{number}{1} (\bibinfo{year}{2014}), \bibinfo{pages}{203--218}.
\newblock


\bibitem[\protect\citeauthoryear{Sontakke, Mehrjou, Itti, and
  Sch{\"o}lkopf}{Sontakke et~al\mbox{.}}{2021}]%
        {sontakke2021causal}
\bibfield{author}{\bibinfo{person}{Sumedh~A Sontakke}, \bibinfo{person}{Arash
  Mehrjou}, \bibinfo{person}{Laurent Itti}, {and} \bibinfo{person}{Bernhard
  Sch{\"o}lkopf}.} \bibinfo{year}{2021}\natexlab{}.
\newblock \showarticletitle{Causal curiosity: Rl agents discovering
  self-supervised experiments for causal representation learning}. In
  \bibinfo{booktitle}{\emph{International conference on machine learning}}.
  PMLR, \bibinfo{pages}{9848--9858}.
\newblock


\bibitem[\protect\citeauthoryear{Spirtes, Glymour, and Scheines}{Spirtes
  et~al\mbox{.}}{2000}]%
        {spirtes2000causation}
\bibfield{author}{\bibinfo{person}{Peter Spirtes}, \bibinfo{person}{Clark~N
  Glymour}, {and} \bibinfo{person}{Richard Scheines}.}
  \bibinfo{year}{2000}\natexlab{}.
\newblock \bibinfo{booktitle}{\emph{Causation, prediction, and search}}.
\newblock \bibinfo{publisher}{MIT press}.
\newblock


\bibitem[\protect\citeauthoryear{Spirtes}{Spirtes}{2013}]%
        {spirtes2013calculation}
\bibfield{author}{\bibinfo{person}{Peter~L Spirtes}.}
  \bibinfo{year}{2013}\natexlab{}.
\newblock \showarticletitle{Calculation of entailed rank constraints in
  partially non-linear and cyclic models}.
\newblock \bibinfo{journal}{\emph{arXiv preprint arXiv:1309.7004}}
  (\bibinfo{year}{2013}).
\newblock


\bibitem[\protect\citeauthoryear{Squires, Yun, Nichani, Agrawal, and
  Uhler}{Squires et~al\mbox{.}}{2022}]%
        {squires2022causal}
\bibfield{author}{\bibinfo{person}{Chandler Squires}, \bibinfo{person}{Annie
  Yun}, \bibinfo{person}{Eshaan Nichani}, \bibinfo{person}{Raj Agrawal}, {and}
  \bibinfo{person}{Caroline Uhler}.} \bibinfo{year}{2022}\natexlab{}.
\newblock \showarticletitle{Causal structure discovery between clusters of
  nodes induced by latent factors}. In \bibinfo{booktitle}{\emph{Conference on
  Causal Learning and Reasoning}}. PMLR, \bibinfo{pages}{669--687}.
\newblock


\bibitem[\protect\citeauthoryear{Sridhar and Blei}{Sridhar and Blei}{2022}]%
        {sridhar2022causal}
\bibfield{author}{\bibinfo{person}{Dhanya Sridhar} {and}
  \bibinfo{person}{David~M Blei}.} \bibinfo{year}{2022}\natexlab{}.
\newblock \showarticletitle{Causal inference from text: A commentary}.
\newblock \bibinfo{journal}{\emph{Science Advances}} \bibinfo{volume}{8},
  \bibinfo{number}{42} (\bibinfo{year}{2022}), \bibinfo{pages}{eade6585}.
\newblock


\bibitem[\protect\citeauthoryear{Stein and McKenna}{Stein and McKenna}{2013}]%
        {stein2013combining}
\bibfield{author}{\bibinfo{person}{Sebastian Stein} {and}
  \bibinfo{person}{Stephen~J McKenna}.} \bibinfo{year}{2013}\natexlab{}.
\newblock \showarticletitle{Combining embedded accelerometers with computer
  vision for recognizing food preparation activities}. In
  \bibinfo{booktitle}{\emph{Proceedings of the 2013 ACM international joint
  conference on Pervasive and ubiquitous computing}}.
  \bibinfo{pages}{729--738}.
\newblock


\bibitem[\protect\citeauthoryear{Stuart}{Stuart}{2010}]%
        {stuart2010matching}
\bibfield{author}{\bibinfo{person}{Elizabeth~A Stuart}.}
  \bibinfo{year}{2010}\natexlab{}.
\newblock \showarticletitle{Matching methods for causal inference: A review and
  a look forward}.
\newblock \bibinfo{journal}{\emph{Statistical science: a review journal of the
  Institute of Mathematical Statistics}} \bibinfo{volume}{25},
  \bibinfo{number}{1} (\bibinfo{year}{2010}), \bibinfo{pages}{1}.
\newblock


\bibitem[\protect\citeauthoryear{Sullivant, Talaska, and Draisma}{Sullivant
  et~al\mbox{.}}{2010}]%
        {sullivant2010trek}
\bibfield{author}{\bibinfo{person}{Seth Sullivant}, \bibinfo{person}{Kelli
  Talaska}, {and} \bibinfo{person}{Jan Draisma}.}
  \bibinfo{year}{2010}\natexlab{}.
\newblock \showarticletitle{Trek separation for Gaussian graphical models}.
\newblock  (\bibinfo{year}{2010}).
\newblock


\bibitem[\protect\citeauthoryear{Sun, Lu, Cheng, Cao, Dong, Tang, Zhou, and
  Mo}{Sun et~al\mbox{.}}{2022}]%
        {sun2022multi}
\bibfield{author}{\bibinfo{person}{Caiqi Sun}, \bibinfo{person}{Penghao Lu},
  \bibinfo{person}{Lei Cheng}, \bibinfo{person}{Zhenfu Cao},
  \bibinfo{person}{Xiaolei Dong}, \bibinfo{person}{Yili Tang},
  \bibinfo{person}{Jun Zhou}, {and} \bibinfo{person}{Linjian Mo}.}
  \bibinfo{year}{2022}\natexlab{}.
\newblock \showarticletitle{Multi-interest sequence modeling for recommendation
  with causal embedding}. In \bibinfo{booktitle}{\emph{Proceedings of the 2022
  SIAM International Conference on Data Mining (SDM)}}. SIAM,
  \bibinfo{pages}{406--414}.
\newblock


\bibitem[\protect\citeauthoryear{Sun, Taylor, and Bollt}{Sun
  et~al\mbox{.}}{2015}]%
        {sun2015causal}
\bibfield{author}{\bibinfo{person}{Jie Sun}, \bibinfo{person}{Dane Taylor},
  {and} \bibinfo{person}{Erik~M Bollt}.} \bibinfo{year}{2015}\natexlab{}.
\newblock \showarticletitle{Causal network inference by optimal causation
  entropy}.
\newblock \bibinfo{journal}{\emph{SIAM Journal on Applied Dynamical Systems}}
  \bibinfo{volume}{14}, \bibinfo{number}{1} (\bibinfo{year}{2015}),
  \bibinfo{pages}{73--106}.
\newblock


\bibitem[\protect\citeauthoryear{Sun, Chen, Pei, and Ren}{Sun
  et~al\mbox{.}}{2018}]%
        {sun2018emotional}
\bibfield{author}{\bibinfo{person}{Xiao Sun}, \bibinfo{person}{Xinmiao Chen},
  \bibinfo{person}{Zhengmeng Pei}, {and} \bibinfo{person}{Fuji Ren}.}
  \bibinfo{year}{2018}\natexlab{}.
\newblock \showarticletitle{Emotional human machine conversation generation
  based on SeqGAN}. In \bibinfo{booktitle}{\emph{2018 First Asian Conference on
  Affective Computing and Intelligent Interaction (ACII Asia)}}. IEEE,
  \bibinfo{pages}{1--6}.
\newblock


\bibitem[\protect\citeauthoryear{Sun, Janzing, and Sch{\"o}lkopf}{Sun
  et~al\mbox{.}}{2006}]%
        {sun2006causal}
\bibfield{author}{\bibinfo{person}{Xiaohai Sun}, \bibinfo{person}{Dominik
  Janzing}, {and} \bibinfo{person}{Bernhard Sch{\"o}lkopf}.}
  \bibinfo{year}{2006}\natexlab{}.
\newblock \showarticletitle{Causal inference by choosing graphs with most
  plausible Markov kernels}. In \bibinfo{booktitle}{\emph{Ninth International
  Symposium on Artificial Intelligence and Mathematics (AIMath 2006)}}.
  \bibinfo{pages}{1--11}.
\newblock


\bibitem[\protect\citeauthoryear{Sun, Schulte, Liu, and Poupart}{Sun
  et~al\mbox{.}}{2021}]%
        {sun2021nts}
\bibfield{author}{\bibinfo{person}{Xiangyu Sun}, \bibinfo{person}{Oliver
  Schulte}, \bibinfo{person}{Guiliang Liu}, {and} \bibinfo{person}{Pascal
  Poupart}.} \bibinfo{year}{2021}\natexlab{}.
\newblock \showarticletitle{NTS-NOTEARS: Learning Nonparametric DBNs With Prior
  Knowledge}.
\newblock \bibinfo{journal}{\emph{arXiv preprint arXiv:2109.04286}}
  (\bibinfo{year}{2021}).
\newblock


\bibitem[\protect\citeauthoryear{Talmor, Herzig, Lourie, and Berant}{Talmor
  et~al\mbox{.}}{2018}]%
        {talmor2018commonsenseqa}
\bibfield{author}{\bibinfo{person}{Alon Talmor}, \bibinfo{person}{Jonathan
  Herzig}, \bibinfo{person}{Nicholas Lourie}, {and} \bibinfo{person}{Jonathan
  Berant}.} \bibinfo{year}{2018}\natexlab{}.
\newblock \showarticletitle{Commonsenseqa: A question answering challenge
  targeting commonsense knowledge}.
\newblock \bibinfo{journal}{\emph{arXiv preprint arXiv:1811.00937}}
  (\bibinfo{year}{2018}).
\newblock


\bibitem[\protect\citeauthoryear{Tan, H{\"u}rriyeto{\u{g}}lu, Caselli,
  Oostdijk, Nomoto, Hettiarachchi, Ameer, Uca, Liza, and Hu}{Tan
  et~al\mbox{.}}{2022}]%
        {tan2022causal}
\bibfield{author}{\bibinfo{person}{Fiona~Anting Tan}, \bibinfo{person}{Ali
  H{\"u}rriyeto{\u{g}}lu}, \bibinfo{person}{Tommaso Caselli},
  \bibinfo{person}{Nelleke Oostdijk}, \bibinfo{person}{Tadashi Nomoto},
  \bibinfo{person}{Hansi Hettiarachchi}, \bibinfo{person}{Iqra Ameer},
  \bibinfo{person}{Onur Uca}, \bibinfo{person}{Farhana~Ferdousi Liza}, {and}
  \bibinfo{person}{Tiancheng Hu}.} \bibinfo{year}{2022}\natexlab{}.
\newblock \showarticletitle{The Causal News Corpus: Annotating Causal Relations
  in Event Sentences from News}.
\newblock \bibinfo{journal}{\emph{arXiv preprint arXiv:2204.11714}}
  (\bibinfo{year}{2022}).
\newblock


\bibitem[\protect\citeauthoryear{Tank, Cover, Foti, Shojaie, and Fox}{Tank
  et~al\mbox{.}}{2017}]%
        {tank2017interpretable}
\bibfield{author}{\bibinfo{person}{Alex Tank}, \bibinfo{person}{Ian Cover},
  \bibinfo{person}{Nicholas~J Foti}, \bibinfo{person}{Ali Shojaie}, {and}
  \bibinfo{person}{Emily~B Fox}.} \bibinfo{year}{2017}\natexlab{}.
\newblock \showarticletitle{An interpretable and sparse neural network model
  for nonlinear granger causality discovery}.
\newblock \bibinfo{journal}{\emph{arXiv preprint arXiv:1711.08160}}
  (\bibinfo{year}{2017}).
\newblock


\bibitem[\protect\citeauthoryear{Taori, Gulrajani, Zhang, Dubois, Li, Guestrin,
  Liang, and Hashimoto}{Taori et~al\mbox{.}}{2023}]%
        {taori2023stanford}
\bibfield{author}{\bibinfo{person}{Rohan Taori}, \bibinfo{person}{Ishaan
  Gulrajani}, \bibinfo{person}{Tianyi Zhang}, \bibinfo{person}{Yann Dubois},
  \bibinfo{person}{Xuechen Li}, \bibinfo{person}{Carlos Guestrin},
  \bibinfo{person}{Percy Liang}, {and} \bibinfo{person}{Tatsunori~B
  Hashimoto}.} \bibinfo{year}{2023}\natexlab{}.
\newblock \bibinfo{title}{Stanford alpaca: An instruction-following llama
  model}.
\newblock
\newblock


\bibitem[\protect\citeauthoryear{Touvron, Lavril, Izacard, Martinet, Lachaux,
  Lacroix, Rozi{\`e}re, Goyal, Hambro, Azhar, et~al\mbox{.}}{Touvron
  et~al\mbox{.}}{2023}]%
        {touvron2023llama}
\bibfield{author}{\bibinfo{person}{Hugo Touvron}, \bibinfo{person}{Thibaut
  Lavril}, \bibinfo{person}{Gautier Izacard}, \bibinfo{person}{Xavier
  Martinet}, \bibinfo{person}{Marie-Anne Lachaux},
  \bibinfo{person}{Timoth{\'e}e Lacroix}, \bibinfo{person}{Baptiste
  Rozi{\`e}re}, \bibinfo{person}{Naman Goyal}, \bibinfo{person}{Eric Hambro},
  \bibinfo{person}{Faisal Azhar}, {et~al\mbox{.}}}
  \bibinfo{year}{2023}\natexlab{}.
\newblock \showarticletitle{Llama: Open and efficient foundation language
  models}.
\newblock \bibinfo{journal}{\emph{arXiv preprint arXiv:2302.13971}}
  (\bibinfo{year}{2023}).
\newblock


\bibitem[\protect\citeauthoryear{Trabasso and Van Den~Broek}{Trabasso and Van
  Den~Broek}{1985}]%
        {trabasso1985causal}
\bibfield{author}{\bibinfo{person}{Tom Trabasso} {and} \bibinfo{person}{Paul
  Van Den~Broek}.} \bibinfo{year}{1985}\natexlab{}.
\newblock \showarticletitle{Causal thinking and the representation of narrative
  events}.
\newblock \bibinfo{journal}{\emph{Journal of memory and language}}
  \bibinfo{volume}{24}, \bibinfo{number}{5} (\bibinfo{year}{1985}),
  \bibinfo{pages}{612--630}.
\newblock


\bibitem[\protect\citeauthoryear{Tu, Zhang, Ackermann, Mohan, Kjellstr{\"o}m,
  and Zhang}{Tu et~al\mbox{.}}{2019}]%
        {tu2019causal}
\bibfield{author}{\bibinfo{person}{Ruibo Tu}, \bibinfo{person}{Cheng Zhang},
  \bibinfo{person}{Paul Ackermann}, \bibinfo{person}{Karthika Mohan},
  \bibinfo{person}{Hedvig Kjellstr{\"o}m}, {and} \bibinfo{person}{Kun Zhang}.}
  \bibinfo{year}{2019}\natexlab{}.
\newblock \showarticletitle{Causal discovery in the presence of missing data}.
  In \bibinfo{booktitle}{\emph{The 22nd International Conference on Artificial
  Intelligence and Statistics}}. PMLR, \bibinfo{pages}{1762--1770}.
\newblock


\bibitem[\protect\citeauthoryear{Ukita}{Ukita}{2020}]%
        {ukita2020causal}
\bibfield{author}{\bibinfo{person}{Jumpei Ukita}.}
  \bibinfo{year}{2020}\natexlab{}.
\newblock \showarticletitle{Causal importance of low-level feature selectivity
  for generalization in image recognition}.
\newblock \bibinfo{journal}{\emph{Neural Networks}}  \bibinfo{volume}{125}
  (\bibinfo{year}{2020}), \bibinfo{pages}{185--193}.
\newblock


\bibitem[\protect\citeauthoryear{Van~den Bulcke, Van~Leemput, Naudts, van
  Remortel, Ma, Verschoren, De~Moor, and Marchal}{Van~den Bulcke
  et~al\mbox{.}}{2006}]%
        {van2006syntren}
\bibfield{author}{\bibinfo{person}{Tim Van~den Bulcke},
  \bibinfo{person}{Koenraad Van~Leemput}, \bibinfo{person}{Bart Naudts},
  \bibinfo{person}{Piet van Remortel}, \bibinfo{person}{Hongwu Ma},
  \bibinfo{person}{Alain Verschoren}, \bibinfo{person}{Bart De~Moor}, {and}
  \bibinfo{person}{Kathleen Marchal}.} \bibinfo{year}{2006}\natexlab{}.
\newblock \showarticletitle{SynTReN: a generator of synthetic gene expression
  data for design and analysis of structure learning algorithms}.
\newblock \bibinfo{journal}{\emph{BMC bioinformatics}} \bibinfo{volume}{7},
  \bibinfo{number}{1} (\bibinfo{year}{2006}), \bibinfo{pages}{1--12}.
\newblock


\bibitem[\protect\citeauthoryear{Vapnik}{Vapnik}{1999}]%
        {vapnik1999nature}
\bibfield{author}{\bibinfo{person}{Vladimir Vapnik}.}
  \bibinfo{year}{1999}\natexlab{}.
\newblock \bibinfo{booktitle}{\emph{The nature of statistical learning
  theory}}.
\newblock \bibinfo{publisher}{Springer science \& business media}.
\newblock


\bibitem[\protect\citeauthoryear{Veitch, Sridhar, and Blei}{Veitch
  et~al\mbox{.}}{2020}]%
        {veitch2020adapting}
\bibfield{author}{\bibinfo{person}{Victor Veitch}, \bibinfo{person}{Dhanya
  Sridhar}, {and} \bibinfo{person}{David Blei}.}
  \bibinfo{year}{2020}\natexlab{}.
\newblock \showarticletitle{Adapting text embeddings for causal inference}. In
  \bibinfo{booktitle}{\emph{Conference on Uncertainty in Artificial
  Intelligence}}. PMLR, \bibinfo{pages}{919--928}.
\newblock


\bibitem[\protect\citeauthoryear{Vowels, Camgoz, and Bowden}{Vowels
  et~al\mbox{.}}{2022}]%
        {vowels2022d}
\bibfield{author}{\bibinfo{person}{Matthew~J Vowels},
  \bibinfo{person}{Necati~Cihan Camgoz}, {and} \bibinfo{person}{Richard
  Bowden}.} \bibinfo{year}{2022}\natexlab{}.
\newblock \showarticletitle{D’ya like dags? a survey on structure learning
  and causal discovery}.
\newblock \bibinfo{journal}{\emph{Comput. Surveys}} \bibinfo{volume}{55},
  \bibinfo{number}{4} (\bibinfo{year}{2022}), \bibinfo{pages}{1--36}.
\newblock


\bibitem[\protect\citeauthoryear{Wager and Athey}{Wager and Athey}{2018}]%
        {wager2018estimation}
\bibfield{author}{\bibinfo{person}{Stefan Wager} {and} \bibinfo{person}{Susan
  Athey}.} \bibinfo{year}{2018}\natexlab{}.
\newblock \showarticletitle{Estimation and inference of heterogeneous treatment
  effects using random forests}.
\newblock \bibinfo{journal}{\emph{J. Amer. Statist. Assoc.}}
  \bibinfo{volume}{113}, \bibinfo{number}{523} (\bibinfo{year}{2018}),
  \bibinfo{pages}{1228--1242}.
\newblock


\bibitem[\protect\citeauthoryear{Wah, Branson, Welinder, Perona, and
  Belongie}{Wah et~al\mbox{.}}{2011}]%
        {wah2011caltech}
\bibfield{author}{\bibinfo{person}{Catherine Wah}, \bibinfo{person}{Steve
  Branson}, \bibinfo{person}{Peter Welinder}, \bibinfo{person}{Pietro Perona},
  {and} \bibinfo{person}{Serge Belongie}.} \bibinfo{year}{2011}\natexlab{}.
\newblock \showarticletitle{The caltech-ucsd birds-200-2011 dataset}.
\newblock  (\bibinfo{year}{2011}).
\newblock


\bibitem[\protect\citeauthoryear{Wang and Zhou}{Wang and Zhou}{2021}]%
        {wang2021actively}
\bibfield{author}{\bibinfo{person}{Tian-Zuo Wang} {and}
  \bibinfo{person}{Zhi-Hua Zhou}.} \bibinfo{year}{2021}\natexlab{}.
\newblock \showarticletitle{Actively identifying causal effects with latent
  variables given only response variable observable}.
\newblock \bibinfo{journal}{\emph{Advances in Neural Information Processing
  Systems}}  \bibinfo{volume}{34} (\bibinfo{year}{2021}),
  \bibinfo{pages}{15007--15018}.
\newblock


\bibitem[\protect\citeauthoryear{Wang, Lin, Feng, He, Lin, and Chua}{Wang
  et~al\mbox{.}}{2022a}]%
        {wang2022causalr}
\bibfield{author}{\bibinfo{person}{Wenjie Wang}, \bibinfo{person}{Xinyu Lin},
  \bibinfo{person}{Fuli Feng}, \bibinfo{person}{Xiangnan He},
  \bibinfo{person}{Min Lin}, {and} \bibinfo{person}{Tat-Seng Chua}.}
  \bibinfo{year}{2022}\natexlab{a}.
\newblock \showarticletitle{Causal representation learning for
  out-of-distribution recommendation}. In \bibinfo{booktitle}{\emph{Proceedings
  of the ACM Web Conference 2022}}. \bibinfo{pages}{3562--3571}.
\newblock


\bibitem[\protect\citeauthoryear{Wang, Du, Zhu, Ke, Chen, Hao, and Wang}{Wang
  et~al\mbox{.}}{2021}]%
        {wang2021ordering}
\bibfield{author}{\bibinfo{person}{Xiaoqiang Wang}, \bibinfo{person}{Yali Du},
  \bibinfo{person}{Shengyu Zhu}, \bibinfo{person}{Liangjun Ke},
  \bibinfo{person}{Zhitang Chen}, \bibinfo{person}{Jianye Hao}, {and}
  \bibinfo{person}{Jun Wang}.} \bibinfo{year}{2021}\natexlab{}.
\newblock \showarticletitle{Ordering-based causal discovery with reinforcement
  learning}.
\newblock \bibinfo{journal}{\emph{arXiv preprint arXiv:2105.06631}}
  (\bibinfo{year}{2021}).
\newblock


\bibitem[\protect\citeauthoryear{Wang, Hutchinson, and Mitchell}{Wang
  et~al\mbox{.}}{2003}]%
        {wang2003training}
\bibfield{author}{\bibinfo{person}{Xuerui Wang}, \bibinfo{person}{Rebecca
  Hutchinson}, {and} \bibinfo{person}{Tom~M Mitchell}.}
  \bibinfo{year}{2003}\natexlab{}.
\newblock \showarticletitle{Training fMRI classifiers to detect cognitive
  states across multiple human subjects}.
\newblock \bibinfo{journal}{\emph{Advances in neural information processing
  systems}}  \bibinfo{volume}{16} (\bibinfo{year}{2003}).
\newblock


\bibitem[\protect\citeauthoryear{Wang and Drton}{Wang and Drton}{2020}]%
        {wang2020high}
\bibfield{author}{\bibinfo{person}{Y~Samuel Wang} {and}
  \bibinfo{person}{Mathias Drton}.} \bibinfo{year}{2020}\natexlab{}.
\newblock \showarticletitle{High-dimensional causal discovery under
  non-Gaussianity}.
\newblock \bibinfo{journal}{\emph{Biometrika}} \bibinfo{volume}{107},
  \bibinfo{number}{1} (\bibinfo{year}{2020}), \bibinfo{pages}{41--59}.
\newblock


\bibitem[\protect\citeauthoryear{Wang, Xiao, Xu, Zhu, and Stone}{Wang
  et~al\mbox{.}}{2022b}]%
        {wang2022causal}
\bibfield{author}{\bibinfo{person}{Zizhao Wang}, \bibinfo{person}{Xuesu Xiao},
  \bibinfo{person}{Zifan Xu}, \bibinfo{person}{Yuke Zhu}, {and}
  \bibinfo{person}{Peter Stone}.} \bibinfo{year}{2022}\natexlab{b}.
\newblock \showarticletitle{Causal dynamics learning for task-independent state
  abstraction}.
\newblock \bibinfo{journal}{\emph{arXiv preprint arXiv:2206.13452}}
  (\bibinfo{year}{2022}).
\newblock


\bibitem[\protect\citeauthoryear{Ward, Hartley, and Tillyer}{Ward
  et~al\mbox{.}}{2016}]%
        {ward2016unpacking}
\bibfield{author}{\bibinfo{person}{Jeffrey~T Ward}, \bibinfo{person}{Richard~D
  Hartley}, {and} \bibinfo{person}{Rob Tillyer}.}
  \bibinfo{year}{2016}\natexlab{}.
\newblock \showarticletitle{Unpacking gender and racial/ethnic biases in the
  federal sentencing of drug offenders: A causal mediation approach}.
\newblock \bibinfo{journal}{\emph{Journal of Criminal Justice}}
  \bibinfo{volume}{46} (\bibinfo{year}{2016}), \bibinfo{pages}{196--206}.
\newblock


\bibitem[\protect\citeauthoryear{Wei, Wang, Schuurmans, Bosma, Xia, Chi, Le,
  Zhou, et~al\mbox{.}}{Wei et~al\mbox{.}}{2022}]%
        {wei2022chain}
\bibfield{author}{\bibinfo{person}{Jason Wei}, \bibinfo{person}{Xuezhi Wang},
  \bibinfo{person}{Dale Schuurmans}, \bibinfo{person}{Maarten Bosma},
  \bibinfo{person}{Fei Xia}, \bibinfo{person}{Ed Chi}, \bibinfo{person}{Quoc~V
  Le}, \bibinfo{person}{Denny Zhou}, {et~al\mbox{.}}}
  \bibinfo{year}{2022}\natexlab{}.
\newblock \showarticletitle{Chain-of-thought prompting elicits reasoning in
  large language models}.
\newblock \bibinfo{journal}{\emph{Advances in Neural Information Processing
  Systems}}  \bibinfo{volume}{35} (\bibinfo{year}{2022}),
  \bibinfo{pages}{24824--24837}.
\newblock


\bibitem[\protect\citeauthoryear{Wei, Zhao, and Mao}{Wei et~al\mbox{.}}{2020}]%
        {wei-etal-2020-effective}
\bibfield{author}{\bibinfo{person}{Penghui Wei}, \bibinfo{person}{Jiahao Zhao},
  {and} \bibinfo{person}{Wenji Mao}.} \bibinfo{year}{2020}\natexlab{}.
\newblock \showarticletitle{Effective Inter-Clause Modeling for End-to-End
  Emotion-Cause Pair Extraction}. In \bibinfo{booktitle}{\emph{Proceedings of
  the 58th Annual Meeting of the Association for Computational Linguistics}}.
  \bibinfo{publisher}{Association for Computational Linguistics},
  \bibinfo{address}{Online}, \bibinfo{pages}{3171--3181}.
\newblock
\urldef\tempurl%
\url{https://doi.org/10.18653/v1/2020.acl-main.289}
\showDOI{\tempurl}


\bibitem[\protect\citeauthoryear{Weinstein, Collisson, Mills, Shaw, Ozenberger,
  Ellrott, Shmulevich, Sander, and Stuart}{Weinstein et~al\mbox{.}}{2013}]%
        {weinstein2013cancer}
\bibfield{author}{\bibinfo{person}{John~N Weinstein}, \bibinfo{person}{Eric~A
  Collisson}, \bibinfo{person}{Gordon~B Mills}, \bibinfo{person}{Kenna~R Shaw},
  \bibinfo{person}{Brad~A Ozenberger}, \bibinfo{person}{Kyle Ellrott},
  \bibinfo{person}{Ilya Shmulevich}, \bibinfo{person}{Chris Sander}, {and}
  \bibinfo{person}{Joshua~M Stuart}.} \bibinfo{year}{2013}\natexlab{}.
\newblock \showarticletitle{The cancer genome atlas pan-cancer analysis
  project}.
\newblock \bibinfo{journal}{\emph{Nature genetics}} \bibinfo{volume}{45},
  \bibinfo{number}{10} (\bibinfo{year}{2013}), \bibinfo{pages}{1113--1120}.
\newblock


\bibitem[\protect\citeauthoryear{Wijmans and Baker}{Wijmans and Baker}{1995}]%
        {wijmans1995solution}
\bibfield{author}{\bibinfo{person}{Johannes~G Wijmans} {and}
  \bibinfo{person}{Richard~W Baker}.} \bibinfo{year}{1995}\natexlab{}.
\newblock \showarticletitle{The solution-diffusion model: a review}.
\newblock \bibinfo{journal}{\emph{Journal of membrane science}}
  \bibinfo{volume}{107}, \bibinfo{number}{1-2} (\bibinfo{year}{1995}),
  \bibinfo{pages}{1--21}.
\newblock


\bibitem[\protect\citeauthoryear{Wu, Breuel, Skuhersky, and Kautz}{Wu
  et~al\mbox{.}}{2020}]%
        {wu2020discovering}
\bibfield{author}{\bibinfo{person}{Tailin Wu}, \bibinfo{person}{Thomas Breuel},
  \bibinfo{person}{Michael Skuhersky}, {and} \bibinfo{person}{Jan Kautz}.}
  \bibinfo{year}{2020}\natexlab{}.
\newblock \showarticletitle{Discovering nonlinear relations with minimum
  predictive information regularization}.
\newblock \bibinfo{journal}{\emph{arXiv preprint arXiv:2001.01885}}
  (\bibinfo{year}{2020}).
\newblock


\bibitem[\protect\citeauthoryear{Wu, Wu, Wang, Liu, and Chen}{Wu
  et~al\mbox{.}}{2022}]%
        {wu2022nonlinear}
\bibfield{author}{\bibinfo{person}{Tianhao Wu}, \bibinfo{person}{Xingyu Wu},
  \bibinfo{person}{Xin Wang}, \bibinfo{person}{Shikang Liu}, {and}
  \bibinfo{person}{Huanhuan Chen}.} \bibinfo{year}{2022}\natexlab{}.
\newblock \showarticletitle{Nonlinear Causal Discovery in Time Series}. In
  \bibinfo{booktitle}{\emph{Proceedings of the 31st ACM International
  Conference on Information \& Knowledge Management}}.
  \bibinfo{pages}{4575--4579}.
\newblock


\bibitem[\protect\citeauthoryear{Wu, Wang, Zhang, He, and Chua}{Wu
  et~al\mbox{.}}{2021}]%
        {wu2021discovering}
\bibfield{author}{\bibinfo{person}{Yingxin Wu}, \bibinfo{person}{Xiang Wang},
  \bibinfo{person}{An Zhang}, \bibinfo{person}{Xiangnan He}, {and}
  \bibinfo{person}{Tat-Seng Chua}.} \bibinfo{year}{2021}\natexlab{}.
\newblock \showarticletitle{Discovering Invariant Rationales for Graph Neural
  Networks}. In \bibinfo{booktitle}{\emph{International Conference on Learning
  Representations}}.
\newblock


\bibitem[\protect\citeauthoryear{Wu, Ramsundar, Feinberg, Gomes, Geniesse,
  Pappu, Leswing, and Pande}{Wu et~al\mbox{.}}{2018}]%
        {wu2018moleculenet}
\bibfield{author}{\bibinfo{person}{Zhenqin Wu}, \bibinfo{person}{Bharath
  Ramsundar}, \bibinfo{person}{Evan~N Feinberg}, \bibinfo{person}{Joseph
  Gomes}, \bibinfo{person}{Caleb Geniesse}, \bibinfo{person}{Aneesh~S Pappu},
  \bibinfo{person}{Karl Leswing}, {and} \bibinfo{person}{Vijay Pande}.}
  \bibinfo{year}{2018}\natexlab{}.
\newblock \showarticletitle{MoleculeNet: a benchmark for molecular machine
  learning}.
\newblock \bibinfo{journal}{\emph{Chemical science}} \bibinfo{volume}{9},
  \bibinfo{number}{2} (\bibinfo{year}{2018}), \bibinfo{pages}{513--530}.
\newblock


\bibitem[\protect\citeauthoryear{Xie, Cai, Huang, Glymour, Hao, and Zhang}{Xie
  et~al\mbox{.}}{2020}]%
        {xie2020generalized}
\bibfield{author}{\bibinfo{person}{Feng Xie}, \bibinfo{person}{Ruichu Cai},
  \bibinfo{person}{Biwei Huang}, \bibinfo{person}{Clark Glymour},
  \bibinfo{person}{Zhifeng Hao}, {and} \bibinfo{person}{Kun Zhang}.}
  \bibinfo{year}{2020}\natexlab{}.
\newblock \showarticletitle{Generalized independent noise condition for
  estimating latent variable causal graphs}.
\newblock \bibinfo{journal}{\emph{Advances in neural information processing
  systems}}  \bibinfo{volume}{33} (\bibinfo{year}{2020}),
  \bibinfo{pages}{14891--14902}.
\newblock


\bibitem[\protect\citeauthoryear{Xie, Huang, Chen, He, Geng, and Zhang}{Xie
  et~al\mbox{.}}{2022}]%
        {xie2022identification}
\bibfield{author}{\bibinfo{person}{Feng Xie}, \bibinfo{person}{Biwei Huang},
  \bibinfo{person}{Zhengming Chen}, \bibinfo{person}{Yangbo He},
  \bibinfo{person}{Zhi Geng}, {and} \bibinfo{person}{Kun Zhang}.}
  \bibinfo{year}{2022}\natexlab{}.
\newblock \showarticletitle{Identification of linear non-gaussian latent
  hierarchical structure}. In \bibinfo{booktitle}{\emph{International
  Conference on Machine Learning}}. PMLR, \bibinfo{pages}{24370--24387}.
\newblock


\bibitem[\protect\citeauthoryear{Xie and Mu}{Xie and Mu}{2019}]%
        {xie2019boosting}
\bibfield{author}{\bibinfo{person}{Zhipeng Xie} {and} \bibinfo{person}{Feiteng
  Mu}.} \bibinfo{year}{2019}\natexlab{}.
\newblock \showarticletitle{Boosting Causal Embeddings via Potential
  Verb-Mediated Causal Patterns.}. In \bibinfo{booktitle}{\emph{IJCAI}}.
  \bibinfo{pages}{1921--1927}.
\newblock


\bibitem[\protect\citeauthoryear{Xu, Skoularidou, Cuesta-Infante, and
  Veeramachaneni}{Xu et~al\mbox{.}}{2019}]%
        {xu2019modeling}
\bibfield{author}{\bibinfo{person}{Lei Xu}, \bibinfo{person}{Maria
  Skoularidou}, \bibinfo{person}{Alfredo Cuesta-Infante}, {and}
  \bibinfo{person}{Kalyan Veeramachaneni}.} \bibinfo{year}{2019}\natexlab{}.
\newblock \showarticletitle{Modeling tabular data using conditional gan}.
\newblock \bibinfo{journal}{\emph{Advances in neural information processing
  systems}}  \bibinfo{volume}{32} (\bibinfo{year}{2019}).
\newblock


\bibitem[\protect\citeauthoryear{Xu, Mou, Li, Chen, Peng, and Jin}{Xu
  et~al\mbox{.}}{2015}]%
        {xu2015classifying}
\bibfield{author}{\bibinfo{person}{Yan Xu}, \bibinfo{person}{Lili Mou},
  \bibinfo{person}{Ge Li}, \bibinfo{person}{Yunchuan Chen},
  \bibinfo{person}{Hao Peng}, {and} \bibinfo{person}{Zhi Jin}.}
  \bibinfo{year}{2015}\natexlab{}.
\newblock \showarticletitle{Classifying relations via long short term memory
  networks along shortest dependency paths}. In
  \bibinfo{booktitle}{\emph{Proceedings of the 2015 conference on empirical
  methods in natural language processing}}. \bibinfo{pages}{1785--1794}.
\newblock


\bibitem[\protect\citeauthoryear{Yang, Han, and Poon}{Yang
  et~al\mbox{.}}{2022}]%
        {yang2022survey}
\bibfield{author}{\bibinfo{person}{Jie Yang}, \bibinfo{person}{Soyeon~Caren
  Han}, {and} \bibinfo{person}{Josiah Poon}.} \bibinfo{year}{2022}\natexlab{}.
\newblock \showarticletitle{A survey on extraction of causal relations from
  natural language text}.
\newblock \bibinfo{journal}{\emph{Knowledge and Information Systems}}
  \bibinfo{volume}{64}, \bibinfo{number}{5} (\bibinfo{year}{2022}),
  \bibinfo{pages}{1161--1186}.
\newblock


\bibitem[\protect\citeauthoryear{Yao, Chu, Li, Li, Gao, and Zhang}{Yao
  et~al\mbox{.}}{2021}]%
        {yao2021survey}
\bibfield{author}{\bibinfo{person}{Liuyi Yao}, \bibinfo{person}{Zhixuan Chu},
  \bibinfo{person}{Sheng Li}, \bibinfo{person}{Yaliang Li},
  \bibinfo{person}{Jing Gao}, {and} \bibinfo{person}{Aidong Zhang}.}
  \bibinfo{year}{2021}\natexlab{}.
\newblock \showarticletitle{A survey on causal inference}.
\newblock \bibinfo{journal}{\emph{ACM Transactions on Knowledge Discovery from
  Data (TKDD)}} \bibinfo{volume}{15}, \bibinfo{number}{5}
  (\bibinfo{year}{2021}), \bibinfo{pages}{1--46}.
\newblock


\bibitem[\protect\citeauthoryear{Yao and Ge}{Yao and Ge}{2023}]%
        {yao2023causal}
\bibfield{author}{\bibinfo{person}{Le Yao} {and} \bibinfo{person}{Zhiqiang
  Ge}.} \bibinfo{year}{2023}\natexlab{}.
\newblock \showarticletitle{Causal variable selection for industrial process
  quality prediction via attention-based GRU network}.
\newblock \bibinfo{journal}{\emph{Engineering Applications of Artificial
  Intelligence}}  \bibinfo{volume}{118} (\bibinfo{year}{2023}),
  \bibinfo{pages}{105658}.
\newblock


\bibitem[\protect\citeauthoryear{Yao, Li, Li, Huai, Gao, and Zhang}{Yao
  et~al\mbox{.}}{2019a}]%
        {yao2019ace}
\bibfield{author}{\bibinfo{person}{Liuyi Yao}, \bibinfo{person}{Sheng Li},
  \bibinfo{person}{Yaliang Li}, \bibinfo{person}{Mengdi Huai},
  \bibinfo{person}{Jing Gao}, {and} \bibinfo{person}{Aidong Zhang}.}
  \bibinfo{year}{2019}\natexlab{a}.
\newblock \showarticletitle{Ace: Adaptively similarity-preserved representation
  learning for individual treatment effect estimation}. In
  \bibinfo{booktitle}{\emph{2019 IEEE International Conference on Data Mining
  (ICDM)}}. IEEE, \bibinfo{pages}{1432--1437}.
\newblock


\bibitem[\protect\citeauthoryear{Yao, Li, Li, Xue, Gao, and Zhang}{Yao
  et~al\mbox{.}}{2019b}]%
        {yao2019estimation}
\bibfield{author}{\bibinfo{person}{Liuyi Yao}, \bibinfo{person}{Sheng Li},
  \bibinfo{person}{Yaliang Li}, \bibinfo{person}{Hongfei Xue},
  \bibinfo{person}{Jing Gao}, {and} \bibinfo{person}{Aidong Zhang}.}
  \bibinfo{year}{2019}\natexlab{b}.
\newblock \showarticletitle{On the estimation of treatment effect with text
  covariates}. In \bibinfo{booktitle}{\emph{International Joint Conference on
  Artificial Intelligence}}.
\newblock


\bibitem[\protect\citeauthoryear{Yao, Chen, and Zhang}{Yao
  et~al\mbox{.}}{2022}]%
        {yao2022learning}
\bibfield{author}{\bibinfo{person}{Weiran Yao}, \bibinfo{person}{Guangyi Chen},
  {and} \bibinfo{person}{Kun Zhang}.} \bibinfo{year}{2022}\natexlab{}.
\newblock \showarticletitle{Learning latent causal dynamics}.
\newblock \bibinfo{journal}{\emph{arXiv preprint arXiv:2202.04828}}
  (\bibinfo{year}{2022}).
\newblock


\bibitem[\protect\citeauthoryear{Ying, Bourgeois, You, Zitnik, and
  Leskovec}{Ying et~al\mbox{.}}{2019}]%
        {ying2019gnnexplainer}
\bibfield{author}{\bibinfo{person}{Zhitao Ying}, \bibinfo{person}{Dylan
  Bourgeois}, \bibinfo{person}{Jiaxuan You}, \bibinfo{person}{Marinka Zitnik},
  {and} \bibinfo{person}{Jure Leskovec}.} \bibinfo{year}{2019}\natexlab{}.
\newblock \showarticletitle{Gnnexplainer: Generating explanations for graph
  neural networks}.
\newblock \bibinfo{journal}{\emph{Advances in neural information processing
  systems}}  \bibinfo{volume}{32} (\bibinfo{year}{2019}).
\newblock


\bibitem[\protect\citeauthoryear{Yu, Chen, Gao, and Yu}{Yu
  et~al\mbox{.}}{2019}]%
        {yu2019dag}
\bibfield{author}{\bibinfo{person}{Yue Yu}, \bibinfo{person}{Jie Chen},
  \bibinfo{person}{Tian Gao}, {and} \bibinfo{person}{Mo Yu}.}
  \bibinfo{year}{2019}\natexlab{}.
\newblock \showarticletitle{DAG-GNN: DAG structure learning with graph neural
  networks}. In \bibinfo{booktitle}{\emph{International Conference on Machine
  Learning}}. PMLR, \bibinfo{pages}{7154--7163}.
\newblock


\bibitem[\protect\citeauthoryear{Zhang, B{\"u}tepage, Kjellstr{\"o}m, and
  Mandt}{Zhang et~al\mbox{.}}{2018}]%
        {zhang2018advances}
\bibfield{author}{\bibinfo{person}{Cheng Zhang}, \bibinfo{person}{Judith
  B{\"u}tepage}, \bibinfo{person}{Hedvig Kjellstr{\"o}m}, {and}
  \bibinfo{person}{Stephan Mandt}.} \bibinfo{year}{2018}\natexlab{}.
\newblock \showarticletitle{Advances in variational inference}.
\newblock \bibinfo{journal}{\emph{IEEE transactions on pattern analysis and
  machine intelligence}} \bibinfo{volume}{41}, \bibinfo{number}{8}
  (\bibinfo{year}{2018}), \bibinfo{pages}{2008--2026}.
\newblock


\bibitem[\protect\citeauthoryear{Zhang}{Zhang}{2008}]%
        {zhang2008completeness}
\bibfield{author}{\bibinfo{person}{Jiji Zhang}.}
  \bibinfo{year}{2008}\natexlab{}.
\newblock \showarticletitle{On the completeness of orientation rules for causal
  discovery in the presence of latent confounders and selection bias}.
\newblock \bibinfo{journal}{\emph{Artificial Intelligence}}
  \bibinfo{volume}{172}, \bibinfo{number}{16-17} (\bibinfo{year}{2008}),
  \bibinfo{pages}{1873--1896}.
\newblock


\bibitem[\protect\citeauthoryear{Zhang, Huang, Zhang, Glymour, and
  Sch{\"o}lkopf}{Zhang et~al\mbox{.}}{2017}]%
        {zhang2017causal}
\bibfield{author}{\bibinfo{person}{Kun Zhang}, \bibinfo{person}{Biwei Huang},
  \bibinfo{person}{Jiji Zhang}, \bibinfo{person}{Clark Glymour}, {and}
  \bibinfo{person}{Bernhard Sch{\"o}lkopf}.} \bibinfo{year}{2017}\natexlab{}.
\newblock \showarticletitle{Causal discovery from nonstationary/heterogeneous
  data: Skeleton estimation and orientation determination}. In
  \bibinfo{booktitle}{\emph{IJCAI: Proceedings of the Conference}},
  Vol.~\bibinfo{volume}{2017}. NIH Public Access, \bibinfo{pages}{1347}.
\newblock


\bibitem[\protect\citeauthoryear{Zhang and Hyvarinen}{Zhang and
  Hyvarinen}{2012}]%
        {zhang2012identifiability}
\bibfield{author}{\bibinfo{person}{Kun Zhang} {and} \bibinfo{person}{Aapo
  Hyvarinen}.} \bibinfo{year}{2012}\natexlab{}.
\newblock \showarticletitle{On the identifiability of the post-nonlinear causal
  model}.
\newblock \bibinfo{journal}{\emph{arXiv preprint arXiv:1205.2599}}
  (\bibinfo{year}{2012}).
\newblock


\bibitem[\protect\citeauthoryear{Zhang, Wang, Zhang, and Sch{\"o}lkopf}{Zhang
  et~al\mbox{.}}{2015a}]%
        {zhang2015estimation}
\bibfield{author}{\bibinfo{person}{Kun Zhang}, \bibinfo{person}{Zhikun Wang},
  \bibinfo{person}{Jiji Zhang}, {and} \bibinfo{person}{Bernhard
  Sch{\"o}lkopf}.} \bibinfo{year}{2015}\natexlab{a}.
\newblock \showarticletitle{On estimation of functional causal models: general
  results and application to the post-nonlinear causal model}.
\newblock \bibinfo{journal}{\emph{ACM Transactions on Intelligent Systems and
  Technology (TIST)}} \bibinfo{volume}{7}, \bibinfo{number}{2}
  (\bibinfo{year}{2015}), \bibinfo{pages}{1--22}.
\newblock


\bibitem[\protect\citeauthoryear{Zhang and Wang}{Zhang and Wang}{2020}]%
        {zhang2020multi}
\bibfield{author}{\bibinfo{person}{Lu Zhang} {and} \bibinfo{person}{Mingjiang
  Wang}.} \bibinfo{year}{2020}\natexlab{}.
\newblock \showarticletitle{Multi-Scale TCN: Exploring Better Temporal DNN
  Model for Causal Speech Enhancement.}. In
  \bibinfo{booktitle}{\emph{Interspeech}}. \bibinfo{pages}{2672--2676}.
\newblock


\bibitem[\protect\citeauthoryear{Zhang, Shan, and Little}{Zhang
  et~al\mbox{.}}{2022a}]%
        {zhang2022causal}
\bibfield{author}{\bibinfo{person}{Tao Zhang}, \bibinfo{person}{Hao-Ran Shan},
  {and} \bibinfo{person}{Max~A Little}.} \bibinfo{year}{2022}\natexlab{a}.
\newblock \showarticletitle{Causal GraphSAGE: A robust graph method for
  classification based on causal sampling}.
\newblock \bibinfo{journal}{\emph{Pattern Recognition}}  \bibinfo{volume}{128}
  (\bibinfo{year}{2022}), \bibinfo{pages}{108696}.
\newblock


\bibitem[\protect\citeauthoryear{Zhang, Zhao, and LeCun}{Zhang
  et~al\mbox{.}}{2015b}]%
        {zhang2015character}
\bibfield{author}{\bibinfo{person}{Xiang Zhang}, \bibinfo{person}{Junbo Zhao},
  {and} \bibinfo{person}{Yann LeCun}.} \bibinfo{year}{2015}\natexlab{b}.
\newblock \showarticletitle{Character-level convolutional networks for text
  classification}.
\newblock \bibinfo{journal}{\emph{Advances in neural information processing
  systems}}  \bibinfo{volume}{28} (\bibinfo{year}{2015}).
\newblock


\bibitem[\protect\citeauthoryear{Zhang, Zhang, Li, and Smola}{Zhang
  et~al\mbox{.}}{2022b}]%
        {zhang2022automatic}
\bibfield{author}{\bibinfo{person}{Zhuosheng Zhang}, \bibinfo{person}{Aston
  Zhang}, \bibinfo{person}{Mu Li}, {and} \bibinfo{person}{Alex Smola}.}
  \bibinfo{year}{2022}\natexlab{b}.
\newblock \showarticletitle{Automatic chain of thought prompting in large
  language models}.
\newblock \bibinfo{journal}{\emph{arXiv preprint arXiv:2210.03493}}
  (\bibinfo{year}{2022}).
\newblock


\bibitem[\protect\citeauthoryear{Zhao, Hu, Cai, and Liu}{Zhao
  et~al\mbox{.}}{2021}]%
        {zhao2021modeling}
\bibfield{author}{\bibinfo{person}{Shan Zhao}, \bibinfo{person}{Minghao Hu},
  \bibinfo{person}{Zhiping Cai}, {and} \bibinfo{person}{Fang Liu}.}
  \bibinfo{year}{2021}\natexlab{}.
\newblock \showarticletitle{Modeling dense cross-modal interactions for joint
  entity-relation extraction}. In \bibinfo{booktitle}{\emph{Proceedings of the
  twenty-ninth international conference on international joint conferences on
  artificial intelligence}}. \bibinfo{pages}{4032--4038}.
\newblock


\bibitem[\protect\citeauthoryear{Zhao, Liu, Zhao, Chen, and Nie}{Zhao
  et~al\mbox{.}}{2016}]%
        {zhao2016event}
\bibfield{author}{\bibinfo{person}{Sendong Zhao}, \bibinfo{person}{Ting Liu},
  \bibinfo{person}{Sicheng Zhao}, \bibinfo{person}{Yiheng Chen}, {and}
  \bibinfo{person}{Jian-Yun Nie}.} \bibinfo{year}{2016}\natexlab{}.
\newblock \showarticletitle{Event causality extraction based on connectives
  analysis}.
\newblock \bibinfo{journal}{\emph{Neurocomputing}}  \bibinfo{volume}{173}
  (\bibinfo{year}{2016}), \bibinfo{pages}{1943--1950}.
\newblock


\bibitem[\protect\citeauthoryear{Zheng, Shen, Tian, Wang, Bu, and Tian}{Zheng
  et~al\mbox{.}}{2015}]%
        {zheng2015person}
\bibfield{author}{\bibinfo{person}{Liang Zheng}, \bibinfo{person}{Liyue Shen},
  \bibinfo{person}{Lu Tian}, \bibinfo{person}{Shengjin Wang},
  \bibinfo{person}{Jiahao Bu}, {and} \bibinfo{person}{Qi Tian}.}
  \bibinfo{year}{2015}\natexlab{}.
\newblock \showarticletitle{Person re-identification meets image search}.
\newblock \bibinfo{journal}{\emph{arXiv preprint arXiv:1502.02171}}
  (\bibinfo{year}{2015}).
\newblock


\bibitem[\protect\citeauthoryear{Zheng, Aragam, Ravikumar, and Xing}{Zheng
  et~al\mbox{.}}{2018}]%
        {zheng2018dags}
\bibfield{author}{\bibinfo{person}{Xun Zheng}, \bibinfo{person}{Bryon Aragam},
  \bibinfo{person}{Pradeep~K Ravikumar}, {and} \bibinfo{person}{Eric~P Xing}.}
  \bibinfo{year}{2018}\natexlab{}.
\newblock \showarticletitle{Dags with no tears: Continuous optimization for
  structure learning}.
\newblock \bibinfo{journal}{\emph{Advances in neural information processing
  systems}}  \bibinfo{volume}{31} (\bibinfo{year}{2018}).
\newblock


\bibitem[\protect\citeauthoryear{Zhou, Huang, Zhang, Zhu, and Liu}{Zhou
  et~al\mbox{.}}{2018}]%
        {zhou2018emotional}
\bibfield{author}{\bibinfo{person}{Hao Zhou}, \bibinfo{person}{Minlie Huang},
  \bibinfo{person}{Tianyang Zhang}, \bibinfo{person}{Xiaoyan Zhu}, {and}
  \bibinfo{person}{Bing Liu}.} \bibinfo{year}{2018}\natexlab{}.
\newblock \showarticletitle{Emotional chatting machine: Emotional conversation
  generation with internal and external memory}. In
  \bibinfo{booktitle}{\emph{Proceedings of the AAAI Conference on Artificial
  Intelligence}}, Vol.~\bibinfo{volume}{32}.
\newblock


\bibitem[\protect\citeauthoryear{Zhou, Yu, and Chen}{Zhou
  et~al\mbox{.}}{2022}]%
        {zhou2022causality}
\bibfield{author}{\bibinfo{person}{Wanqi Zhou}, \bibinfo{person}{Shujian Yu},
  {and} \bibinfo{person}{Badong Chen}.} \bibinfo{year}{2022}\natexlab{}.
\newblock \showarticletitle{Causality detection with matrix-based transfer
  entropy}.
\newblock \bibinfo{journal}{\emph{Information Sciences}}  \bibinfo{volume}{613}
  (\bibinfo{year}{2022}), \bibinfo{pages}{357--375}.
\newblock


\bibitem[\protect\citeauthoryear{Zhu, Ng, and Chen}{Zhu et~al\mbox{.}}{2019}]%
        {zhu2019causal}
\bibfield{author}{\bibinfo{person}{Shengyu Zhu}, \bibinfo{person}{Ignavier Ng},
  {and} \bibinfo{person}{Zhitang Chen}.} \bibinfo{year}{2019}\natexlab{}.
\newblock \showarticletitle{Causal discovery with reinforcement learning}.
\newblock \bibinfo{journal}{\emph{arXiv preprint arXiv:1906.04477}}
  (\bibinfo{year}{2019}).
\newblock


\bibitem[\protect\citeauthoryear{Zhu, Zhang, Feng, Yang, Wang, and He}{Zhu
  et~al\mbox{.}}{2022}]%
        {zhu2022mitigating}
\bibfield{author}{\bibinfo{person}{Xinyuan Zhu}, \bibinfo{person}{Yang Zhang},
  \bibinfo{person}{Fuli Feng}, \bibinfo{person}{Xun Yang},
  \bibinfo{person}{Dingxian Wang}, {and} \bibinfo{person}{Xiangnan He}.}
  \bibinfo{year}{2022}\natexlab{}.
\newblock \showarticletitle{Mitigating hidden confounding effects for causal
  recommendation}.
\newblock \bibinfo{journal}{\emph{arXiv preprint arXiv:2205.07499}}
  (\bibinfo{year}{2022}).
\newblock


\bibitem[\protect\citeauthoryear{Zimbor{\'a}s, Farrelly, Farkas, and
  Masanes}{Zimbor{\'a}s et~al\mbox{.}}{2022}]%
        {zimboras2022does}
\bibfield{author}{\bibinfo{person}{Zolt{\'a}n Zimbor{\'a}s},
  \bibinfo{person}{Terry Farrelly}, \bibinfo{person}{Szil{\'a}rd Farkas}, {and}
  \bibinfo{person}{Lluis Masanes}.} \bibinfo{year}{2022}\natexlab{}.
\newblock \showarticletitle{Does causal dynamics imply local interactions?}
\newblock \bibinfo{journal}{\emph{Quantum}}  \bibinfo{volume}{6}
  (\bibinfo{year}{2022}), \bibinfo{pages}{748}.
\newblock


\end{thebibliography}


\end{document}